\newcommand{\final}{1}
\documentclass[sigconf]{acmart} 

\acmSubmissionID{183}

\usepackage{cuted}
\usepackage[ruled]{algorithm2e} 

\SetAlFnt{\small}
\SetAlCapFnt{\small}
\SetAlCapNameFnt{\small}
\SetAlCapHSkip{0pt}

\usepackage{amsthm}

\usepackage{amssymb}
\usepackage{textcomp}
\usepackage{wrapfig}
\usepackage{subfig}
\usepackage{color}
\usepackage{xspace}
\usepackage{subfig}
\usepackage{enumitem}
\usepackage{hyperref}
\usepackage{multirow}
\usepackage{soul}
\usepackage{ifthen}

\acmJournal{TOG}

\newcommand{\onethirdfigurewidth}{0.32}
\newcommand{\onefourthfigurewidth}{0.24}
\newcommand{\onesixthfigurewidth}{0.16}

\newcommand{\revision}[1]{{\color{orange} #1}}
\newcommand{\delete}[1]{{\color{orange} \st{#1}}}

\ifthenelse{\equal{\final}{1}}
{
\renewcommand{\revision}[1]{#1}
\renewcommand{\delete}[1]{}
}
{}

\newcommand{\yl}[1]{#1} 

\newcommand{\yht}[1]{#1} 

\newcommand{\sysname}{TRAvatar\xspace} 

\newcommand{\light}{l}
\newcommand{\direction}{\mathbf{d}}

\newcommand{\etal}{\emph{et al.}}
\newcommand{\ie}{\emph{i.e.}}

\newcommand{\loss}{\mathcal{L}}
\newcommand{\totalloss}{\loss_{total}}
\newcommand{\dataloss}{\loss_{img}}

\newcommand{\regularizationloss}{\loss_{reg}}

\newcommand{\encoder}{\mathcal{E}}
\newcommand{\motionencoder}{\encoder_{m}}
\newcommand{\decoder}{\mathcal{D}}
\newcommand{\meshdecoder}{\decoder_{mesh}}
\newcommand{\transformdecoder}{\decoder_{T}}
\newcommand{\opacitydecoder}{\decoder_{\alpha}}
\newcommand{\colordecoder}{\decoder_{rgb}}

\newcommand{\latentcode}{z}
\newcommand{\expressioncode}{\latentcode_{e}}

\newcommand{\vertex}{\mathbf{v}}
\newcommand{\pixel}{p}

\newcommand{\pos}{\mathbf{v}}
\newcommand{\templatepos}{\hat{\pos}}
\newcommand{\residualpos}{\delta\pos}
\newcommand{\featuremap}{\mathcal{F}}
\newcommand{\linearfeaturemap}{\featuremap_\mathrm{{lin}}}
\newcommand{\nonlinearfeaturemap}{\featuremap_\mathrm{{nlin}}}

\newcommand{\laplacian}{\mathbf{L}}
\newcommand{\blendshapes}{\mathbf{B}}

\newcommand{\voxelcolor}{V_{rgb}}
\newcommand{\voxelopacity}{V_{\alpha}}

\newcommand{\opacityvalue}{I_{\alpha}}
\newcommand{\finalimage}{\hat{I}}
\newcommand{\renderedimage}{I_{rgb}}
\newcommand{\foregroundimage}{I_{rgb}}
\newcommand{\backgroundimage}{I_{BG}}

\copyrightyear{2023}
\acmYear{2023}
\setcopyright{acmlicensed}\acmConference[SA Conference Papers '23]{SIGGRAPH Asia 2023 Conference Papers}{December 12--15, 2023}{Sydney, NSW, Australia}
\acmBooktitle{SIGGRAPH Asia 2023 Conference Papers (SA Conference Papers '23), December 12--15, 2023, Sydney, NSW, Australia}
\acmPrice{15.00}
\acmDOI{10.1145/3610548.3618138}
\acmISBN{979-8-4007-0315-7/23/12}

\begin{document}

\title{Towards Practical Capture of High-Fidelity Relightable Avatars}


\author{Haotian Yang}
\affiliation{
\institution{Kuaishou Technology}
\country{China}
}
\email{yanghaotian03@kuaishou.com}

\author{Mingwu Zheng}
\affiliation{
\institution{Kuaishou Technology}
\country{China}
}
\email{zhengmingwu@kuaishou.com}

\author{Wanquan Feng}
\affiliation{
\institution{Kuaishou Technology}
\country{China}
}
\email{fengwanquan@kuaishou.com}

\author{Haibin Huang}
\affiliation{
\institution{Kuaishou Technology}
\country{China}
}
\email{huanghaibin03@kuaishou.com}

\author{Yu-Kun Lai}
\affiliation{
\institution{Cardiff University}
\country{United Kingdom}
}
\email{laiy4@cardiff.ac.uk}

\author{Pengfei Wan}
\affiliation{
\institution{Kuaishou Technology}
\country{China}
}
\email{wanpengfei@kuaishou.com}

\author{Zhongyuan Wang}
\affiliation{
\institution{Kuaishou Technology}
\country{China}
}
\email{wangzhongyuan@kuaishou.com}

\author{Chongyang Ma*}
\affiliation{
\institution{Kuaishou Technology}
\country{China}
}
\email{chongyangma@kuaishou.com}

\begin{teaserfigure}
\centering
    \includegraphics[width=\linewidth]{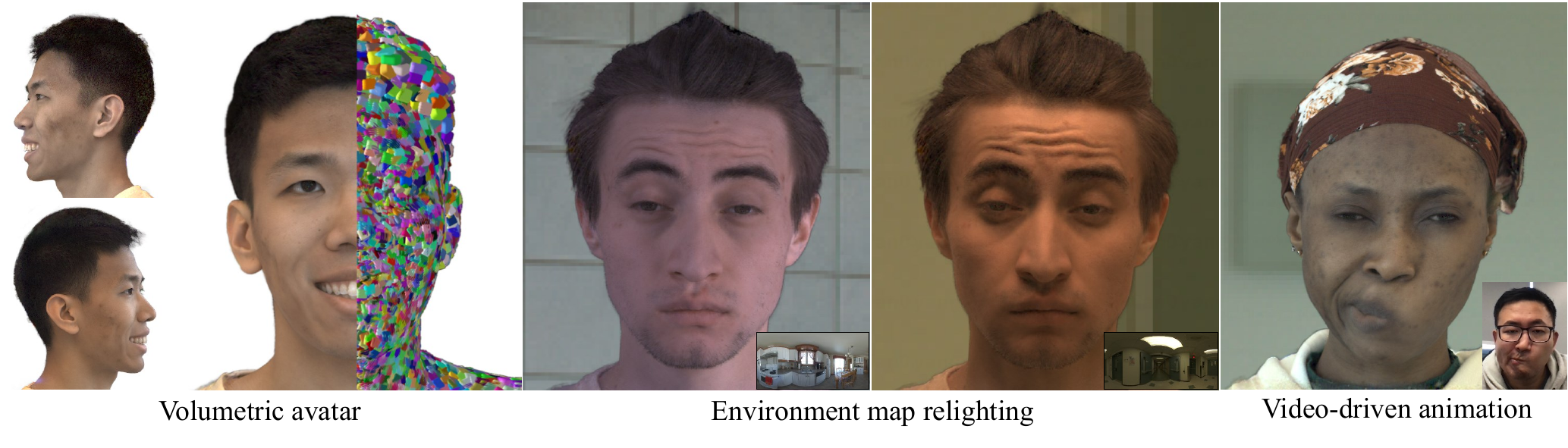}
    \captionof{figure}{We present \sysname, a novel framework to capture and \yl{reconstruct} high-fidelity volumetric avatars. Trained \yl{efficiently} end-to-end on multi-view image sequences under varying illuminations, our virtual avatars can be relighted and animated in real-time of high fidelity.}
\label{fig:teaser}
\end{teaserfigure}

\renewcommand{\shortauthors}{H. Yang, M. Zheng, W. Feng, H. Huang, Y.-K. Lai, P. Wan, Z. Wang, and C. Ma}
\renewcommand{\authors}{Haotian Yang, Mingwu Zheng, Wanquan Feng, Haibin Huang, Yu-Kun Lai, Pengfei Wan, Zhongyuan Wang, and Chongyang Ma}

\begin{abstract}
\ifthenelse{\equal{\final}{1}}
{
\renewcommand{\thefootnote}{}
\footnotetext{* Corresponding author.}
}
{}
In this paper, we propose a novel framework, Tracking-free Relightable Avatar (\sysname), for capturing and reconstructing high-fidelity 3D avatars. 
Compared to previous methods, \sysname  works in a more practical and efficient setting. Specifically, \sysname is trained with dynamic image sequences captured in a Light Stage under varying lighting conditions, enabling realistic relighting and real-time animation for avatars in diverse scenes. 
Additionally, \sysname allows for tracking-free avatar capture and obviates the need for accurate surface tracking under varying illumination conditions.
Our contributions are two-fold: First, we propose a novel network architecture that explicitly builds on and ensures the satisfaction of the linear nature of lighting. Trained on simple group light captures, \sysname can predict the appearance in real-time with a single forward pass, achieving high-quality relighting effects under illuminations of arbitrary environment maps. Second, we jointly optimize the facial geometry and relightable appearance from scratch based on image sequences, where the tracking is implicitly learned. This tracking-free approach brings robustness for establishing temporal correspondences between frames under different lighting conditions. Extensive qualitative and quantitative experiments demonstrate that our framework achieves superior performance for photorealistic avatar animation and relighting.

\end{abstract}

%
%


%
%



\maketitle

\section{Introduction}

In this work, we focus on the 
\yl{capture and reconstruction}
of high-fidelity avatars in a Light Stage environment. As virtual representations of humans, avatars are crucial components in various downstream applications, such as video games, virtual reality, telepresence, and more~\cite{lombardi2018deep, schwartz2020eyes, bi2021deep, guo2019relightables, moser2021semi}. 

Avatar creation has been a popular and challenging research topic in computer graphics and computer vision for decades. Despite considerable progress in this field, there are still many challenges to overcome, including 
\yl{expensive and sophisticated setup for avatar capture, lack of support for realistic relighting and animation, and high resource demands making training time-consuming and real-time deployment difficult to achieve.}
\yht{Traditional frameworks based on graphics pipeline, including geometry reconstruction~\cite{guo2019relightables, beeler2011high, wu2018deep, collet2015high, beeler2010high, riviere2020single} and physically-inspired reflectance capture~\cite{debevec2000acquiring, ma2007rapid, moser2021semi, weyrich2006analysis, ghosh2011multiview}, are often difficult to set up and lack robustness, especially for dynamic subjects and non-facial parts. }
Recent deep learning based methods~\cite{lombardi2018deep, bi2021deep, lombardi2021mixture, remelli2022drivable, cao2022authentic} have demonstrated promising improvements for avatar representation by approximating the geometry and appearance with \yl{neural} networks. However, most learning-based methods struggle to handle relighting effectively and have computationally expensive pre-processing and training steps that cannot meet the aforementioned requirements.

To this end, we propose a novel framework, Tracking-free Relightable Avatar (\sysname), that can 
\yl{circumvent}
the above obstacles, supporting efficient \yl{capture}, high-quality 
\yl{reconstruction},
as well as real-time animation and relighting (see Figure~\ref{fig:teaser}).
Specifically, we improve the entire pipeline at its \yl{two primary} stages, \ie, both the data \yl{capture} and avatar reconstruction. 

For the data \yl{capture} stage, we record \yl{a} subject's performance under various expressions and lighting conditions. To faithfully reproduce the identity and detailed expressions of a specific subject, both dynamic geometry and reflectance should be captured. Considering the complexity of lighting conditions, it is non-trivial for the avatar network to directly learn the mapping from environment maps to the appearance. Furthermore, it is challenging to achieve satisfactory decoupling of lighting and other input conditions. To overcome this challenge, we take advantage of the prior knowledge of lighting, specifically its linear nature, to guide the network design. We design a network structure that \yl{explicitly exploits and guarantees to satisfy}
the linear nature of lighting, making it easy to train and enabling excellent generalization ability. Trained on dynamically captured image sequences in simple controllable group light illumination~\cite{bi2021deep}, our model can predict the appearance under arbitrary and complex lighting condition in a single forward pass, which facilitates real-time environment relighting. The learned disentangled representation also allows our data-driven avatar to be animated, relighted, and \yl{rendered} under novel viewpoints.


For avatar reconstruction, we generate a 3D model from captured data that can be manipulated in real time. It is a challenging task to estimate temporal correspondences between captured frames with different lighting conditions. Previous learning-based methods typically rely on \yl{a} pre-processing step to compute explicit tracked geometry \yl{(as a deformable base mesh)}, which is computationally expensive and \yl{not} robust to varying light conditions. Therefore, we propose to jointly optimize the relightable appearance and latent geometry from scratch from image sequences, where the tracking 
is implicitly learned. Different from previous methods that separate mesh tracking and avatar creation in two stages, \yl{our tracking-free approach implicitly learns the dynamic deformation of the base mesh directly from the multi-view captured data, along with the relightable appearance in a joint optimization process.}
\yl{In addition to being much more efficient, this joint optimization} allows our model to be directly trained on images in varying illumination, which is challenging for traditional explicit surface tracking. 

Our experiments with \sysname show its effectiveness in creating high-quality and authentic avatars that can be animated and relighted in real-time with superior visual quality and computational efficiency compared to previous methods. 

In summary, our contributions are:

\begin{itemize}[leftmargin=*]

\item We present \sysname, a practical and efficient capture solution for creating high-fidelity avatars that can be animated and relighted in real time. 

\item We propose a novel network architecture that \yl{explicitly exploits} the linear nature of lighting \yl{to improve generalizability, enabling} real-time relighting \yl{with high realism} \yl{for} given environment maps.

\item We propose to jointly optimize the relightable appearance and latent geometry of avatars from image sequences captured under varying lighting conditions, allowing more efficient and effective creation of \yl{relightable} virtual avatars.

\item We demonstrate that \sysname outperforms previous methods in terms of both visual quality and computational efficiency.

\end{itemize}

\section{Related Work}
\label{sec:related}

Creating a data-driven, relightable facial avatar of a specific subject typically involves capturing both dynamic geometry and reflectance. This is followed by constructing a parametric model from the captured data, or alternatively, employing image-based relighting techniques to synthesize the output.
Below, we provide a concise overview of 
most relevant methods.

\paragraph{Geometry and reflectance acquisition.}
3D face reconstruction and performance capture have been \yl{active} research topics for decades.
Accordingly, sophisticated 3D scanning systems have been developed for both static geometry reconstruction~\cite{beeler2010high,ghosh2011multiview} and dynamic performance capture~\cite{beeler2011high, bradley2010high, huang2011leveraging, collet2015high, dou2017motion2fusion, guo2019relightables}.
These methods utilize either multi-view stereo \revision{(MVS)} or structured light for point cloud acquisition and then estimate the deforming geometry to achieve temporally consistent mesh tracking.
\revision{The tracking process often involves time-consuming MVS reconstruction for thousands of frames and dense optical-flow optimization, while existing real-time face tracking algorithms cannot achieve satisfactory accuracy.}

Besides, another crucial aspect of realistic relightable avatars is to estimate the way in which the light interacts with the subject, \ie, the reflectance property. 
Previous methods usually assume physically-inspired reflectance functions modeled as bidirectional reflectance distribution function (BRDF)~\cite{schlick1994inexpensive} and solve the parameters by observing the appearance under active or passive lighting.
Active lighting methods typically require specialized setups with controllable illuminations and synchronized cameras. Debevec~\etal~\shortcite{debevec2000acquiring} pioneer in using \yl{a} Light Stage for facial reflectance acquisition. One-light-at-a-time (OLAT) capture is performed to obtain the dense reflectance field.
Later, polarized~\cite{ma2007rapid, ghosh2011multiview, zhang2022video} and color gradient illuminations~\cite{fyffe2015single, guo2019relightables} are used for rapid acquisition.
Passive capture methods have significantly reduced the necessity for an expensive capture setup.
For example, Riviere~\etal~\shortcite{riviere2020single} and Zheng~\etal~\shortcite{zheng2023neuface} propose to estimate physically-based facial textures via inverse rendering. 

\paragraph{3D face modeling.}
Modeling of facial geometry and appearance has been a fundamental component of human related tasks in computer graphics and computer vision. The seminal work on 3D morphable models (3DMMs)~\cite{blanz1999morphable, cao2013facewarehouse, yang2020facescape} employs \yl{Principal} Component Analyze (PCA) to derive the shape basis from head scans.
Despite its widespread use in various applications such as single-view face reconstruction and tracking~\cite{zhu2017face, thies2016face2face, dou2017end}, the shape space of 3DMMs is limited by its low-dimensional linear representation.
\yl{Follow-up} methods separate the parametric space dimensions~\cite{vlasic2006face, li2017learning, jiang2019disentangled} or use local deformation \yl{models}~\cite{wu2016anatomically} to enhance the representation power of the morphable model. 

In recent years, deep learning based methods~\cite{bagautdinov2018modeling, tran2018nonlinear, tran2019learning, zhang2022video, zheng2022imface} have been widely used to achieve impressive realism in face modeling.
Lombardi~\etal~\shortcite{lombardi2018deep} utilize \yl{a} Variational Autoencoder (VAE) \cite{kingma2013auto} to jointly model the mesh and dynamic texture, which is used for monocular~\cite{yoon2019self} and binocular~\cite{cao2021real} facial performance capture. Bi~\etal~\shortcite{bi2021deep} propose to extend the \yl{VAE-based} deep appearance model by capturing the dynamic performance under controllable group light illuminations to enable relighting.

While mesh-based methods typically require dense correspondence based on sophisticated surface tracking algorithms~\cite{beeler2011high, wu2018deep} for training and degrade in non-facial regions, recent progress in neural volumetric rendering further enables photorealistic avatar creation.
Lombardi~\etal~\shortcite{lombardi2021mixture} propose MVP \yl{(Mixture of Volumetric Primitives)}, a hybrid volumetric and primitive-based representation that produces high-fidelity rendering results with efficient runtime performance.
More recently, Li~\etal~\shortcite{li2023megane} extend MVP with eyeglasses to be relightable following \cite{bi2021deep}. But it requires additional efforts for real-time relighting.

Some other methods have been proposed to create a facial avatar from monocular videos~\cite{zielonka2022instant, gao2022reconstructing} or RGB-D input~\cite{cao2022authentic} without a specialized capturing apparatus.
However, these approaches do not provide a relightable appearance, and their quality cannot match that of avatars built from industrial capture setups.

\begin{figure}
    \centering
    \includegraphics[width=1.0\linewidth]{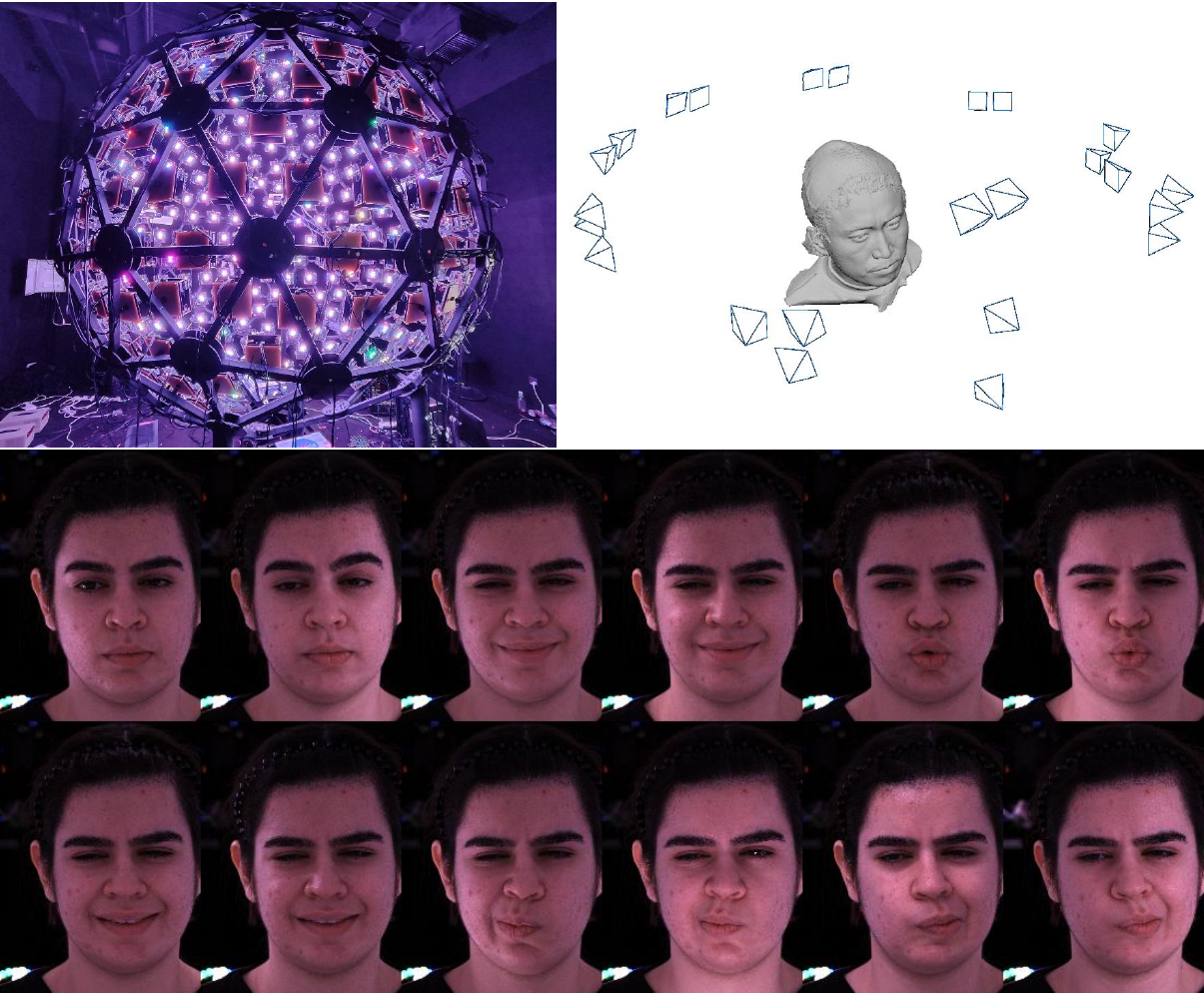}
    \caption{Illustration of our capture setup. Top left: Our customized capturing apparatus. Top right: The layout of 24 cameras. Bottom: Snapshots of captured frames from the frontal camera in a recording. Both the expression and the lighting condition change across different frames.}
\label{fig:hardware}
\end{figure}

\ifthenelse{\equal{\final}{0}}
{
\newcommand{\overviewfigwidth}{0.9}
}
{
\newcommand{\overviewfigwidth}{1}
}

\begin{figure*}
    \centering
    \includegraphics[width=\overviewfigwidth\linewidth]{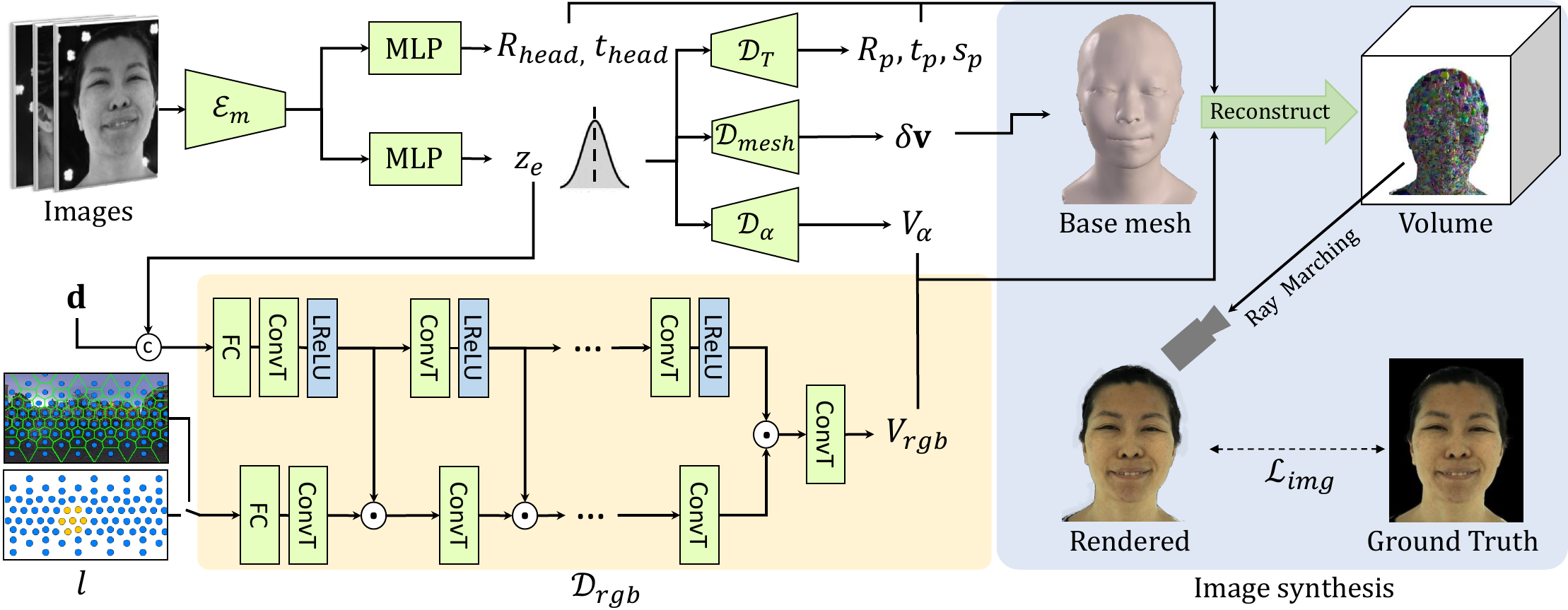}
    \caption{The pipeline of our framework. \protect{\sysname} is a relightable volumetric avatar \yl{representation} learned from multiview image sequences, \yl{including dynamic expressions and varying illuminations}. For each frame, a motion encoder $\motionencoder$ forecasts the disentangled global \yl{rigid} transformation $\{R_{head}, t_{head}\}$ and expression code $\expressioncode$. With the given expression code, lighting condition $\light$, and view direction $\direction$, a series of decoders subsequently predict the \yl{base} mesh and the volumetric primitives \yl{mounted on it}. Notably, a physically-inspired appearance decoder $\colordecoder$ (detailed in Section~\ref{sec:relightable_appearance}) is proposed to facilitate network training. Ultimately, the avatar representation is computed and then rendered, adaptable to any viewpoint and any lighting condition.
    }
\label{fig:method}
\end{figure*}

\paragraph{Image-based relighting.}
In contrast to model-based reflectance acquisition approaches, image-based relighting addresses the problem from an orthogonal perspective.
By exploiting the linear nature of light transport, Debevec~\etal~\shortcite{debevec2000acquiring} \yl{propose} to add up hundreds of images of densely sampled reflectance fields from OLAT capture to synthesize rendering results under novel lighting conditions.
Subsequently, the number of sampled images is reduced by using specifically designed illumination patterns~\cite{peers2009compressive, reddy2012frequency} or employing sparse sampling~\cite{fuchs2007adaptive, wang2009kernel}. Xu~\etal~\shortcite{xu2018deep} \yl{propose} to train a network for relighting a scene from only five input images. Meka~\etal~\shortcite{meka2019deep} show that the full 4D reflectance field of human faces can be regressed from two images under color gradient light illumination. Sun~\etal~\shortcite{sun2020light} propose a learning-based method to achieve higher lighting resolution than the original Light Stage OLAT capture.
Although these approaches achieve photorealistic rendering under novel lighting conditions, they only work from fixed viewpoints.

Meka~\etal~\shortcite{meka2020deep} achieve relightable free viewpoint rendering of dynamic facial performance by extending Meka~\etal~\shortcite{meka2019deep} with explicit 3D reconstruction and multi-view capture.
However, they extract pixel-aligned features from captured raw images under color gradient light illumination to build relightable textures, which limits its usage scenarios to performance replay.
In contrast, our approach enables the creation of virtual avatars that not only \yl{allows} for free viewpoint rendering with a relightable appearance but also \yl{possesses} the capability \yl{of being} controlled by an animation sequence of a different subject.

\section{Capturing Apparatus}
\label{sec:capture}

To create an animatable and relightable avatar with ultra-high realism and specific identity, it is necessary to capture its performance under various expressions and lighting conditions.
To this end, we have constructed an apparatus following the design principles of Light Stages~\cite{debevec2012light, guo2019relightables}. Our customized capturing apparatus is shown in Figure~\ref{fig:hardware}.

Our Light Stage, installed on a spherical structure with a 3.6-meter diameter, comprises 356 lighting units and 24 machine vision cameras.
We strategically place the cameras to capture the subject from multiple angles, and arrange the lighting units for precise control over illumination conditions. The Light Stage is placed in a dark room to prevent environment light interference.

\paragraph{Lighting units.}
The 356 lighting units are uniformly mounted on the sphere and are oriented towards the center. Each customized lighting unit comprises 132 high-brightness Light-Emitting Diodes (LEDs) that are controlled by a programmable embedded system. The LEDs are equipped with diffusers and lenses to ensure equal density illumination at the center.

There are five different types of LEDs on the lighting unit, namely red, green, blue, white 4500K, and white 6500K. The setup follows the latitude-longitude polarization as proposed in~\cite{ghosh2011multiview}, and each type of LED is grouped into three categories with different polarization arrangements. The brightness of each group of lights can be adjusted independently using Pulse Width Modulation up to 100KHz.
All the lighting units are connected to a central control unit and a computer via a CAN bus. The lighting pattern can be shuffled within 2ms, allowing us to capture the subject's performance under various lighting conditions quickly.

\paragraph{Cameras.}

Our apparatus includes 24 machine vision cameras installed around the sphere, with a focus on the center. The cameras consist of four 31M RGB cameras, $12$ 5M RGB cameras, and eight 12M monochrome cameras. The trigger ports of these cameras are linked to the central control unit, which synchronizes the cameras and lighting units to capture the subject's performance under various lighting conditions. We have disabled postprocessing features such as automatic gain adjustments in the cameras to ensure a linear response to the illuminance.

Depending on the camera types, we transmit the captured images to seven PCs via 10G Ethernet or USB ports. We calibrate the camera array with a 250mm calibration sphere similar to \cite{beeler2010high} and undistort the images to ensure high-quality reconstruction.
The mean reprojection error is less than 0.4 pixels, which facilitates high-quality creation of the target avatar.

\section{Method}
\label{sec:method}

In this section, we formally introduce our novel framework, namely \sysname, which learns a disentangled representation for the target avatar to be animated, relighted, and rendered from novel viewpoints.
As shown in Figure~\ref{fig:method}, our approach is based on a variational autoencoder (VAE)~\cite{kingma2013auto} architecture, 
where the latent space is designed to be disentangled with linear responses to varying lighting \yl{conditions} , providing efficient and accurate modeling of dynamic geometry and reflectance fields.

We will first describe the details of our \sysname, including the training framework and network \yl{architecture} (Section~\ref{sec:core_algorithm}).
The details of our specifically designed appearance decoder will be explained in Section~\ref{sec:relightable_appearance}.
We will then \revision{describe how to use }\delete{discuss }our Light Stage \delete{setup }for data capture under various illuminations (Section~\ref{sec:data_acquisition}).
Finally, we will introduce the loss functions and regularization terms used for end-to-end network training (Section~\ref{sec:network_training}).

\subsection{\sysname}
\label{sec:core_algorithm}

\begin{figure}
    \centering
    \begin{minipage}[t]{\onethirdfigurewidth\linewidth}
        \centering
        \includegraphics[width=\linewidth]{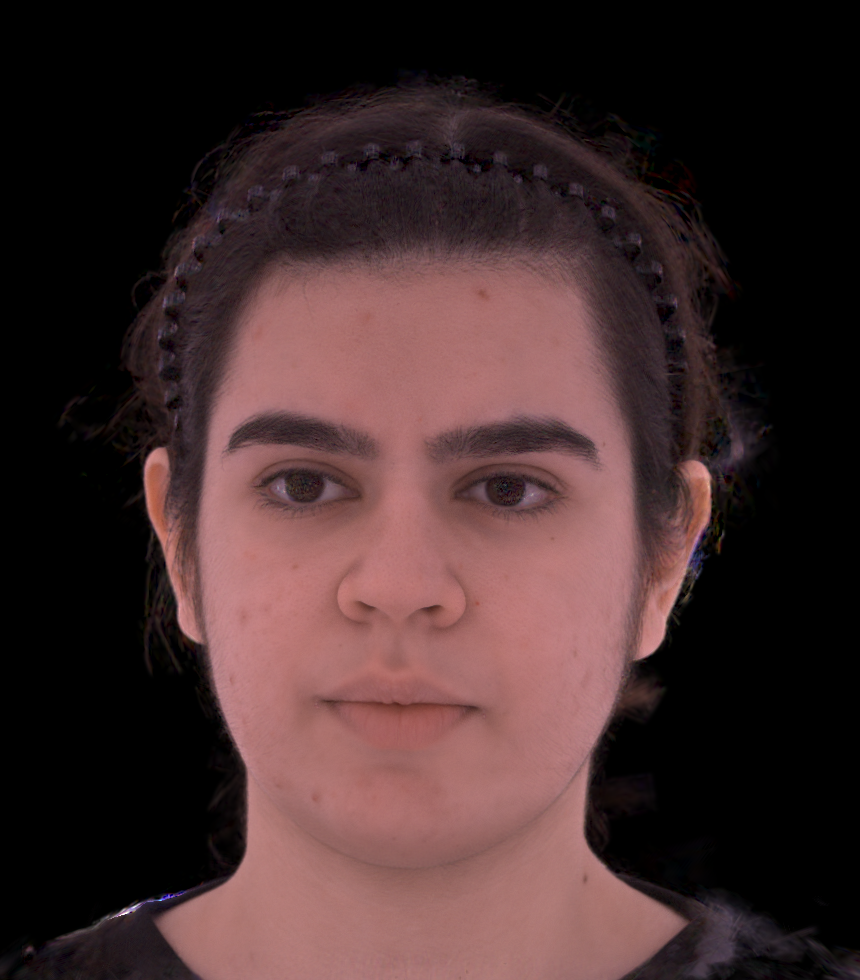}
        \includegraphics[width=\linewidth]{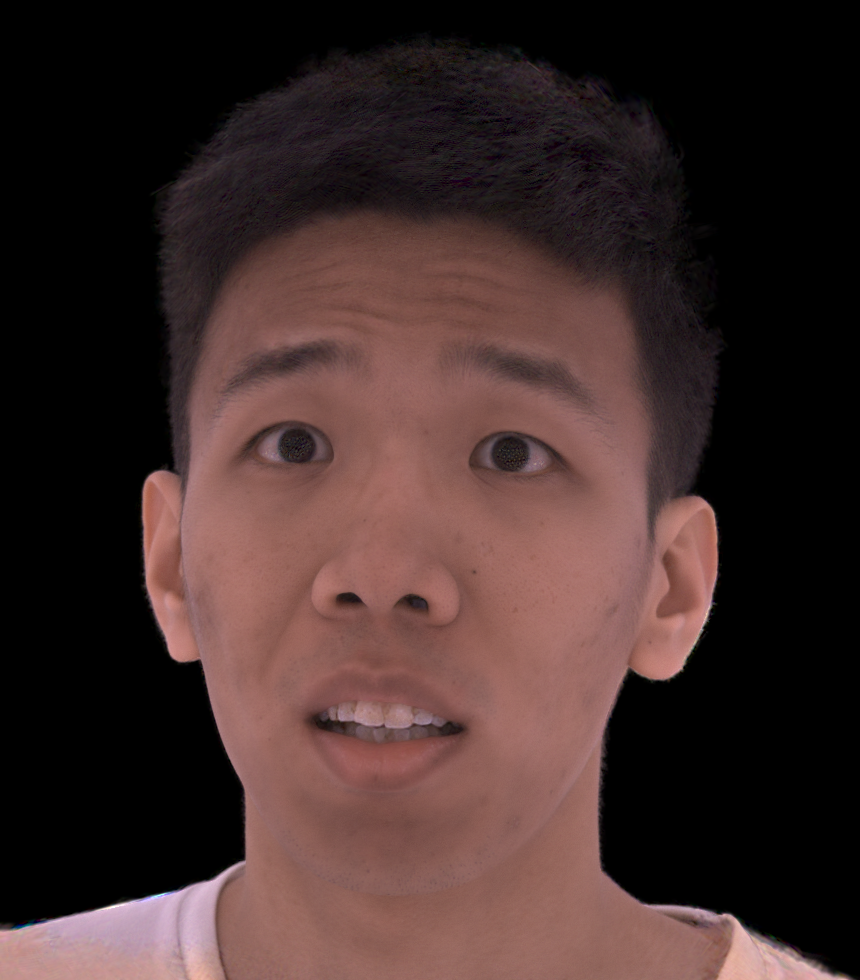}
        \subfloat{Captured image}
    \end{minipage}
    \begin{minipage}[t]{\onethirdfigurewidth\linewidth}
        \centering
        \includegraphics[width=\linewidth]{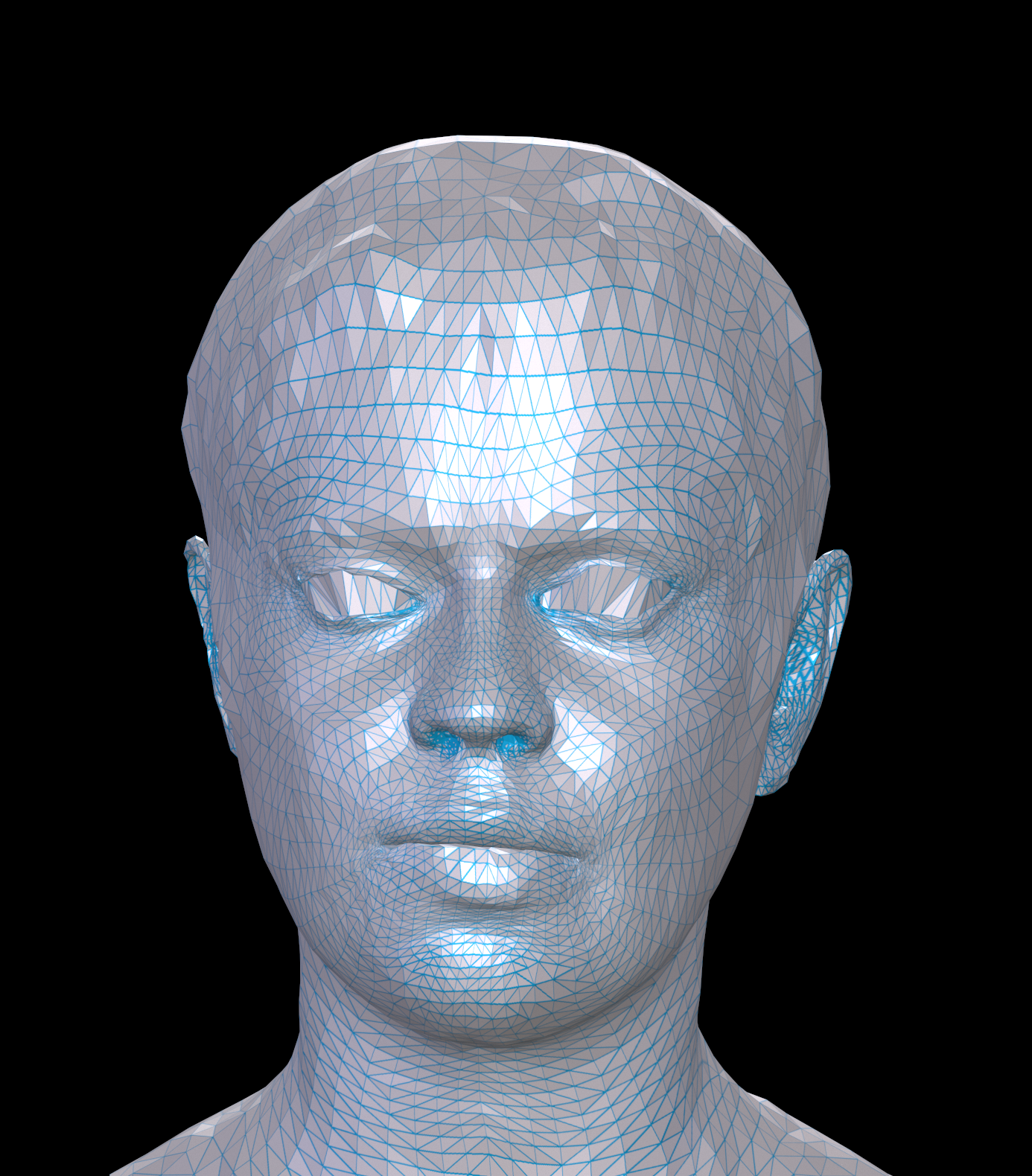}
        \includegraphics[width=\linewidth]{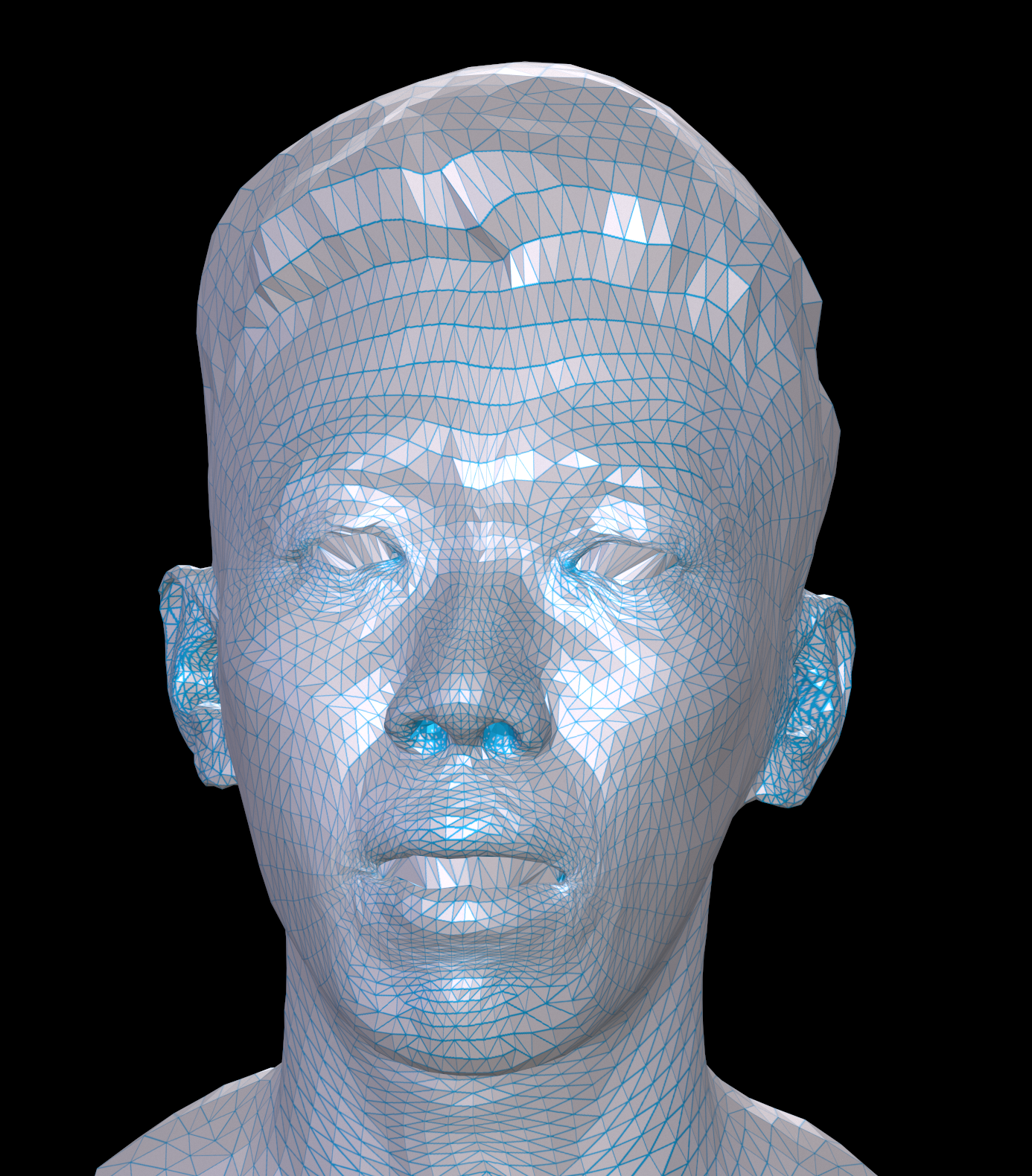}
        \subfloat{Base mesh}
    \end{minipage}
    \begin{minipage}[t]{\onethirdfigurewidth\linewidth}
        \centering
        \includegraphics[width=\linewidth]{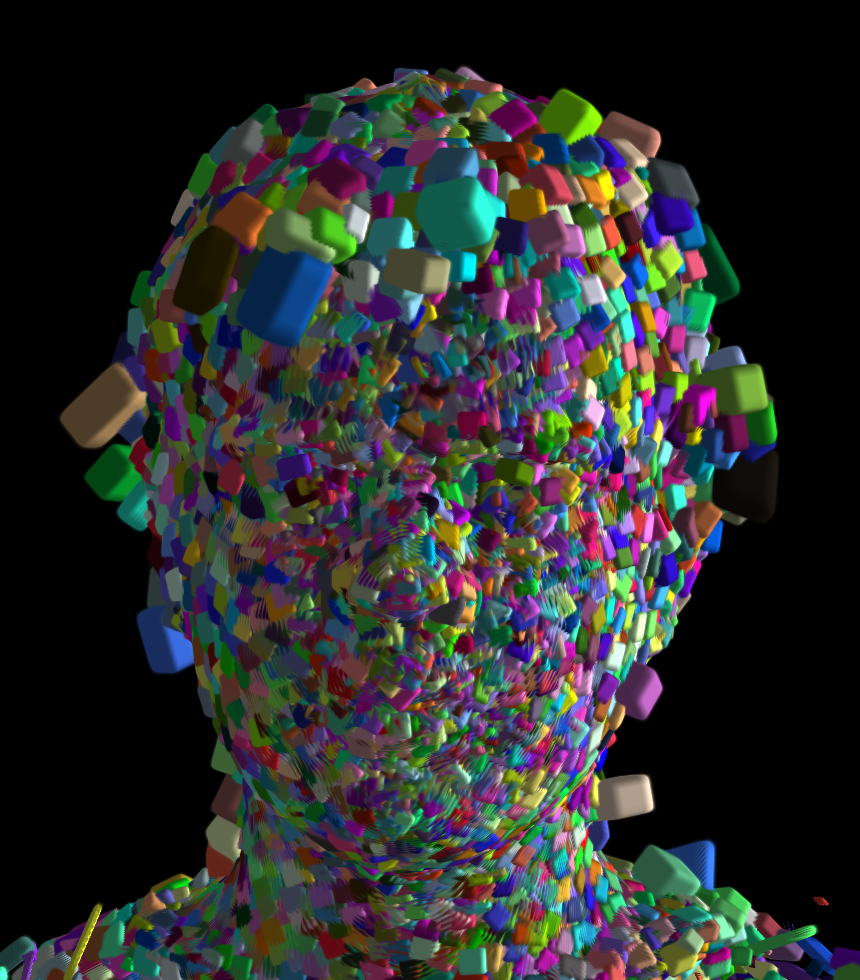}
        \includegraphics[width=\linewidth]{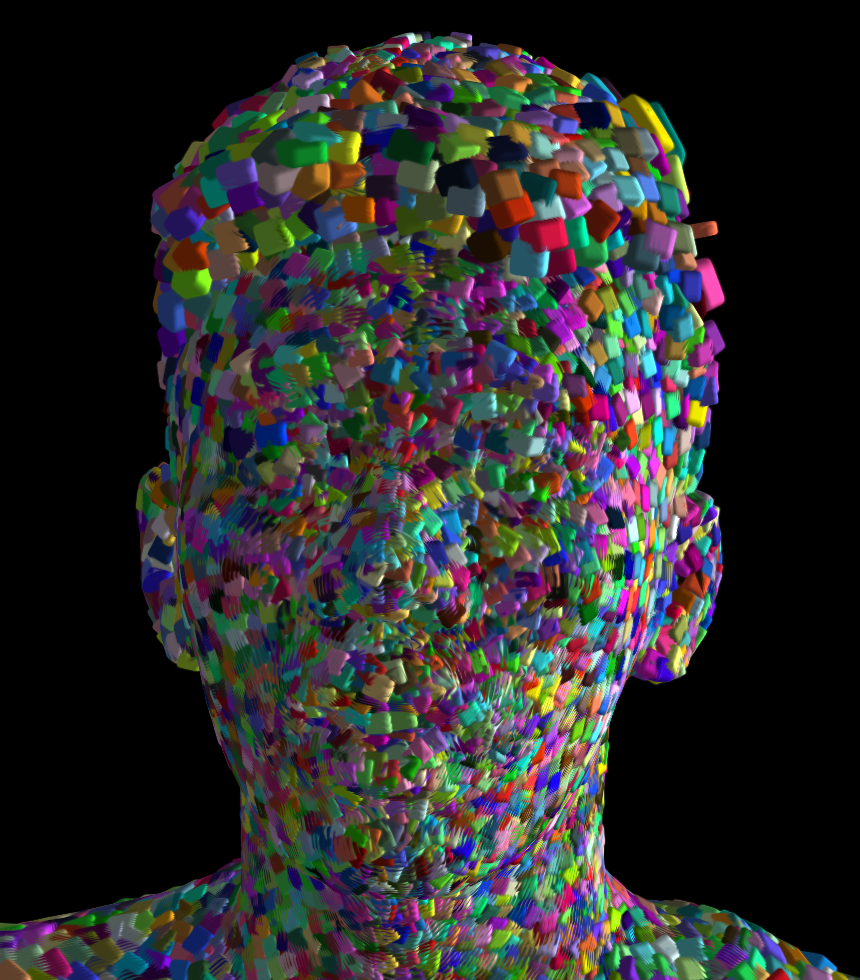}
        \subfloat{Volumetric primitives}
    \end{minipage}
    \caption{Illustration of our hybrid avatar representation. The base mesh and the volumetric primitives have consistent structures which provide flexible control such as video driven animation.}
\label{fig:topology}
\end{figure}

Our volumetric avatar is built upon Mixed Volumetric Primitives (MVP)~\cite{lombardi2021mixture}, which is a generalized hybrid representation using both a base mesh and volumetric primitives (see Figure~\ref{fig:topology}).
Each primitive is mounted to the base mesh and is represented as a volumetric grid with a resolution of $M^3$.
We set $M=8$ in our implementation.

Inspired by the success of image based relighting methods, our lighting condition is modeled as a vector $l \in {\mathbb{R}_+}^{356}$ representing the incoming light field of 356 densely \yl{sampled} directions corresponding to the light positions of the Light Stage. We employ a VAE based architecture to train our relightable avatar.
Different from previous methods~\cite{bi2021deep, remelli2022drivable}, we do not require tracked geometry in training.
Note that the motion of a human head can be separated into global rigid motion and expression related motion.
\yl{We} utilize a motion encoder $\motionencoder$, to predict the disentangled motion.
During training, for each frame, the convolutional motion encoder $\motionencoder$ takes a subset of the camera views as input and outputs the global head rotation $R_{head} \in SO(3)$ and translation $t_{head} \in \mathbb{R}^{3}$ as well as the mean $\mu \in \mathbb{R}^{256}$ and the standard deviation $\sigma \in {\mathbb{R}_+}^{256}$ of a Gaussian distribution $\mathcal{N}(\mu, \sigma^{2})$. The expression code $\expressioncode \in \mathbb{R}^{256}$ is sampled from this Gaussian distribution and represents expression related motion.

Taking the expression code $\expressioncode$, the lighting condition $\light$, and the view direction $\direction$ as input, we use several decoders to predict the base mesh and volumetric primitives for output synthesis.
Specifically, a mesh decoder $\meshdecoder:\mathbb{R}^{256} \rightarrow \mathbb{R}^{3 \times N_{mesh}}$, which is a multilayer perceptron, predicts the residual vertex positions $\residualpos$ based on the vertex positions $\templatepos$ of a template \yl{mesh} with a fixed topology, where $N_{mesh}$ is the number of mesh vertices. 
Then the resulting vertex position $\pos$ of the base mesh is computed as $\pos=R_{head}(\templatepos+\residualpos)+t_{head}$.

Following \cite{lombardi2021mixture}, three decoders $\transformdecoder$, $\opacitydecoder$, and ${\colordecoder}$ with 2D convolutional architectures predict the volumetric primitives upon the base mesh.
Specifically, the transformation decoder $\transformdecoder:\mathbb{R}^{256} \rightarrow \mathbb{R}^{9 \times N_{prim}}$ computes the rotation \yht{$R_{p}$}, translation \yht{$t_{p}$}, and scale \yht{$s_{p}$} of $N_{prim}$ primitives relative to the tangent space of the base mesh, which compensate for the motion that is not modeled by the mesh vertex $\vertex$.
The opacity decoder $\opacitydecoder:\mathbb{R}^{256} \rightarrow \mathbb{R}^{M^3 \times N_{prim}}$ also takes the expression code $\expressioncode$ as input and decodes the voxel opacity $\voxelopacity$ of the primitives.
The appearance decoder $\colordecoder:\mathbb{R}^{256+356+3} \rightarrow \mathbb{R}^{3 \times M^3 \times N_{prim}}$ takes the expression code $\expressioncode$, the lighting condition $\light$, and the view direction $\direction$ as input and predicts the RGB colors $\voxelcolor$ of the primitives.
The architecture of our relightable appearance decoder is designed to leverage the linear nature of lighting (see Section~\ref{sec:relightable_appearance}).

\paragraph{Output synthesis.}
Given the volumetric primitives, we use a differentiable accumulative ray marching algorithm~\cite{lombardi2021mixture,karras2013fast} to render the output images. Specifically, for a ray $\mathbf{r}_\pixel(t)=\mathbf{o}_\pixel + t \mathbf{d}_\pixel$ with a direction $\mathbf{d}_\pixel$ starting from a pixel $\pixel$ with a 3D position $\mathbf{o}_\pixel$, we compute the pixel color $\renderedimage(\pixel)$ as:
\begin{equation}
\begin{aligned}
\renderedimage(\pixel) = \int_{t_{\min}}^{t_{\max}}\voxelcolor(\mathbf{r}_\pixel(t))\frac{dT(\pixel, t)}{dt},
\end{aligned}
\end{equation}
\begin{equation}
\begin{aligned}
T(\pixel, t)=\min\Big(1, \int_{t_{\min}}^{t}\voxelopacity\big(\mathbf{r}_p(t)\big)\Big),
\end{aligned}
\end{equation}
where $t_{\min}$ and $t_{\max}$ are the predefined near and far bounds of \yl{the} rendering range. The opacity of a pixel $\pixel$ is set as $\opacityvalue(\pixel) = T(\pixel, t_{\max})$.

\subsection{Relightable Appearance}
\label{sec:relightable_appearance}

In this section, we detail our specially designed appearance decoder $\colordecoder$ that enables high-fidelity real-time relighting using environment maps.
Although the appearance changes drastically when lighting condition changes, previous methods~\cite{basri2003lambertian, xu2018deep} have shown that the relighted images often lie in low-dimensional subspaces. For example, nearly all the lighting effects are linear\delete{ly additive}~\cite{debevec2000acquiring, chandrasekhar2013radiative} and the full reflectance field can be predicted from \yl{a} few images of the object in specific lighting conditions~\cite{xu2018deep, meka2019deep}. However, directly predicting all OLAT images and adding them up for environment map relighting is not feasible for real-time rendering.
Our key observation is that we can design a network architecture upon the disentangled representation for our appearance decoder $\colordecoder$ to strictly satisfy the linear\delete{ly additive} nature of lighting, \ie:
\begin{equation}
\begin{aligned}
\colordecoder(\expressioncode, k_1\light_1 + k_2\light_2, \direction) &= k_1\colordecoder(\expressioncode, \light_1, \direction) \\& + k_2\colordecoder(\expressioncode, \light_2, \direction), \forall k_1 \,\mathrm{and}\, \  k_2 \in\mathbb{R}.
\end{aligned}
\end{equation}
We show the architecture of $\colordecoder$ in Figure~\ref{fig:method}. Considering the spatially structured effect for each light, we use a convolutional architecture for $\colordecoder$.
The expression code $\expressioncode$ and the view direction $\direction$ are fed into an ordinary \emph{non-linear} branch.
The lighting condition $\light$ is injected in a separate \emph{linear} branch, where the activation layers and the bias in the fully connected layer and transposed convolutional layers are removed. The feature maps of the linear branch $\linearfeaturemap$ is point-wise multiplied with the feature maps from the non-linear branch $\nonlinearfeaturemap$ at each stage:
\begin{equation}
\begin{aligned}
\linearfeaturemap^{i+1}=ConvT\big(\linearfeaturemap^{i} \odot (\nonlinearfeaturemap^{i} + 1)\big),
\end{aligned}
\end{equation}
where $i$ is the index of the stage, $ConvT$ represents the transposed convolution operation, and $\odot$ is point-wise multiplication. The plus one term acts as a residual connection that stabilizes training (this term is omitted in Figure~\ref{fig:method} to avoid clutter). In this way, the appearance decoder $\colordecoder$ is strictly linear to the lighting condition $\light$ while being non-linear to the expression code $\expressioncode$ and the view direction $\direction$ that does not limit the representation power.
We empirically find that our architecture significantly improves the generalization ability for novel lighting conditions (see Section~\ref{sec:ablation_study} for some related evaluation results).

\subsection{Data Acquisition}
\label{sec:data_acquisition}

Capturing each transient facial expression under a variety of lighting conditions for relightable appearance poses a significant challenge.
Instead, for each subject, we record image sequences of dynamic expressions with different lighting conditions in each frame and rely on our self-supervised 
training framework for disentanglement by using information across frames. Following \cite{bi2021deep, li2023megane}, we use group light patterns for capture, \ie, for each frame seven randomly selected adjacent lights are turned to the maximum. 
Differently, since we \yl{do not} use interleaved full-on frames for tracking, we find a large part of the face is dark in group light conditions that makes the implicit tracking in our network unstable. To provide basic illumination, we set all lights not included in the selected group to a known low brightness. Thanks to the linear nature of light and our network architecture design, the fully disentangled relightable appearance can be learned from such coalescent lighting conditions.

During the capture process, a subject is asked to perform 41 predefined expressions and read out two paragraphs. Then a freestyle performance is captured to cover extreme and complex expression combinations. We capture 10200 frames for each subject at 20fps. We show a snapshot of our captured images in Figure~\ref{fig:hardware}. The background without the subject is also captured.


\subsection{Network Training}
\label{sec:network_training}

Our model is trained end-to-end on the multi-view image sequences under varying illuminations. The training loss $\totalloss$ consists of two parts: $\totalloss=\dataloss+\regularizationloss$, where $\dataloss$ is the data term and $\regularizationloss$ is the regularization term.

The data term $\dataloss$ contains three components and measures the similarity between the captured input and the rendered output:
\begin{equation}
\begin{aligned}
\dataloss=\loss_{1}+{\lambda}_\mathrm{VGG}\loss_\mathrm{VGG}+{\lambda}_\mathrm{GAN}\loss_\mathrm{GAN},
\end{aligned}
\end{equation}
where $\loss_{1}$ is the MAE loss, $\loss_\mathrm{VGG}$ is the perceptual loss, and $\loss_\mathrm{GAN}$ is the adversarial loss that improves the visual quality. ${\lambda}_\mathrm{VGG}$ and ${\lambda}_\mathrm{GAN}$ are the balancing weights. We clip the pixel values of the rendered images $\renderedimage$ before calculating loss to simulate the truncation of the imaging process.

The regularization loss $\regularizationloss$ comprises four components:
\begin{equation}
\begin{aligned}
\regularizationloss={\lambda}_\mathrm{Lap}\loss_\mathrm{Lap}+{\lambda}_{pR}\loss_{pR}+{\lambda}_{vol}\loss_{vol}+{\lambda}_\mathrm{KLD}\loss_\mathrm{KLD},
\end{aligned}
\end{equation}
where $\loss_\mathrm{Lap}=||\laplacian(\vertex - \vertex_{base})||^2$ is the expression-aware Laplacian loss to encourage a smooth base mesh. $\laplacian$ is the sparse Laplacian matrix. 
$\vertex_{base} = \blendshapes(\blendshapes^T \blendshapes)^{-1}\blendshapes^T\vertex$ is calculated in a least-squares manner based on the 51 predefined expression blendshapes $\blendshapes \in \mathbb{R}^{51 \times 3N_{mesh}}$ from the FaceScape dataset~\cite{yang2020facescape}.
$\loss_{pR} =\frac{1}{N_{prim}}||(\transformdecoder)_{R,t}||$ regularizes the predicted rotation and translation $(\transformdecoder)_{R,t}$ to be small.
We apply a predefined mask on the base mesh to assign higher weights of $\loss_\mathrm{Lap}$ and $\loss_{pR}$ on facial regions \yl{compared} to non-facial parts.
$\loss_{vol}$ and $\loss_\mathrm{KLD}$ are the volume minimization prior and KL-divergence
loss as in \cite{lombardi2021mixture}, respectively.
${\lambda}_\mathrm{Lap}$, ${\lambda}_{pR}$, ${\lambda}_{vol}$, and ${\lambda}_\mathrm{KLD}$ are balancing weights.

Since our training images are captured under varying illuminations, the background changes across frames.
To prevent the encoding of background flashes into the avatar, the final image $\finalimage$ in training is generated by blending the rendered foreground $\foregroundimage$ with \delete{and }the captured background $\backgroundimage$ based on the computed opacity value $\opacityvalue$:
\begin{equation}
\begin{aligned}
\finalimage=\opacityvalue \foregroundimage +(1-\opacityvalue)\backgroundimage.
\end{aligned}
\end{equation}

We use the Adam optimizer~\cite{kingma2015adam} to train the network with a learning rate of ${10}^{-4}$. We choose frontal, left, and right views as input of the encoder.
The input images are normalized and converted to grayscale to prevent the light from being encoded in the expression code $\expressioncode$.
We use the per-camera color calibration similar to \cite{lombardi2021mixture}.
For monochrome cameras, the rendered images are explicitly converted to grayscale before calculating loss functions. We fit a base mesh on the first frame for initialization.

The network training for each subject takes about two days on a single NVIDIA V100 graphics card.
The decoding and rendering take around 22ms for a frame of a resolution $1280\times960$, enabling real-time relighting and animation.
Please refer to our supplementary materials for implementation details such as network architectures and hyperparameters.

\section{Experiments}

\subsection{Qualitative Evaluation Results}

\paragraph{Mesh-volume representation}
Figure~\ref{fig:topology} shows two examples of our avatars based on the hybrid mesh-volume representation.
Although our avatars are trained without explicit tracking, the base mesh and the volumetric primitives are roughly aligned. The inherently consistent structures enable explicit control and can be naturally used for applications such as video-driven animations and relighting.

\paragraph{Disentanglement of illumination and motion}

\begin{figure*}
    \centering
    \begin{minipage}{\linewidth}
        \centering
        \hspace{\onesixthfigurewidth\linewidth}
        \begin{minipage}{\onesixthfigurewidth\linewidth}
            \centering
            \includegraphics[width=\linewidth]{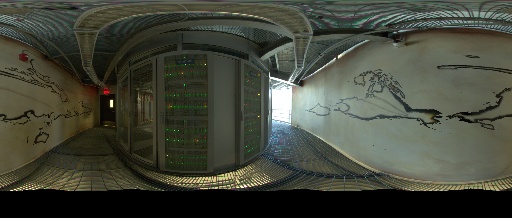}
        \end{minipage}
        \begin{minipage}{\onesixthfigurewidth\linewidth}
            \centering
            \includegraphics[width=\linewidth]{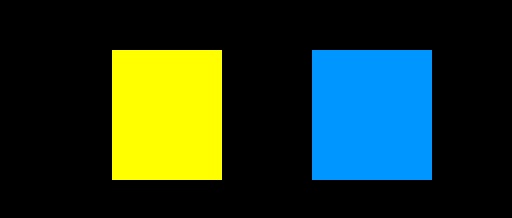}
        \end{minipage}
        \hspace{\onesixthfigurewidth\linewidth}
        \begin{minipage}{\onesixthfigurewidth\linewidth}
            \centering
            \includegraphics[width=\linewidth]{images/disentanglement/lightingA.jpg}
        \end{minipage}
        \begin{minipage}{\onesixthfigurewidth\linewidth}
            \centering
            \includegraphics[width=\linewidth]{images/disentanglement/lightingB.jpg}
        \end{minipage}
    \end{minipage}

    \begin{minipage}[t]{\onesixthfigurewidth\linewidth}
        \centering
        \includegraphics[width=\linewidth]{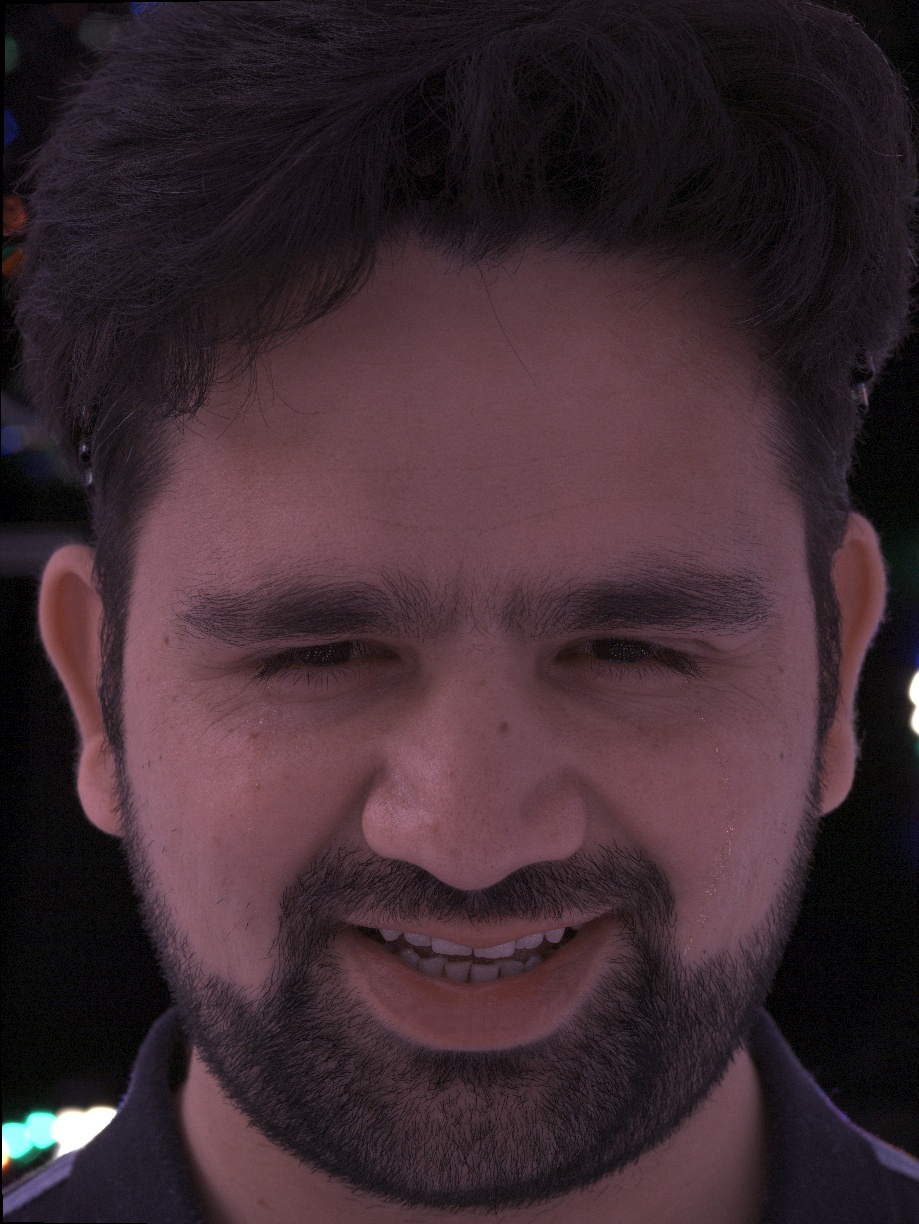}
        \includegraphics[width=\linewidth]{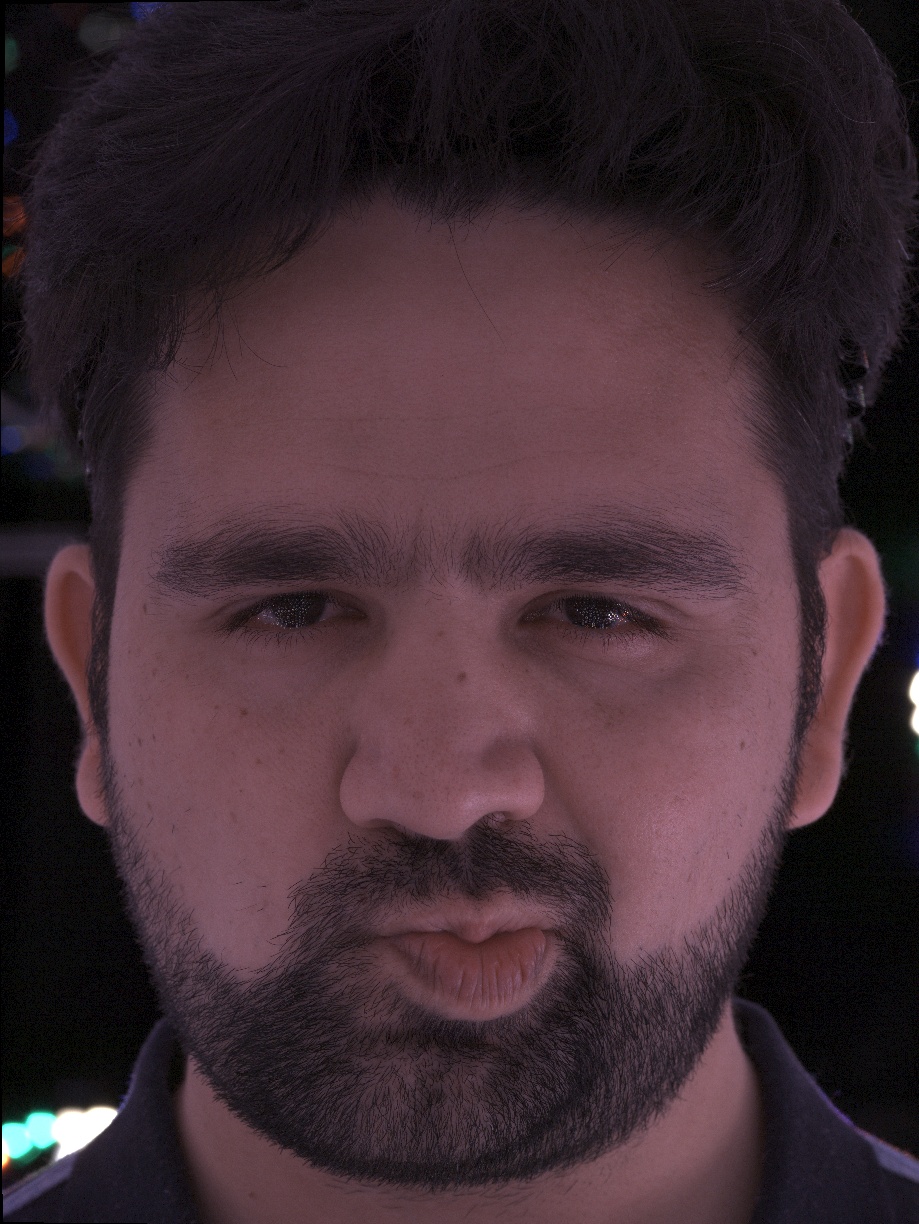}
        \subfloat{Subject D}
        \label{fig:disentanglement_sub1}
    \end{minipage}
    \begin{minipage}[t]{\onesixthfigurewidth\linewidth}
        \centering
        \includegraphics[width=\linewidth]{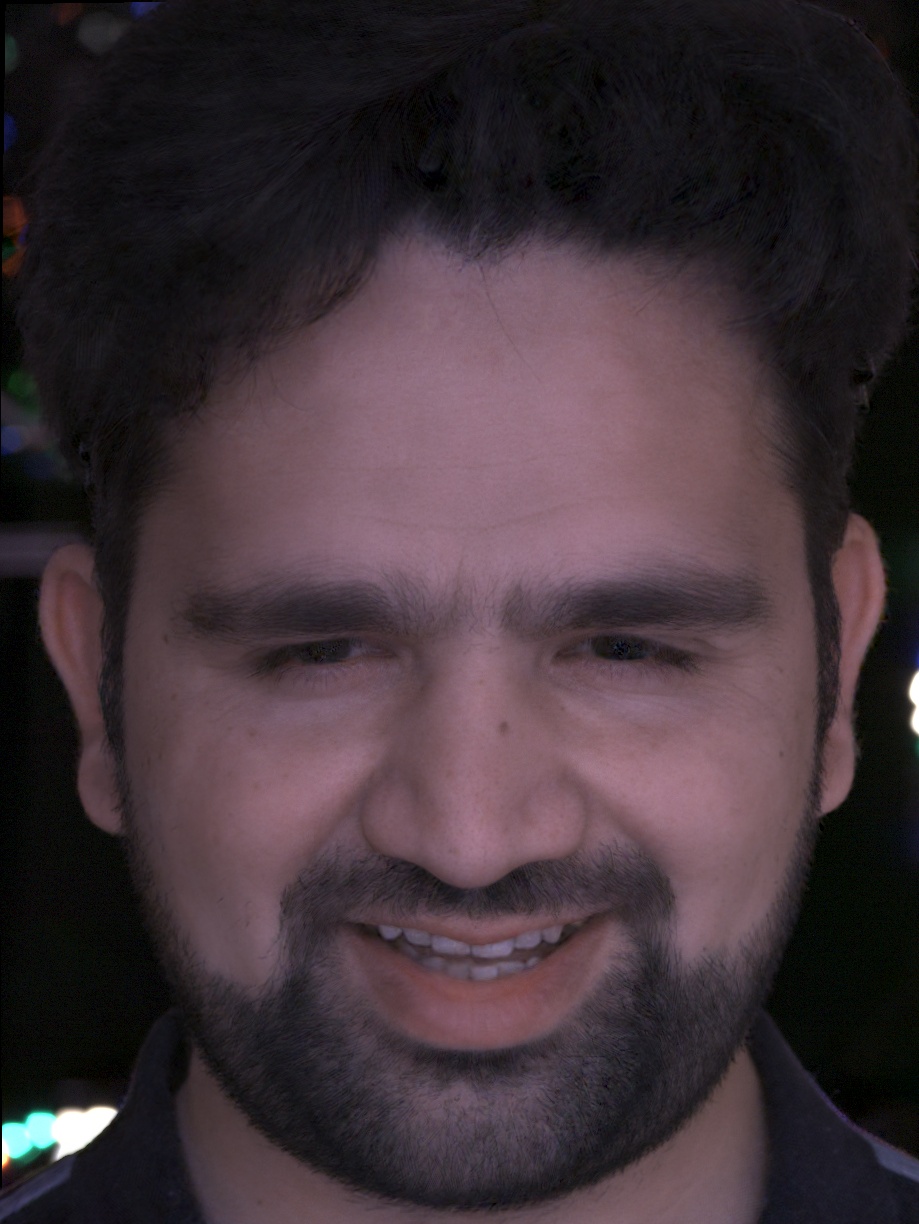}
        \includegraphics[width=\linewidth]{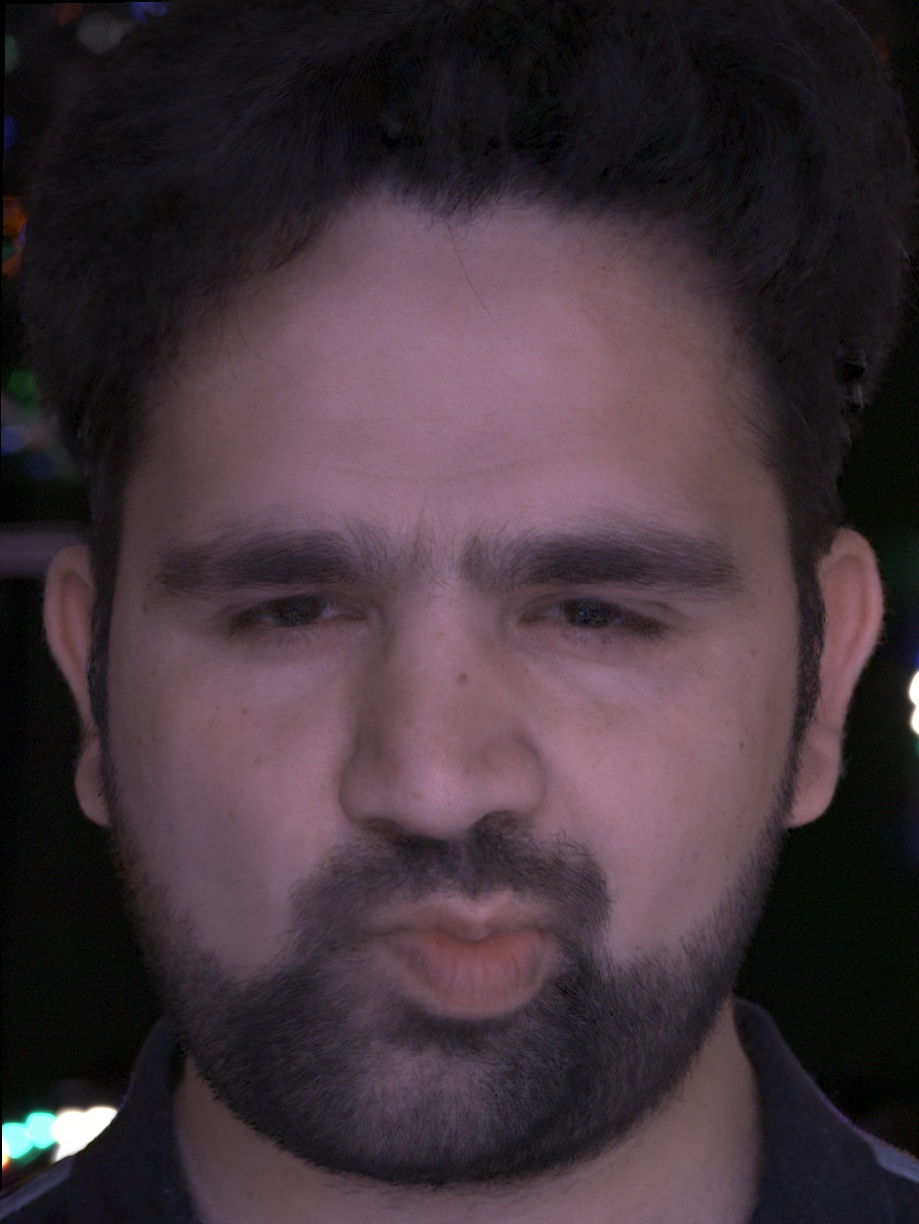}
        \subfloat{Lighting 1}
        \label{fig:disentanglement_sub1_lightA}
    \end{minipage}
    \begin{minipage}[t]{\onesixthfigurewidth\linewidth}
        \centering
        \includegraphics[width=\linewidth]{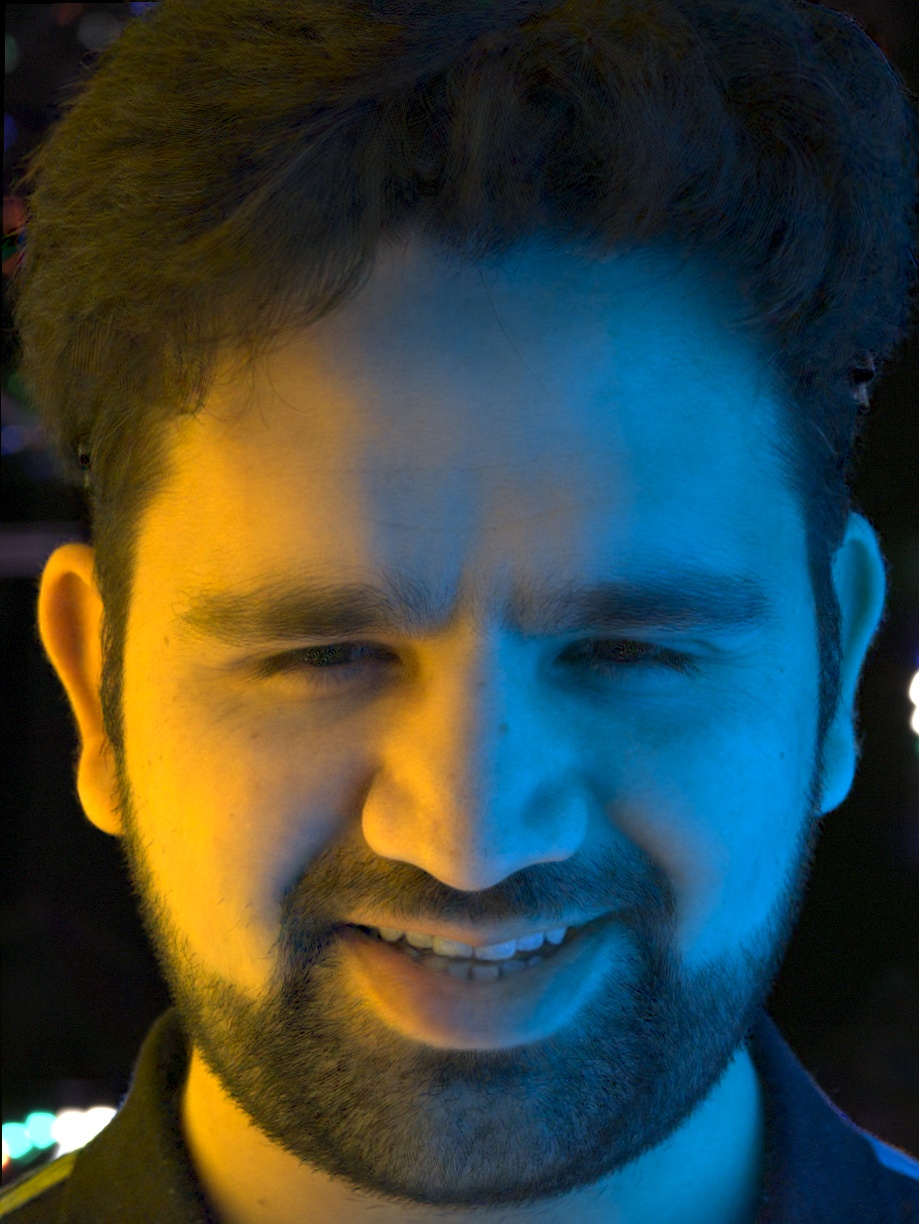}
        \includegraphics[width=\linewidth]{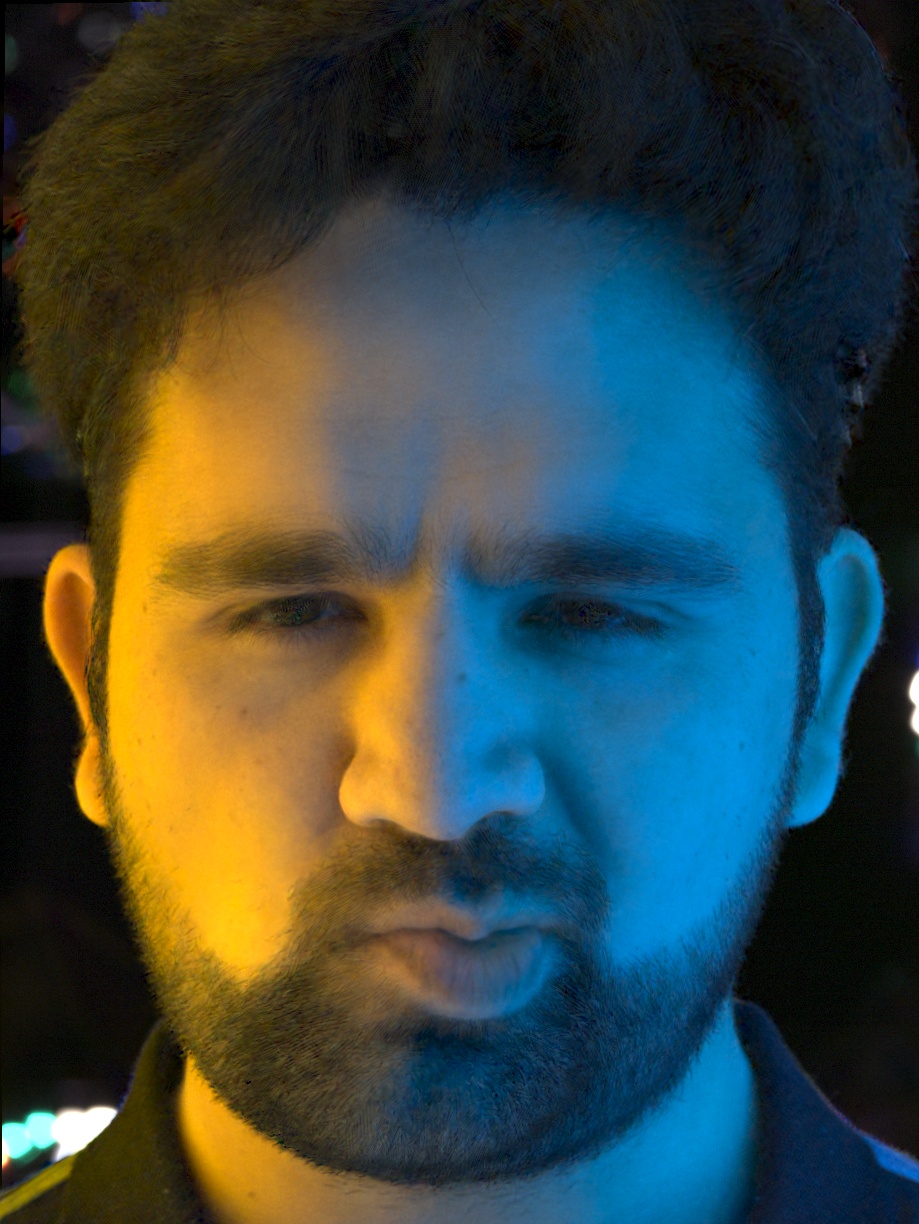}
        \subfloat{Lighting 2}
        \label{fig:disentanglement_sub1_lightB}
    \end{minipage}
    \begin{minipage}[t]{\onesixthfigurewidth\linewidth}
        \centering
        \includegraphics[width=\linewidth]{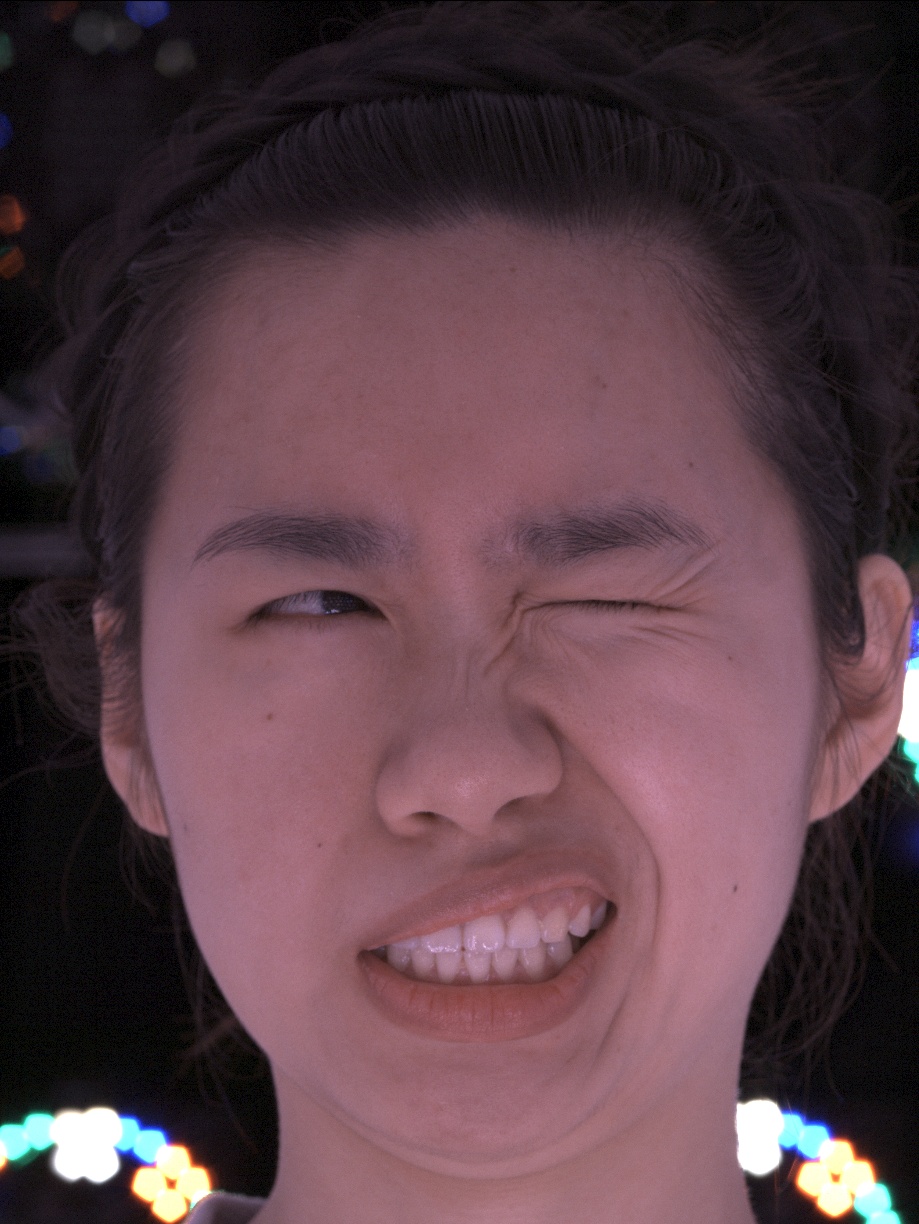}
        \includegraphics[width=\linewidth]{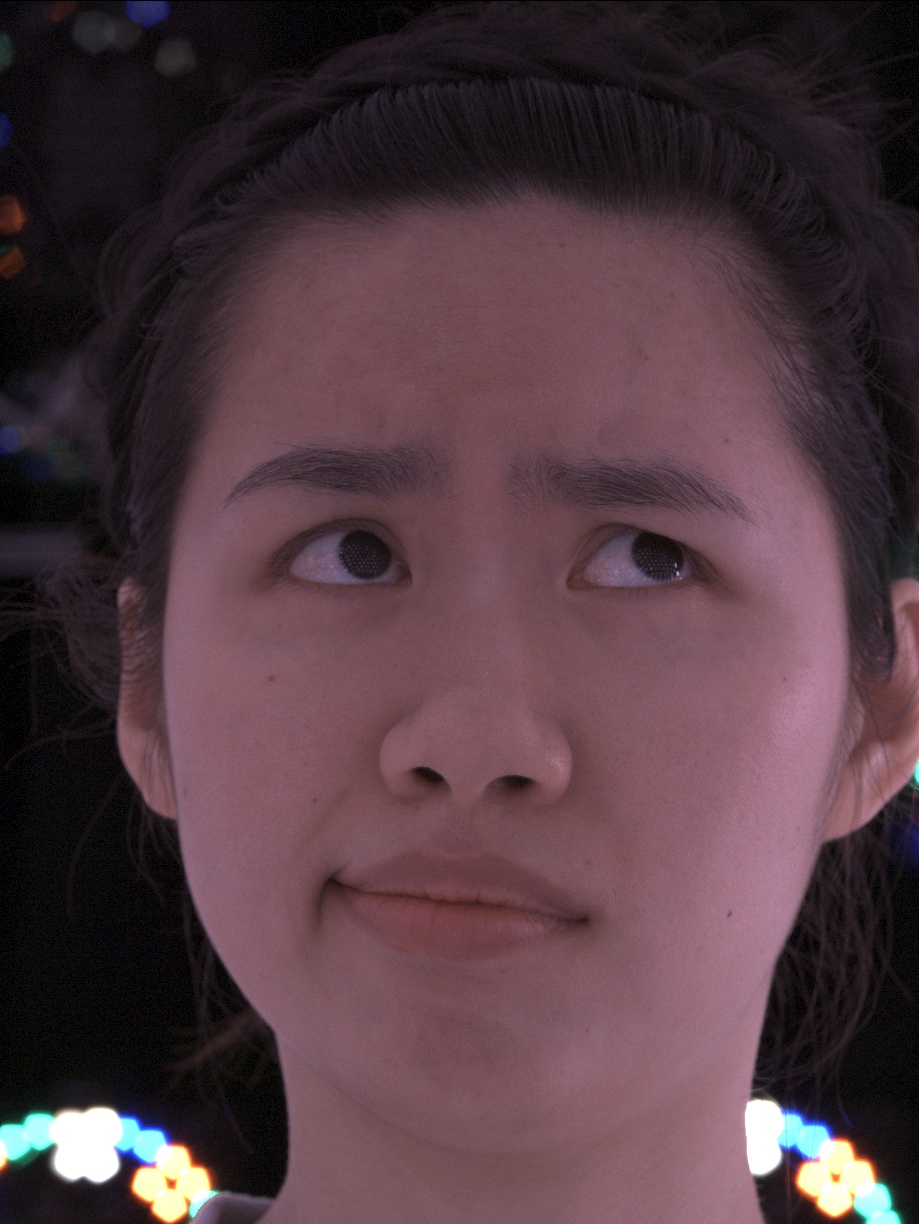}
        \subfloat{Subject E}
        \label{fig:disentanglement_sub2}
    \end{minipage}
    \begin{minipage}[t]{\onesixthfigurewidth\linewidth}
        \centering
        \includegraphics[width=\linewidth]{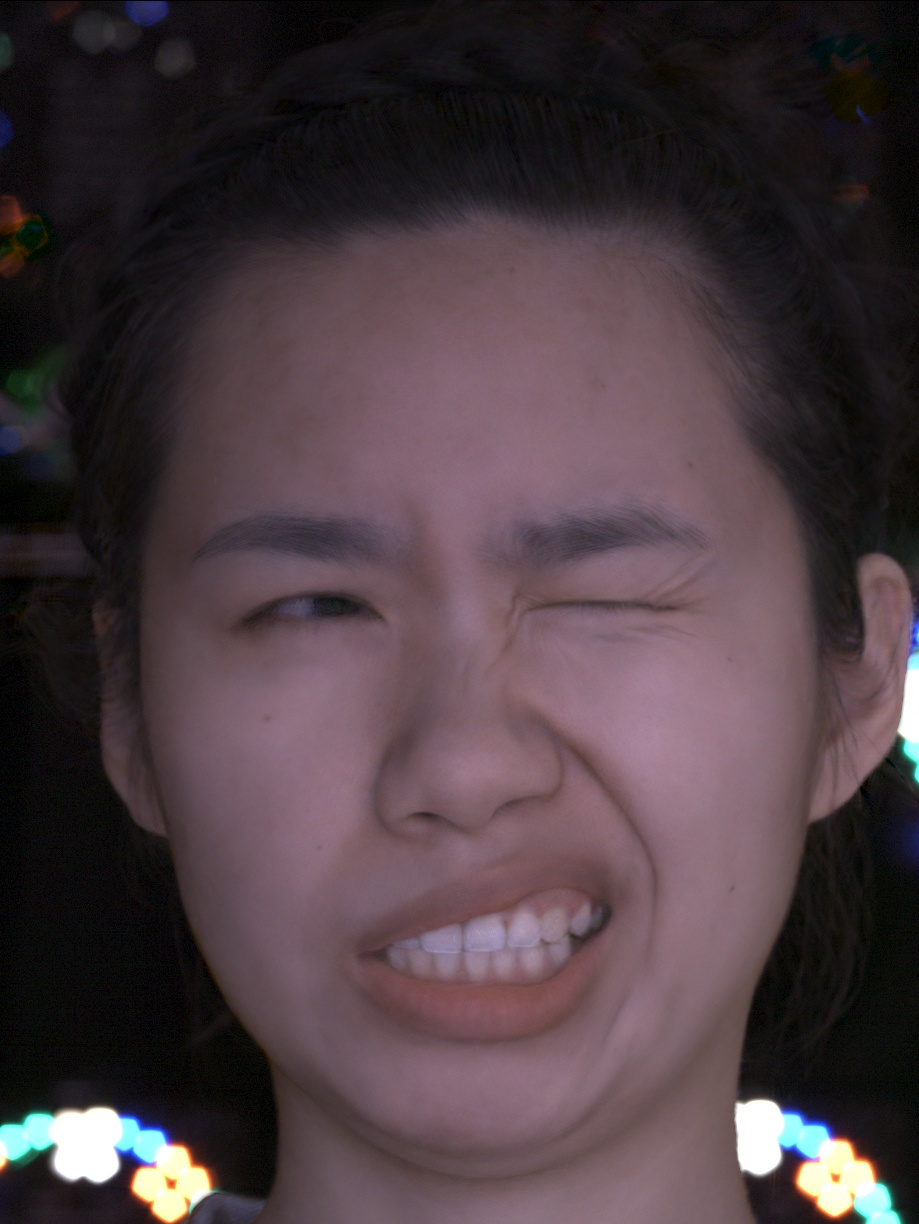}
        \includegraphics[width=\linewidth]{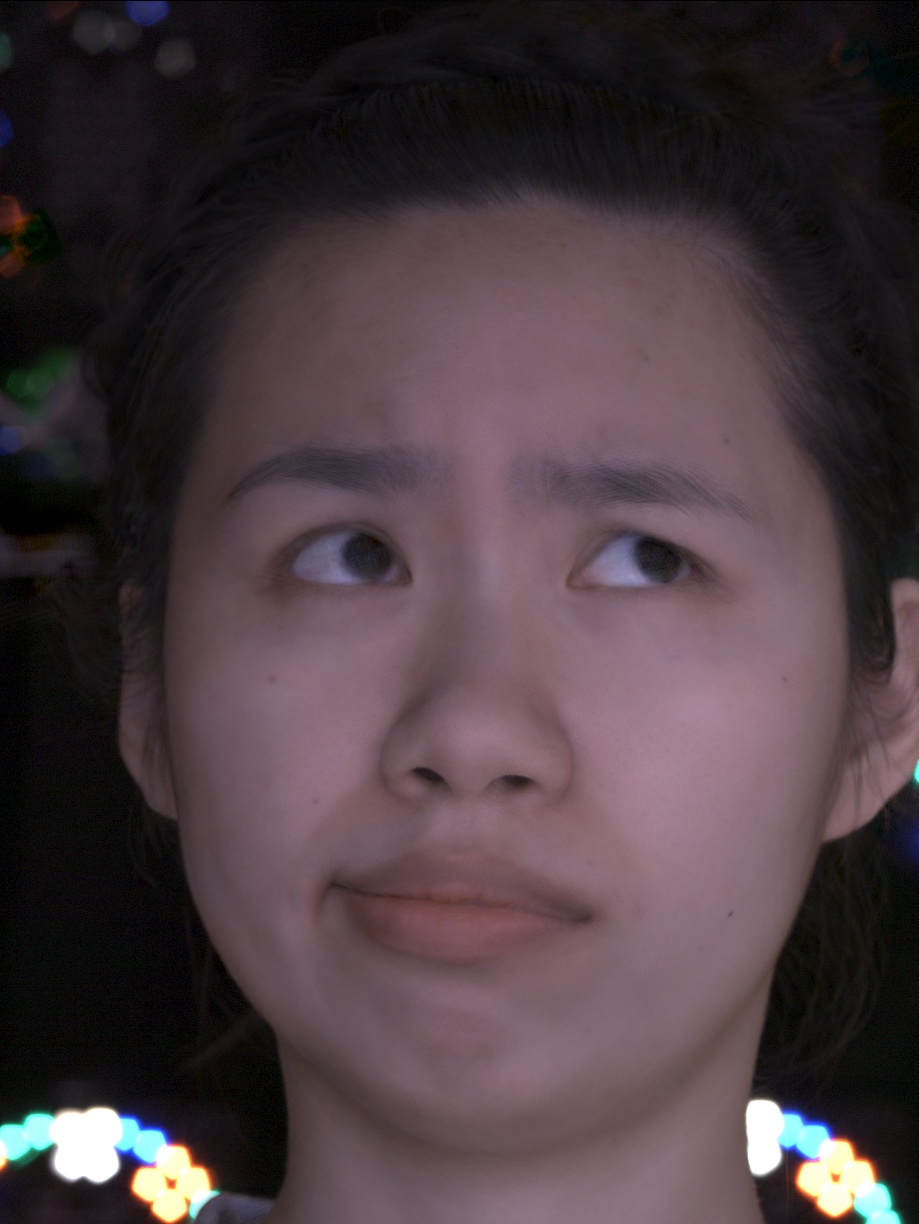}
        \subfloat{Lighting 1}
        \label{fig:disentanglement_sub2_lightA}
    \end{minipage}
    \begin{minipage}[t]{\onesixthfigurewidth\linewidth}
        \centering
        \includegraphics[width=\linewidth]{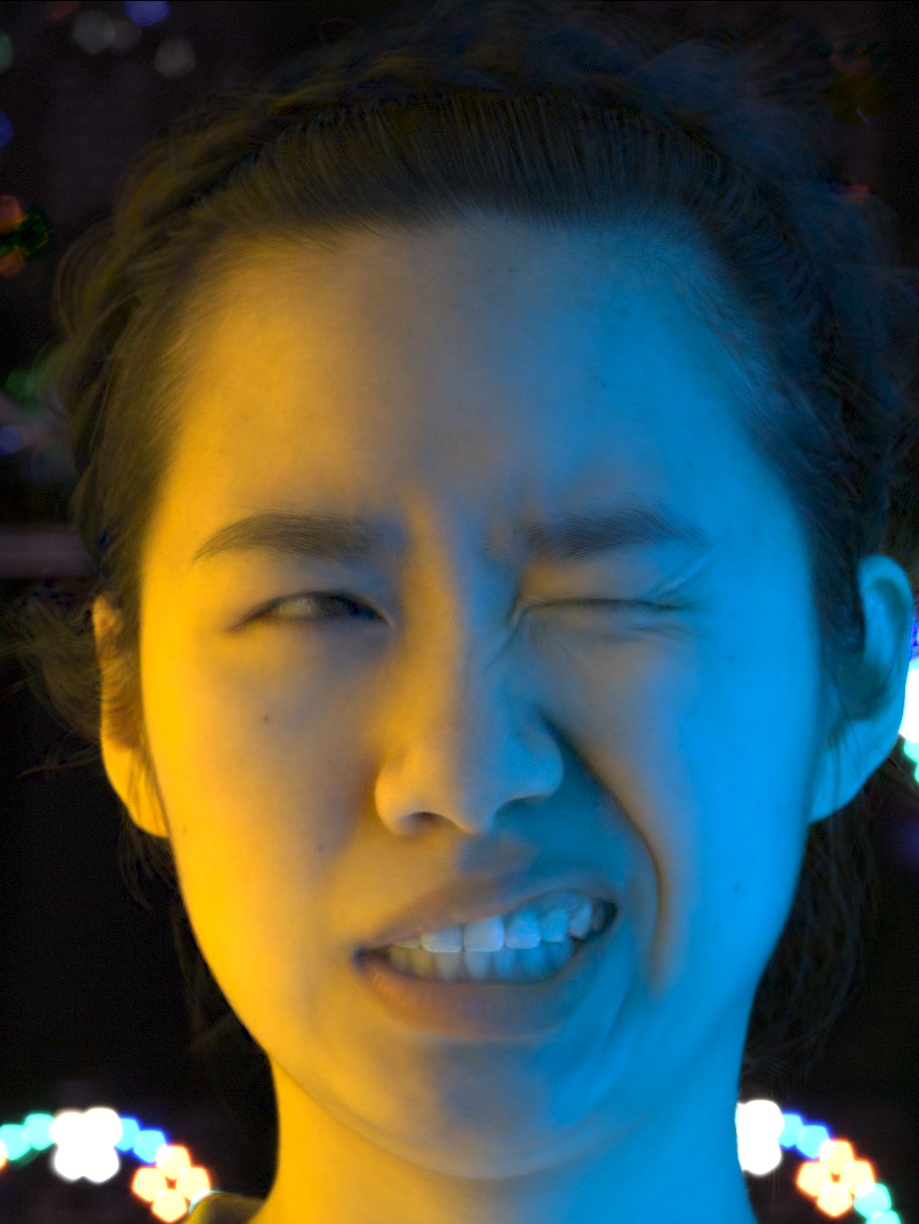}
        \includegraphics[width=\linewidth]{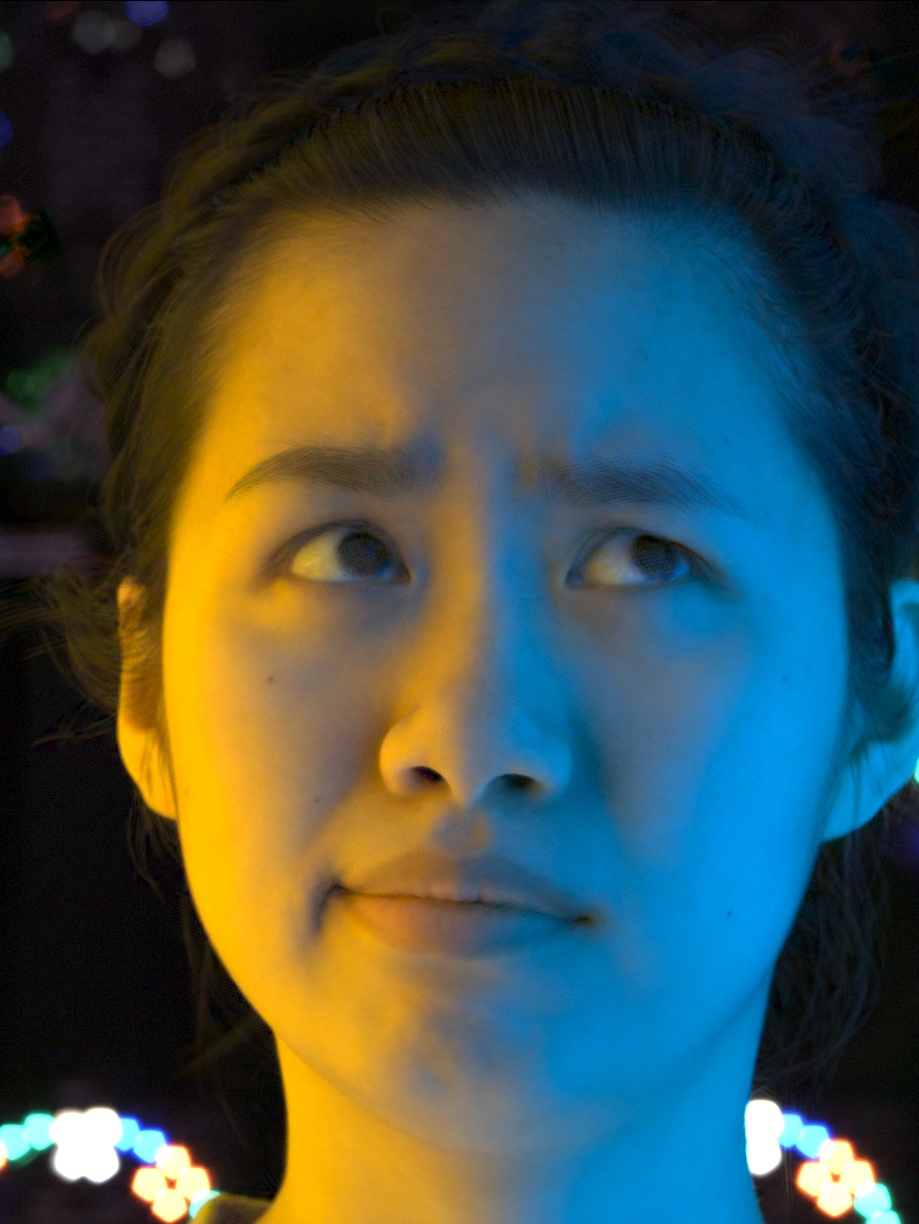}
        \subfloat{Lighting 2}
        \label{fig:disentanglement_sub2_lightB}
    \end{minipage}
    \caption{Evaluation results of lighting and motion disentanglement.
    For both subjects, we show the input frames of two different expressions on the left and the corresponding relighting results in the middle and on the right.
    The two input environment maps for relighting are shown on the top. The relighting effects are consistent with the dynamic expressions.
    }
\label{fig:disentanglement}
\end{figure*}

Both illumination and motion are varied in our captured sequences.
To evaluate the disentanglement of illumination and motion in our model, for each input frame, we keep the extracted expression code $\expressioncode$ fixed and change the lighting condition $\light$ extracted from environment maps to generate the relighting results. We use the appearance decoder to predict the relighted appearance of RGB channels separately for colorful environment map relighting.
As shown in Figure~\ref{fig:disentanglement}, the lighting conditions are fully disentangled from the motion and are consistent across different subjects.

\subsection{Comparisons to Prior Work}

\paragraph{Comparison to MVP}

\begin{figure}
    \centering
    \begin{minipage}[t]{\onefourthfigurewidth\linewidth}
        \centering
        \includegraphics[width=\linewidth]{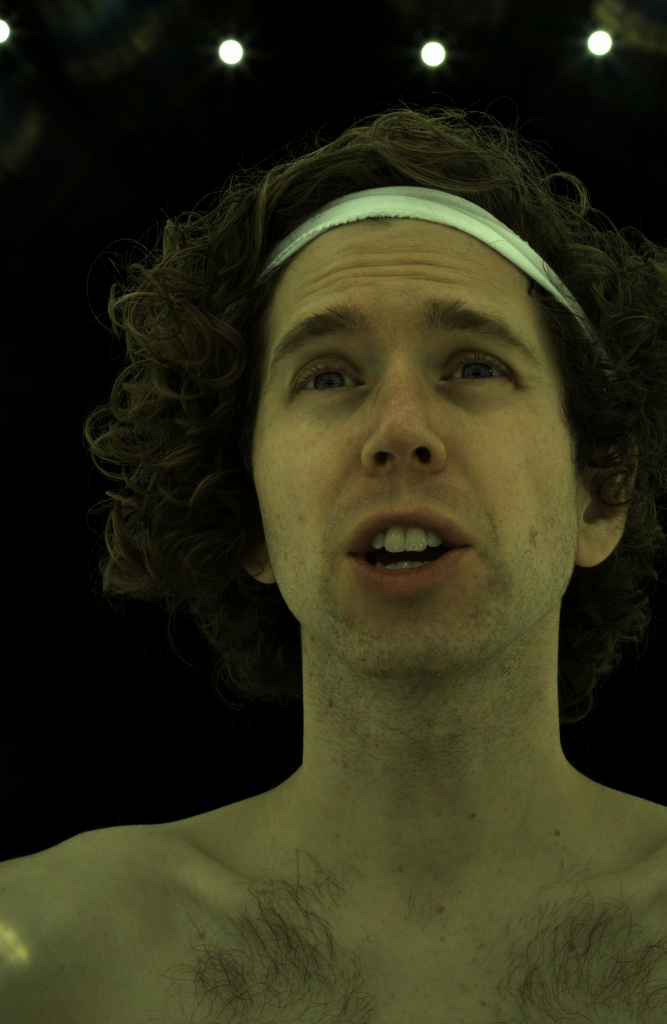}
        \includegraphics[width=\linewidth]{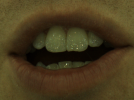}
        \subfloat{Ground-truth}
    \end{minipage}
    \begin{minipage}[t]{\onefourthfigurewidth\linewidth}
        \centering
        \includegraphics[width=\linewidth]{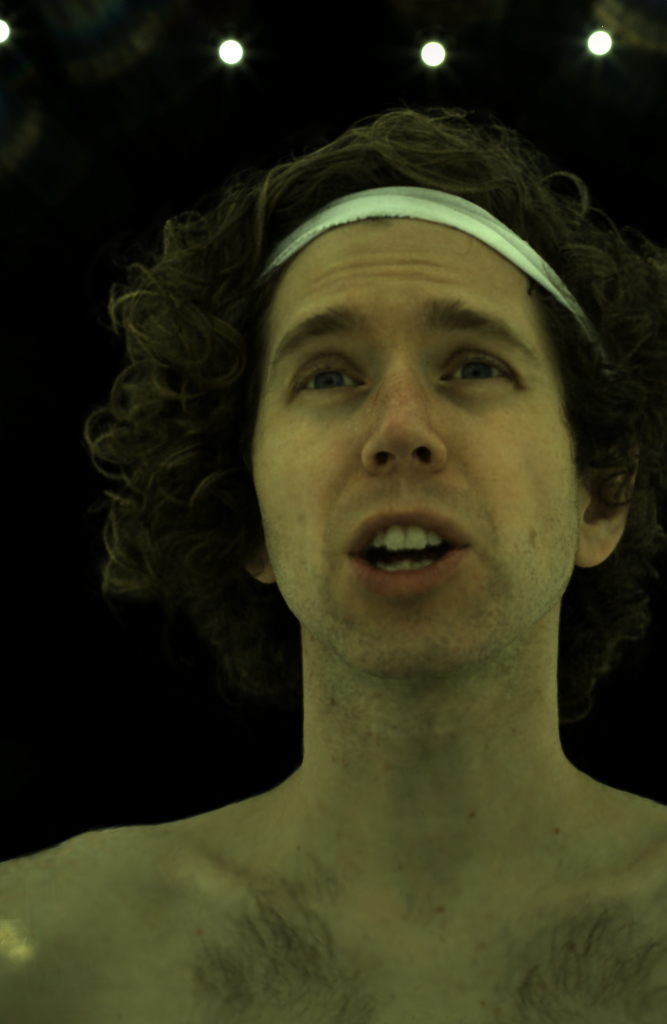}
        \includegraphics[width=\linewidth]{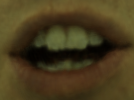}
        \subfloat{MVP}
    \end{minipage}
    \begin{minipage}[t]{\onefourthfigurewidth\linewidth}
        \centering
        \includegraphics[width=\linewidth]{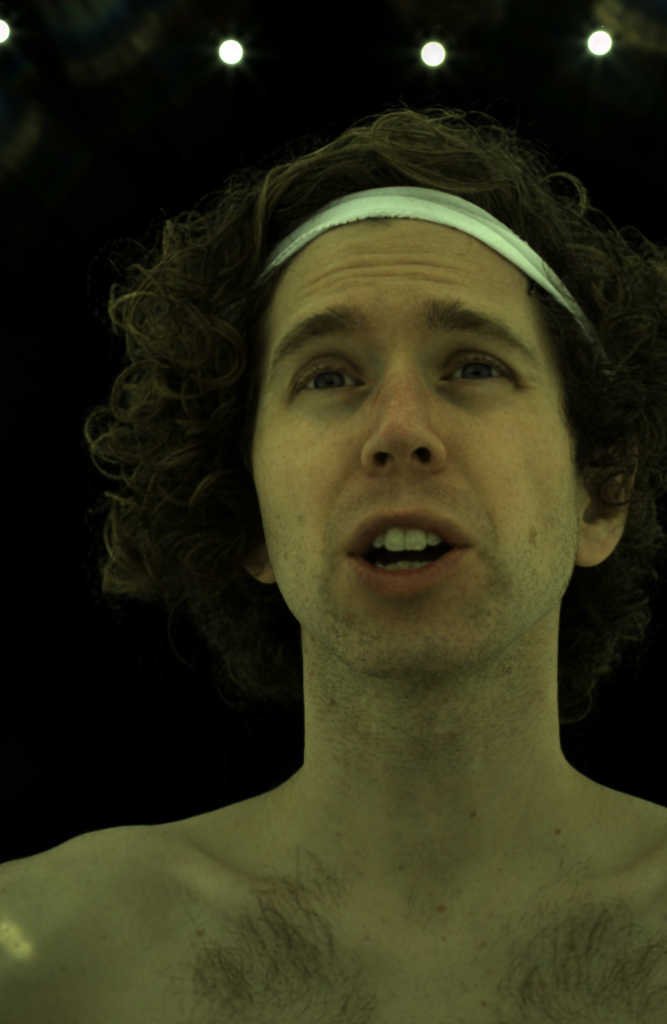}
        \includegraphics[width=\linewidth]{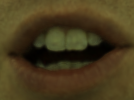}
        \subfloat{MVP+}
    \end{minipage}
    \begin{minipage}[t]{\onefourthfigurewidth\linewidth}
        \centering
        \includegraphics[width=\linewidth]{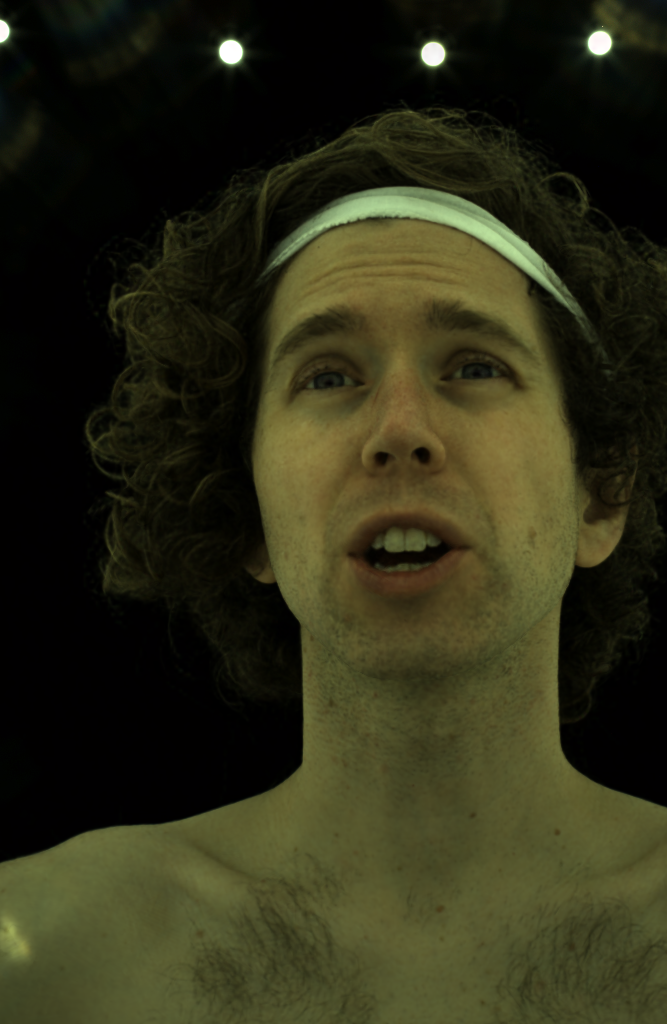}
        \includegraphics[width=\linewidth]{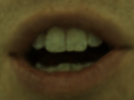}
        \subfloat{Ours}
    \end{minipage}
    \caption{Comparison to MVP~\cite{lombardi2021mixture} on novel view synthesis. Our results are comparable to MVP and MVP+ (an improved version of MVP trained by ourselves) even without explicit tracking of the \yl{base} mesh.}
\label{fig:cmp_mvp}
\end{figure}

\begin{table}
\center 
\caption{Quantitative evaluation results of novel view synthesis in comparison with MVP~\cite{lombardi2021mixture}. The two subjects are from the Multiface Dataset~\cite{wuu2022multiface}.}
\resizebox{\columnwidth}{!}{%
\begin{tabular}{l|ccc|ccc}
\hline
       & \multicolumn{3}{c|}{Subject \#002421669} & \multicolumn{3}{c}{Subject \#5067077} \\ \hline
Method & MAE $\downarrow$ & SSIM $\uparrow$ & LPIPS $\downarrow$ & MAE $\downarrow$ & SSIM $\uparrow$ & LPIPS $\downarrow$\\ \hline
MVP    & 2.08          & 0.910   & 0.273      & 2.21         & 0.923    & 0.232     \\
MVP+   & 1.76          & 0.930   & 0.193      & 2.11         & 0.928    & 0.211     \\
Ours   & \textbf{1.73} & \textbf{0.932} & \textbf{0.186} & \textbf{2.01}& \textbf{0.934}   & \textbf{0.208}      \\ \hline
\end{tabular}
\label{tab:cmp_mvp}%
}
\end{table}

Since existing explicit surface tracking methods~\cite{beeler2011high, wu2018deep} do not generalize well under varying lighting conditions, we compare to MVP~\cite{lombardi2021mixture} on the publicly available Multiface Dataset~\cite{wuu2022multiface}, which consists of high quality multi-view recordings of 13 different identities under fixed illumination.
We perform qualitative and quantitative evaluations on eight held out views of two subjects. 
The vanilla MVP uses an L2 loss during training, which leads to blurry results.
We train an improved version, namely MVP+, using the similar data term as ours for fair comparison. The other components remain \yl{identical} to the vanilla MVP.

The visual comparison on Subject \#002421669 from the dataset is shown in Figure~\ref{fig:cmp_mvp}.
The Mean Absolute Error (MAE), Structural Similarity Index (SSIM), and Learned Perceptual Image Patch Similarity (LPIPS) measurements are reported in Table~\ref{tab:cmp_mvp}.
Both our method and MVP+ generate \yl{clearer} details \yl{compared} to vanilla MVP.
Even without a computationally intensive tracking process, the quantitative reconstruction error of our method is slightly lower than that of MVP+.
We attribute the improvement to the avoidance of information loss in the explicit surface tracking process.

\ifthenelse{\equal{\final}{1}}
{
\newcommand{\dprsubfigwidth}{0.25}
\newcommand{\dprsubfigvspace}{-0.084}

\begin{figure}
    \centering
    \begin{minipage}{\linewidth}
        \centering
    \begin{minipage}{\linewidth}
        \includegraphics[width=\onethirdfigurewidth\linewidth]{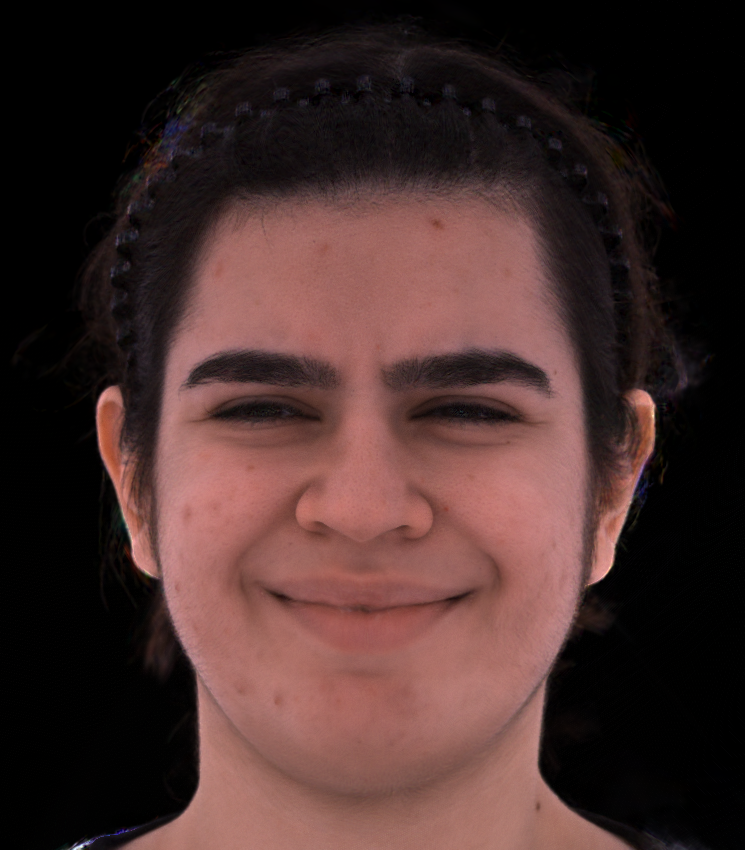}
        \includegraphics[width=\onethirdfigurewidth\linewidth]{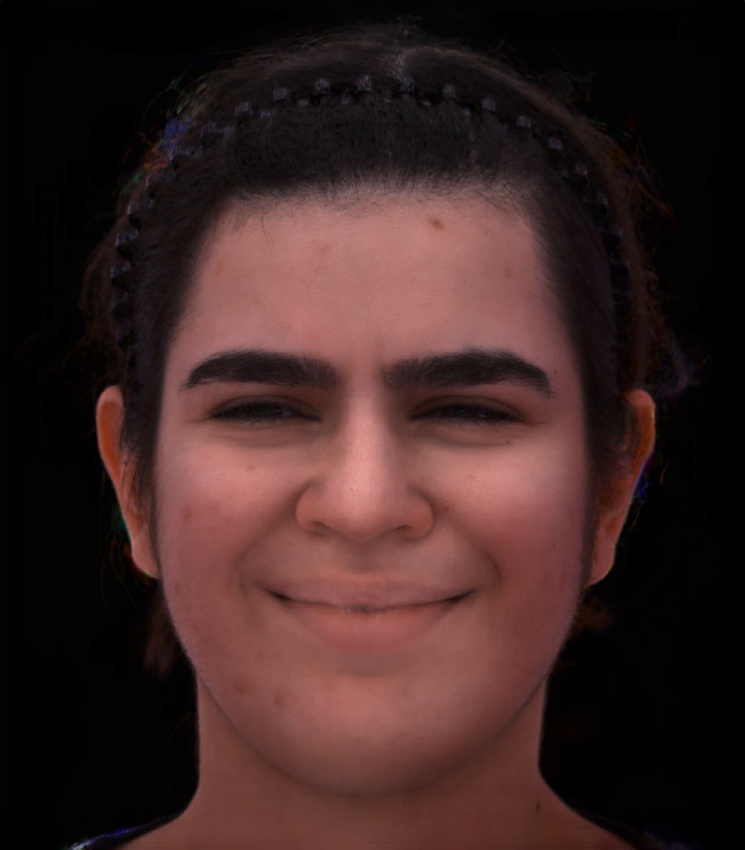}
        \includegraphics[width=\onethirdfigurewidth\linewidth]{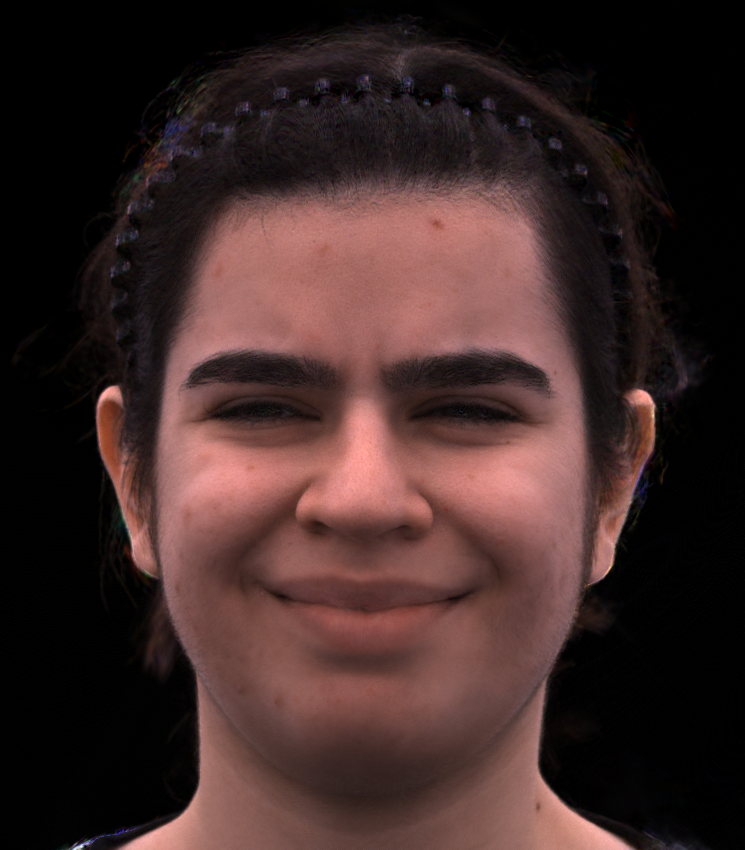}
    \end{minipage}

    \vspace{\dprsubfigvspace\linewidth}
    \begin{minipage}{\linewidth}
        \hspace{\onethirdfigurewidth\linewidth}
        \begin{minipage}{\onethirdfigurewidth\linewidth}
            \includegraphics[width=\dprsubfigwidth\linewidth]{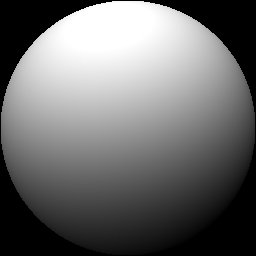}
        \end{minipage}
        \begin{minipage}{\onethirdfigurewidth\linewidth}
            \includegraphics[width=\dprsubfigwidth\linewidth]{images/compare_dpr/light_02.png}
        \end{minipage}
    \end{minipage}

    \begin{minipage}{\linewidth}
        \includegraphics[width=\onethirdfigurewidth\linewidth]{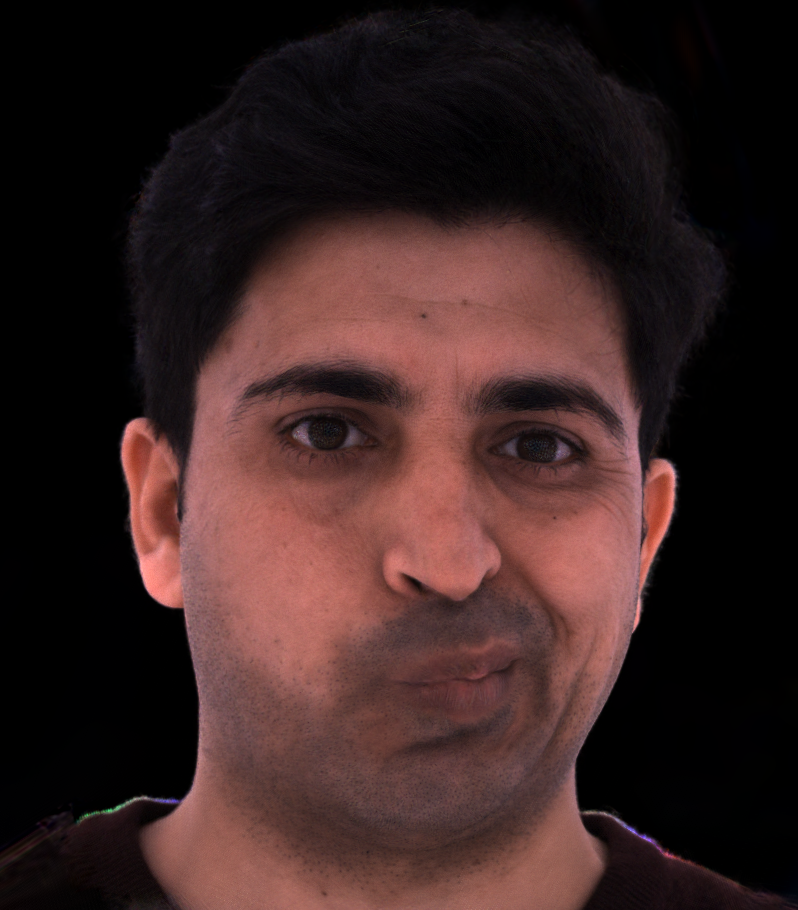}
        \includegraphics[width=\onethirdfigurewidth\linewidth]{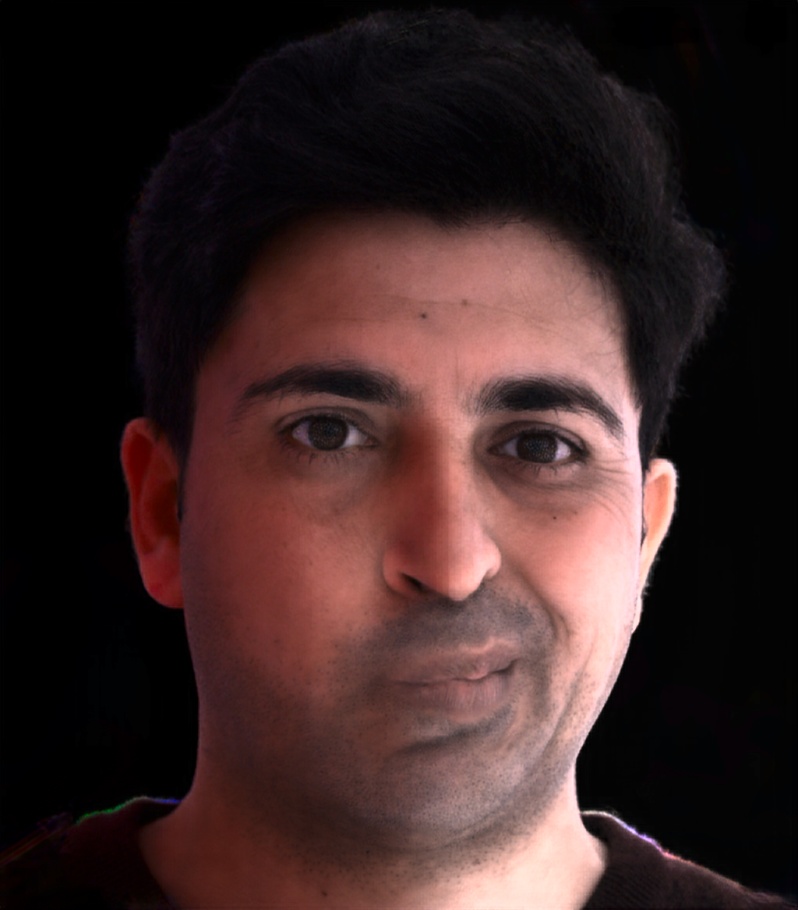}
        \includegraphics[width=\onethirdfigurewidth\linewidth]{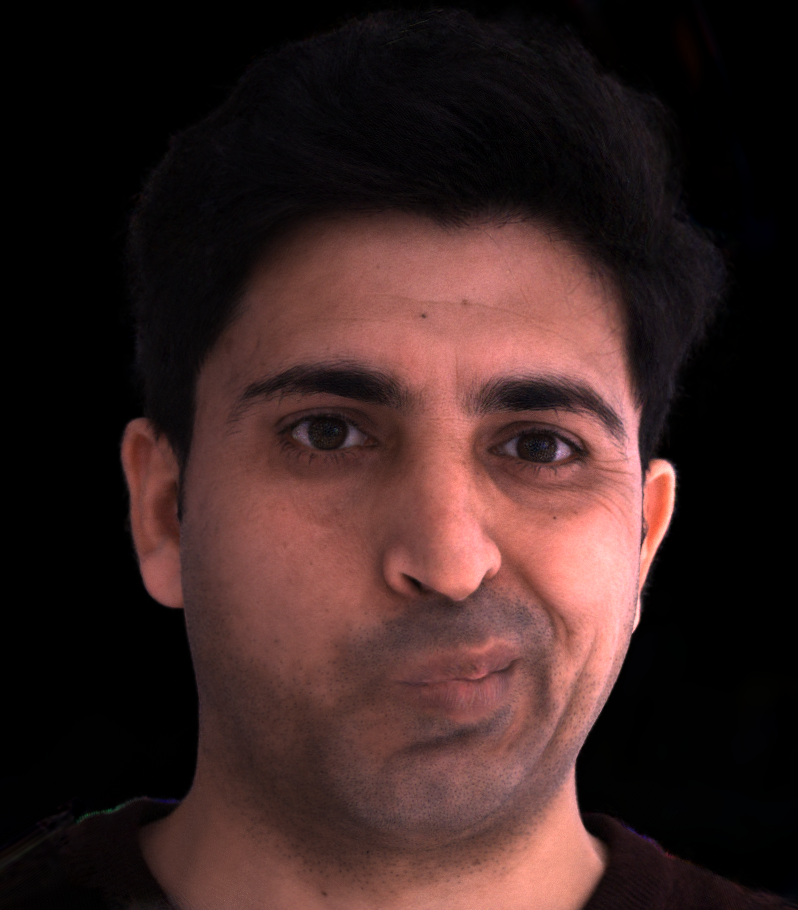}
    \end{minipage}

    \vspace{\dprsubfigvspace\linewidth}
    \begin{minipage}{\linewidth}
        \hspace{\onethirdfigurewidth\linewidth}
        \begin{minipage}{\onethirdfigurewidth\linewidth}
            \includegraphics[width=\dprsubfigwidth\linewidth]{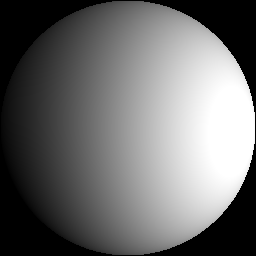}
        \end{minipage}
        \begin{minipage}{\onethirdfigurewidth\linewidth}
            \includegraphics[width=\dprsubfigwidth\linewidth]{images/compare_dpr/light_00.png}
        \end{minipage}
    \end{minipage}

    \vspace{5pt}
    \begin{minipage}{\linewidth}
        \begin{minipage}[t]{\onethirdfigurewidth\linewidth}
            \centering
            \subfloat{Input}
        \end{minipage}
        \begin{minipage}[t]{\onethirdfigurewidth\linewidth}
            \centering
            \subfloat{DPR}
        \end{minipage}
        \begin{minipage}[t]{\onethirdfigurewidth\linewidth}
            \centering
            \subfloat{Ours}
        \end{minipage}
    \end{minipage}
    \end{minipage}
    \caption{Comparison to DPR~\cite{zhou2019deep} on single-view portrait relighting. The input illumination is shown as inset in each relighting result. 
    }
\label{fig:cmp_dpr}
\end{figure}

}
{}

\paragraph{Comparison to single-view portrait relighting methods.}
We compare our method to Deep Portrait Relighting (DPR)~\cite{zhou2019deep} to evaluate the relighting results.
The illumination is represented as the first three bands of Spherical Harmonics (SH) in DPR.
We use their default SH coefficients and calculate the corresponding point light brightness for our model. 
We use a portrait in uniform illumination as the input of DPR.

As shown in Figure~\ref{fig:cmp_dpr}, DPR fails to predict correct relighting effects such as specularities and shadows consistent with the identity-specific geometry and skin material. As a result, the identity is shifted after relighting.
In contrast, our method achieves more faithful portrait relighting results.


\subsection{Ablation Study}
\label{sec:ablation_study}


\begin{table}
\center 
\caption{Quantitative evaluation results of ablation study. In each column, the best number is highlighted in \textbf{bold}. \delete{and the second best one is shown with an {\underline{underscore}}.}
Some corresponding visual results are shown in Figure~\ref{fig:ablation_gt}.
}
\resizebox{\columnwidth}{!}{%
\begin{tabular}{l|ccc|ccc}
\hline
           & \multicolumn{3}{c|}{Subject A} & \multicolumn{3}{c}{Subject B} \\ \hline
Method     & MAE $\downarrow$ & SSIM $\uparrow$ & LPIPS $\downarrow$  & MAE $\downarrow$ & SSIM $\uparrow$ & LPIPS $\downarrow$ \\ \hline
NL        & 10.47         & 0.665   & 0.417      & 13.87          & 0.601  & 0.440       \\
NL + ENV & 7.10         & 0.677     & 0.418    & 9.78          & 0.604    & 0.445     \\
NL + LCL & 9.74         & 0.661     & 0.428    & 12.19          & 0.601    & 0.449     \\
NL + TS  & 8.03         & 0.672      & 0.401   & 9.77          & 0.597     & 0.423    \\
Ours       & \textbf{6.32}         & \textbf{0.707}  & \textbf{0.334}       & \textbf{7.99}          & \textbf{0.635}   & \textbf{0.356}      \\ \hline
\end{tabular}
\label{tab:ablation}%
}
\end{table}

We perform ablation studies to evaluate the effectiveness of our physically-inspired appearance decoder $\colordecoder$. Specifically, we compare our method to four alternative design options:
\begin{enumerate}
    \item NL: We remove the linear \yl{lighting} branch of $\colordecoder$ and directly feed the concatenated lighting condition $\light$ and other latent codes to an ordinary non-linear network with the same layers as for appearance prediction.
    \item NL + ENV: We use the same network architecture as in (1) but use the Light Stage to simulate environment maps~\cite{debevec2002lighting} instead of group lights for training.
    \item NL + LCL: We adopt the same network architecture as in (1) and add a lighting consistency loss inspired by the recent single image portrait relighting method~\cite{yeh2022learning} to enforce the linearity of lighting.
    \item NL + TS: We adopt the same network architecture as in (1) and use a two-stage training framework~\cite{bi2021deep} for relighting. Specifically, we initially train an appearance decoder $\colordecoder$ for OLAT relighting, and subsequently use the trained network to synthesize data for training the environment map relighting appearance decoder.
\end{enumerate}

\begin{figure*}
    \centering
    \begin{minipage}[t]{\linewidth}
        \centering
        \includegraphics[width=\onesixthfigurewidth\linewidth]{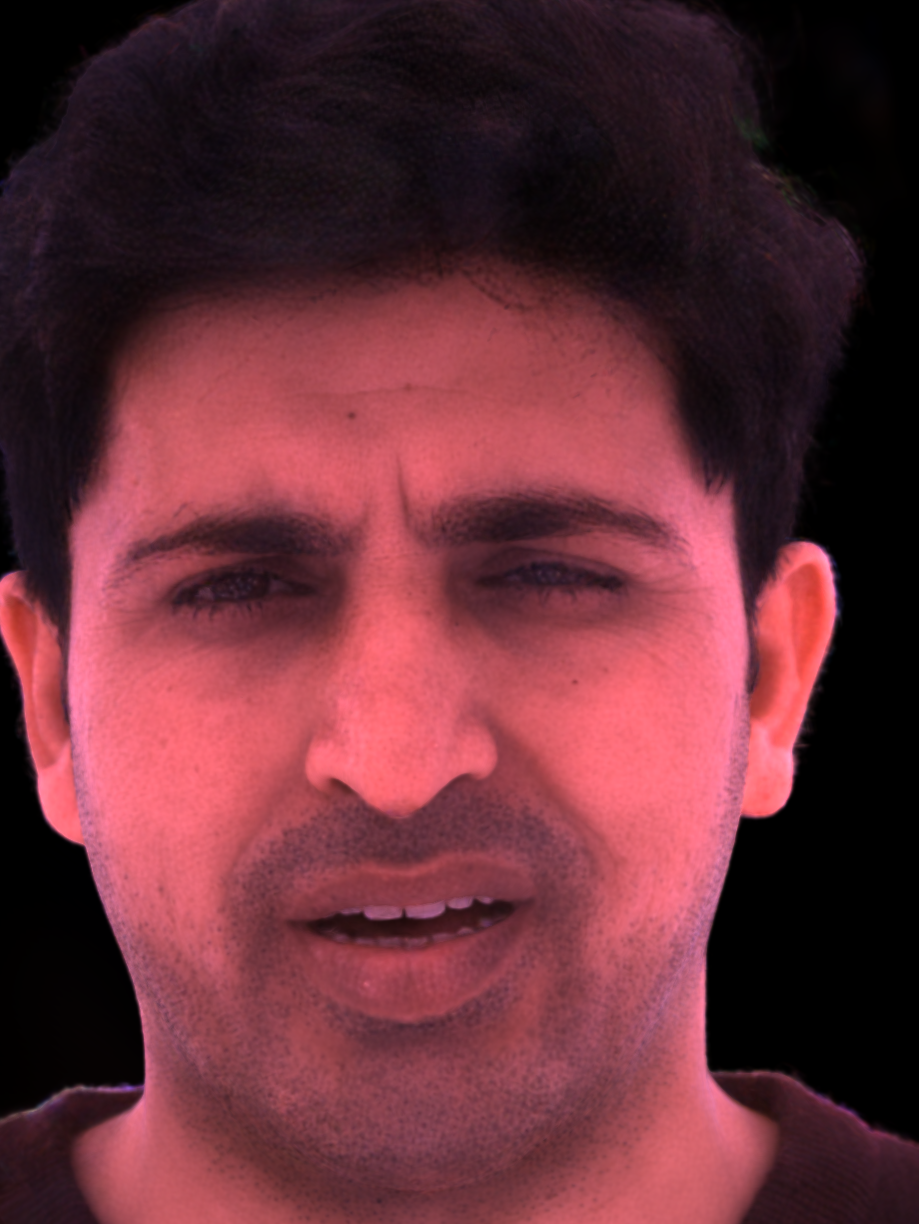}
        \includegraphics[width=\onesixthfigurewidth\linewidth]{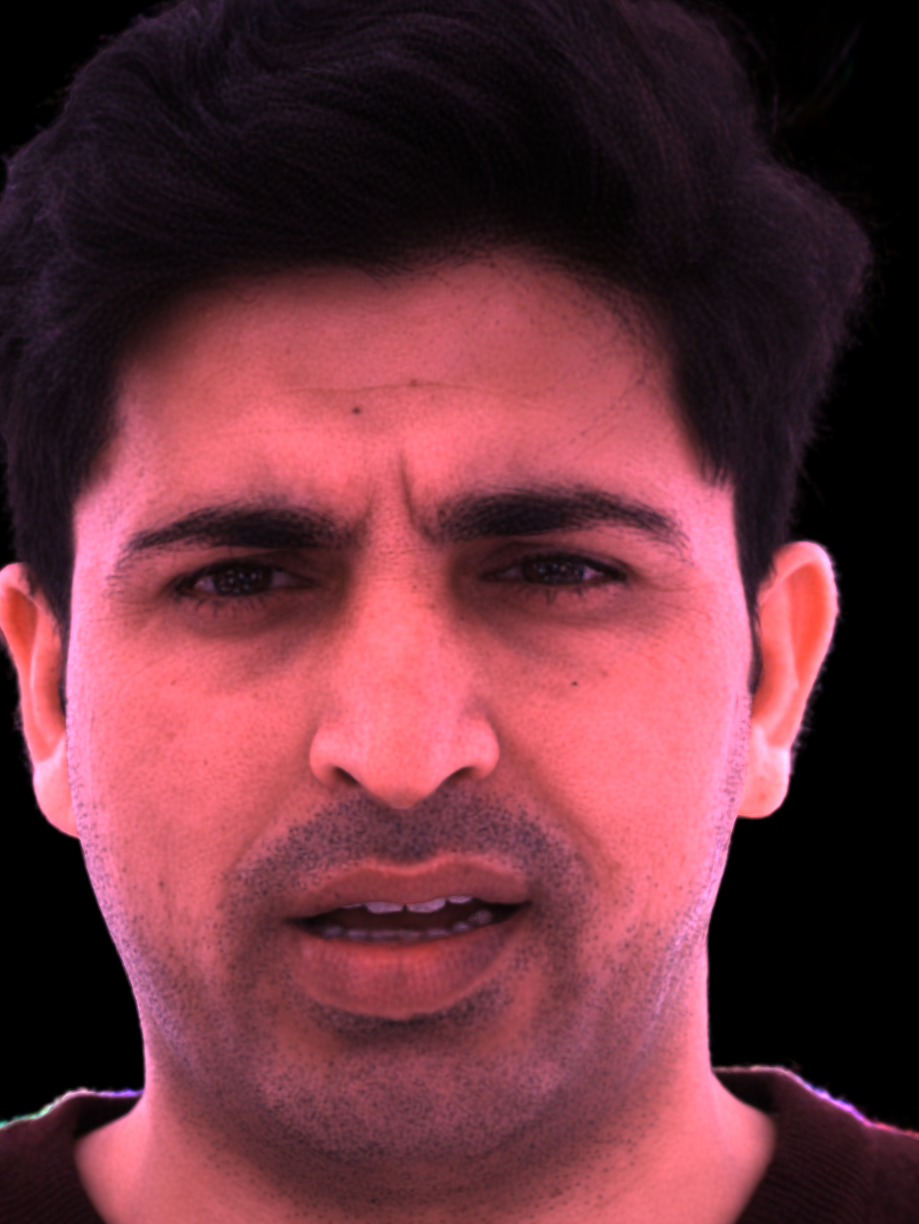}
        \includegraphics[width=\onesixthfigurewidth\linewidth]{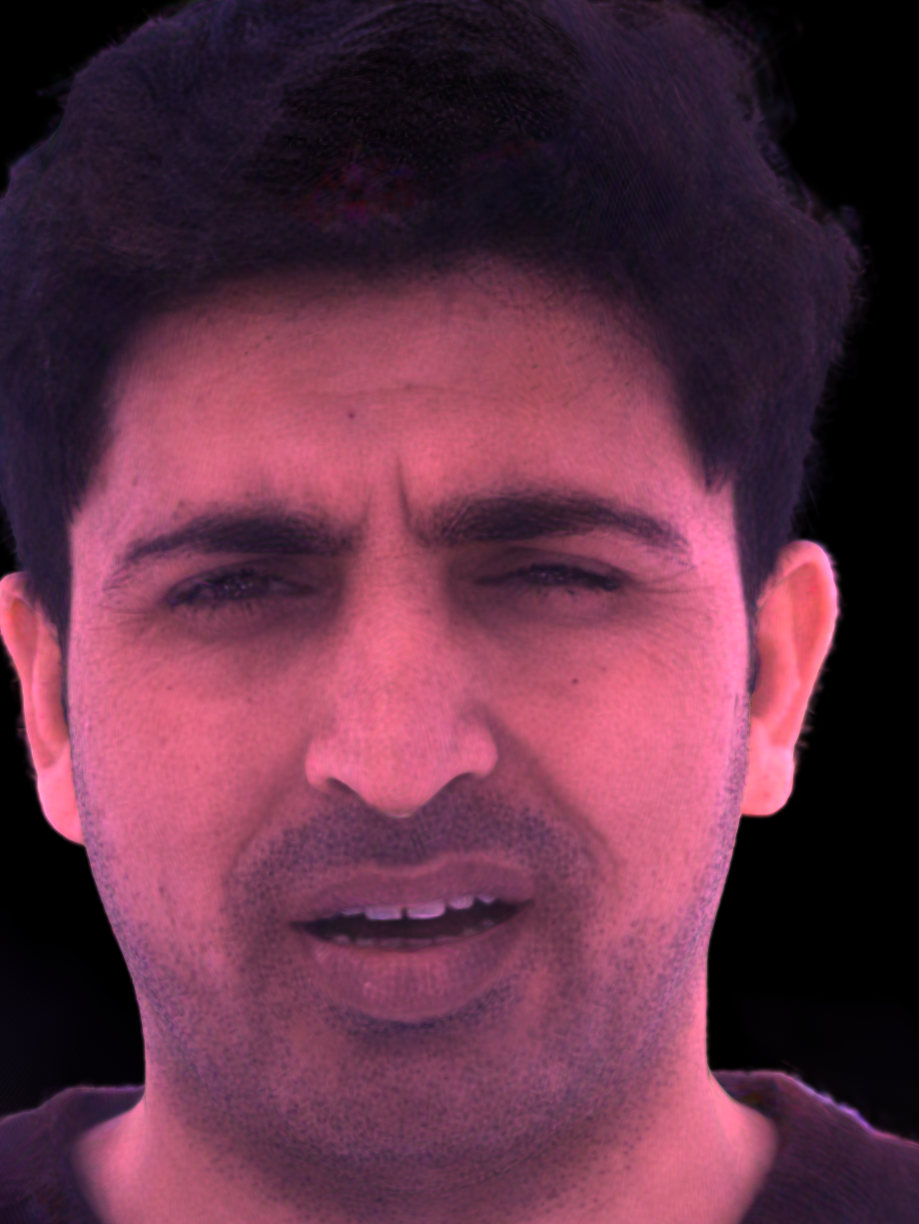}
        \includegraphics[width=\onesixthfigurewidth\linewidth]{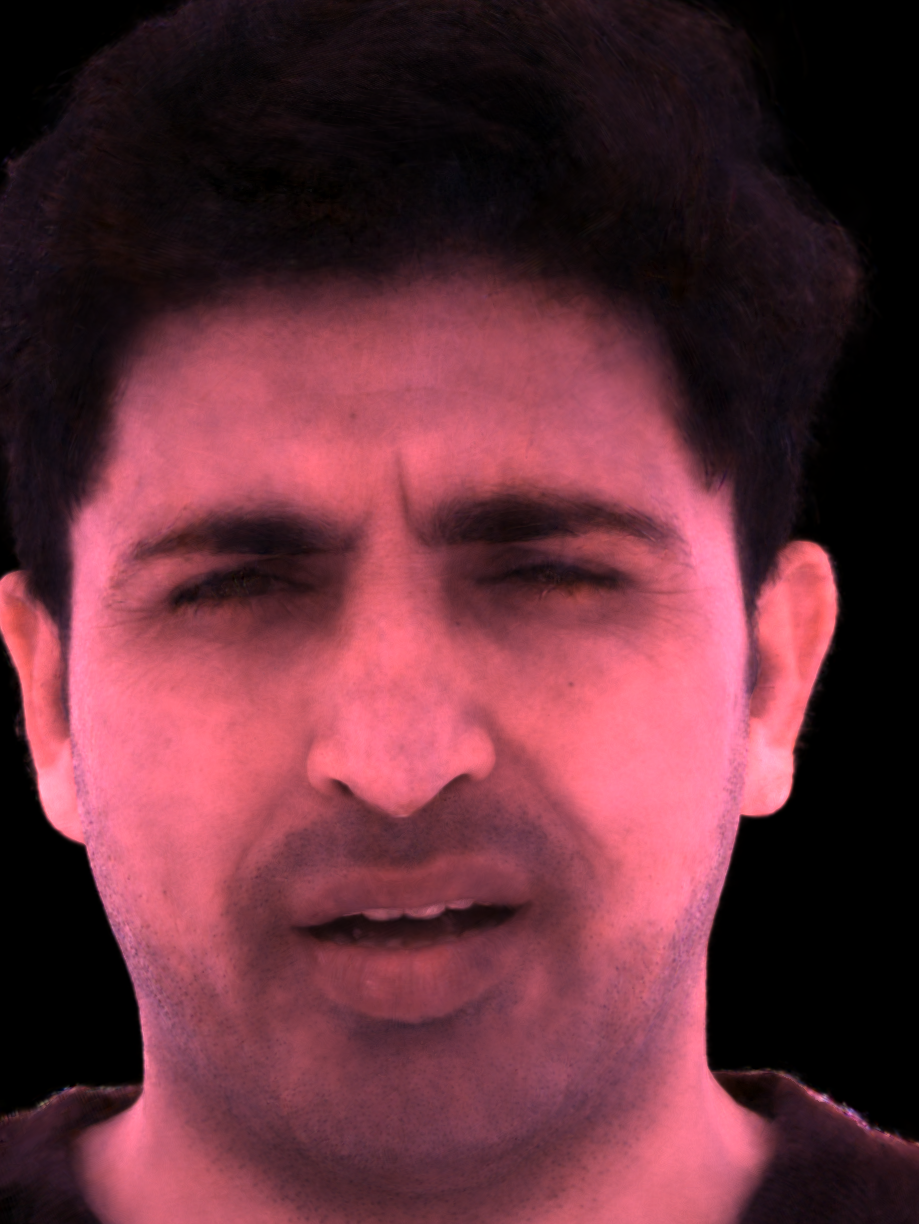}
        \includegraphics[width=\onesixthfigurewidth\linewidth]{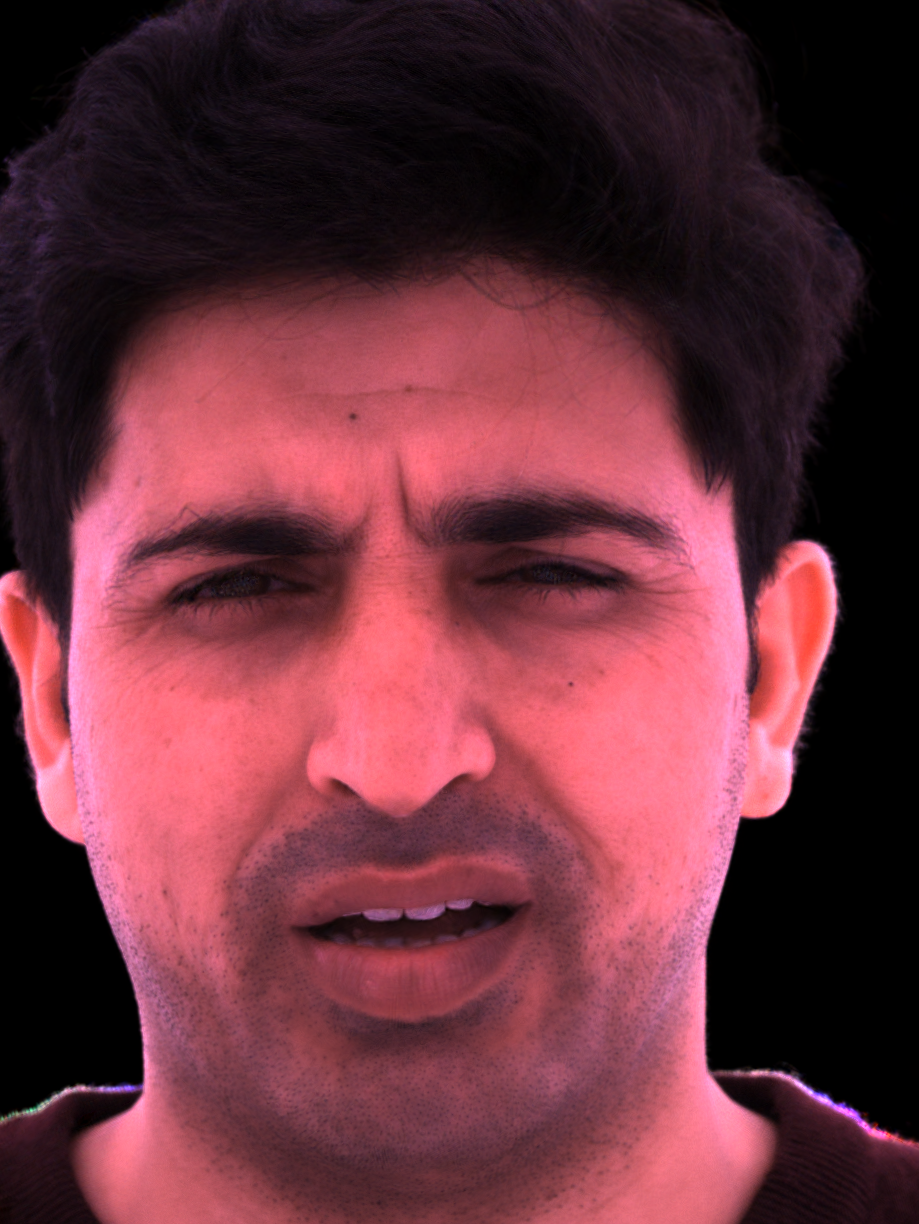}
        \includegraphics[width=\onesixthfigurewidth\linewidth]{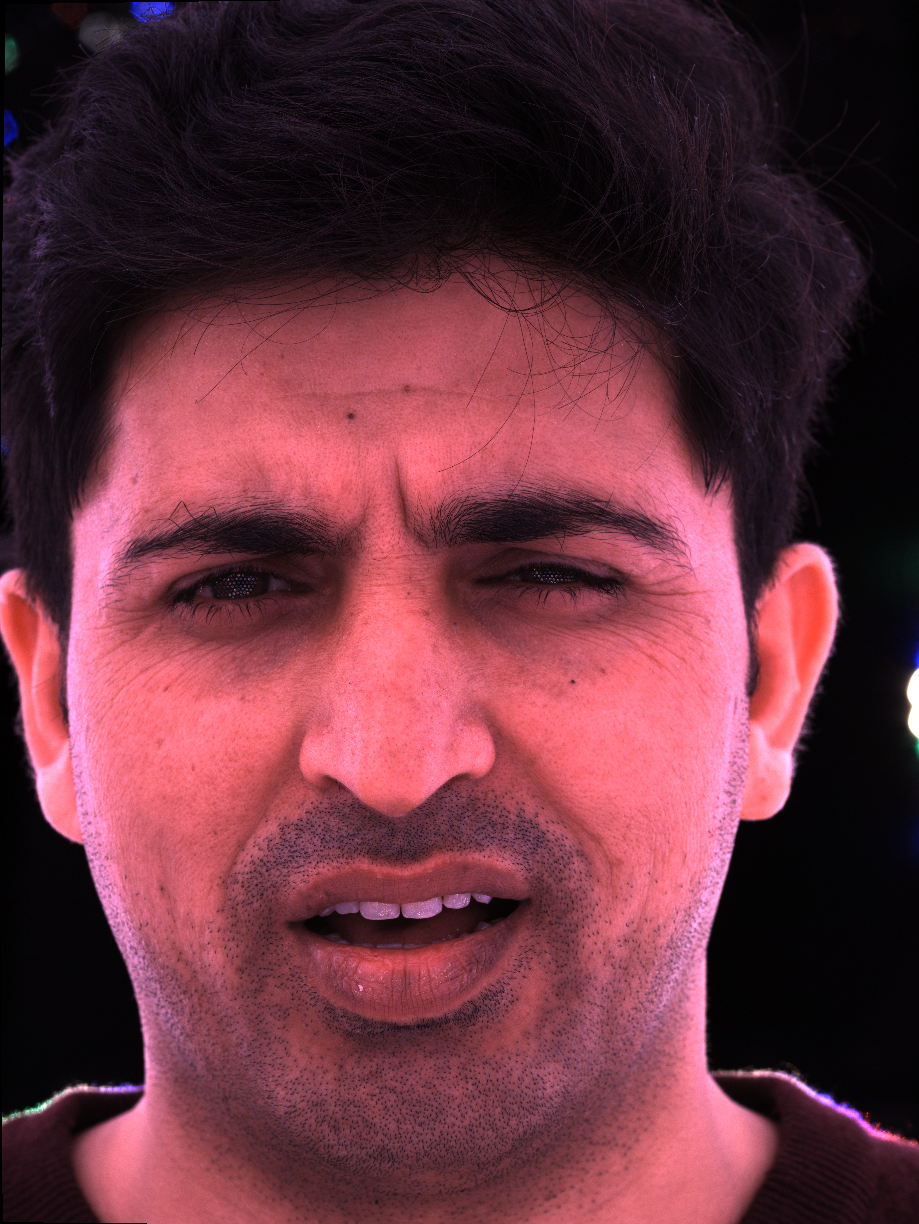}
    \end{minipage}

    \begin{minipage}[t]{\linewidth}
        \centering
        \includegraphics[width=\onesixthfigurewidth\linewidth]{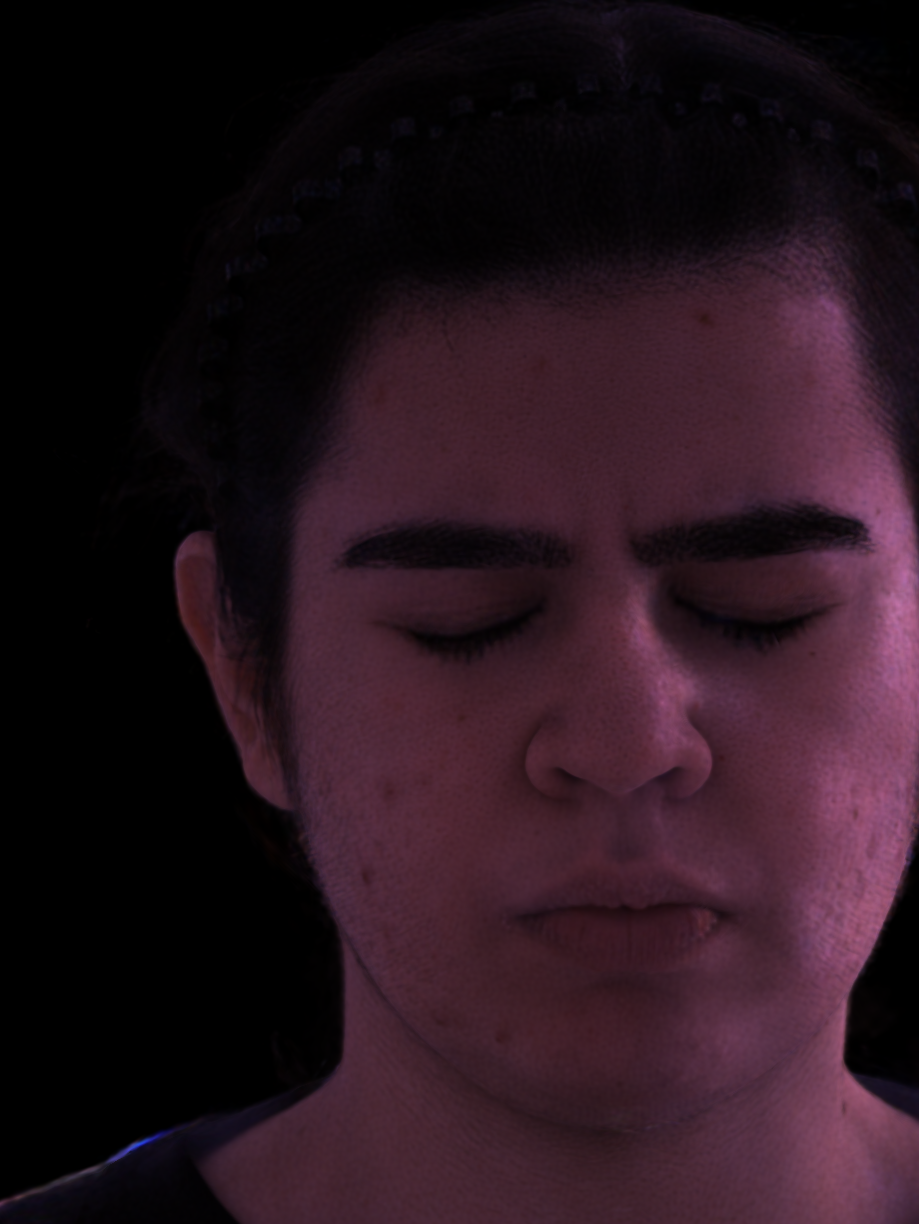}
        \includegraphics[width=\onesixthfigurewidth\linewidth]{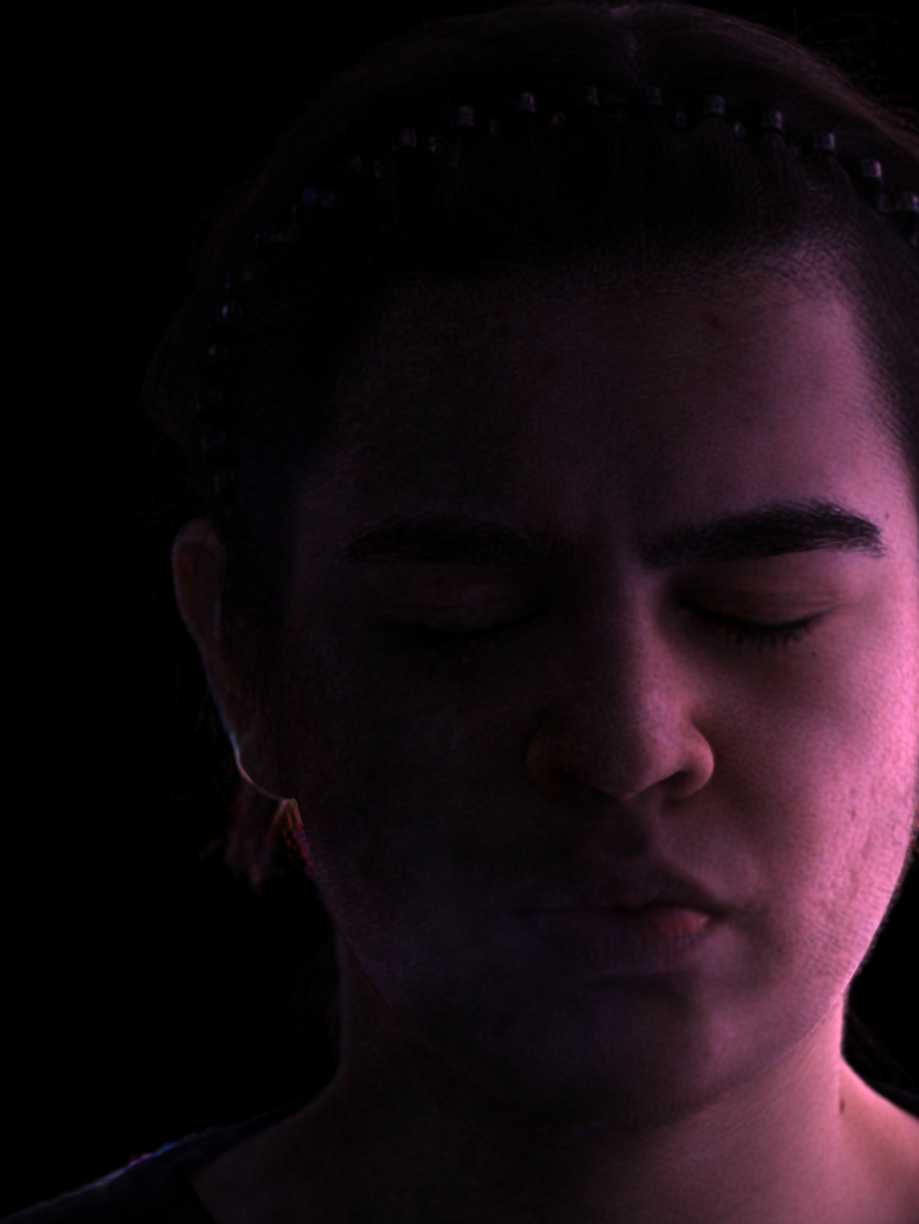}
        \includegraphics[width=\onesixthfigurewidth\linewidth]{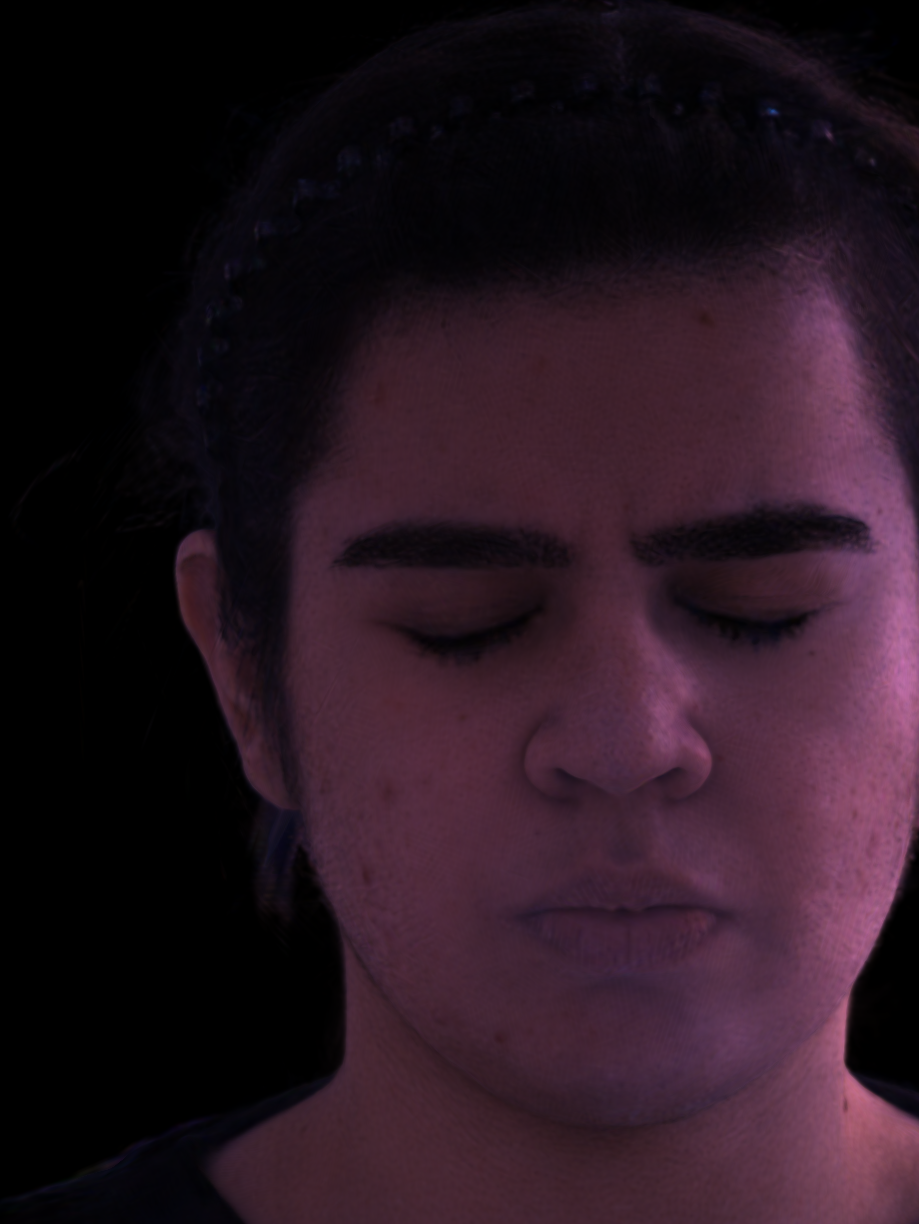}
        \includegraphics[width=\onesixthfigurewidth\linewidth]{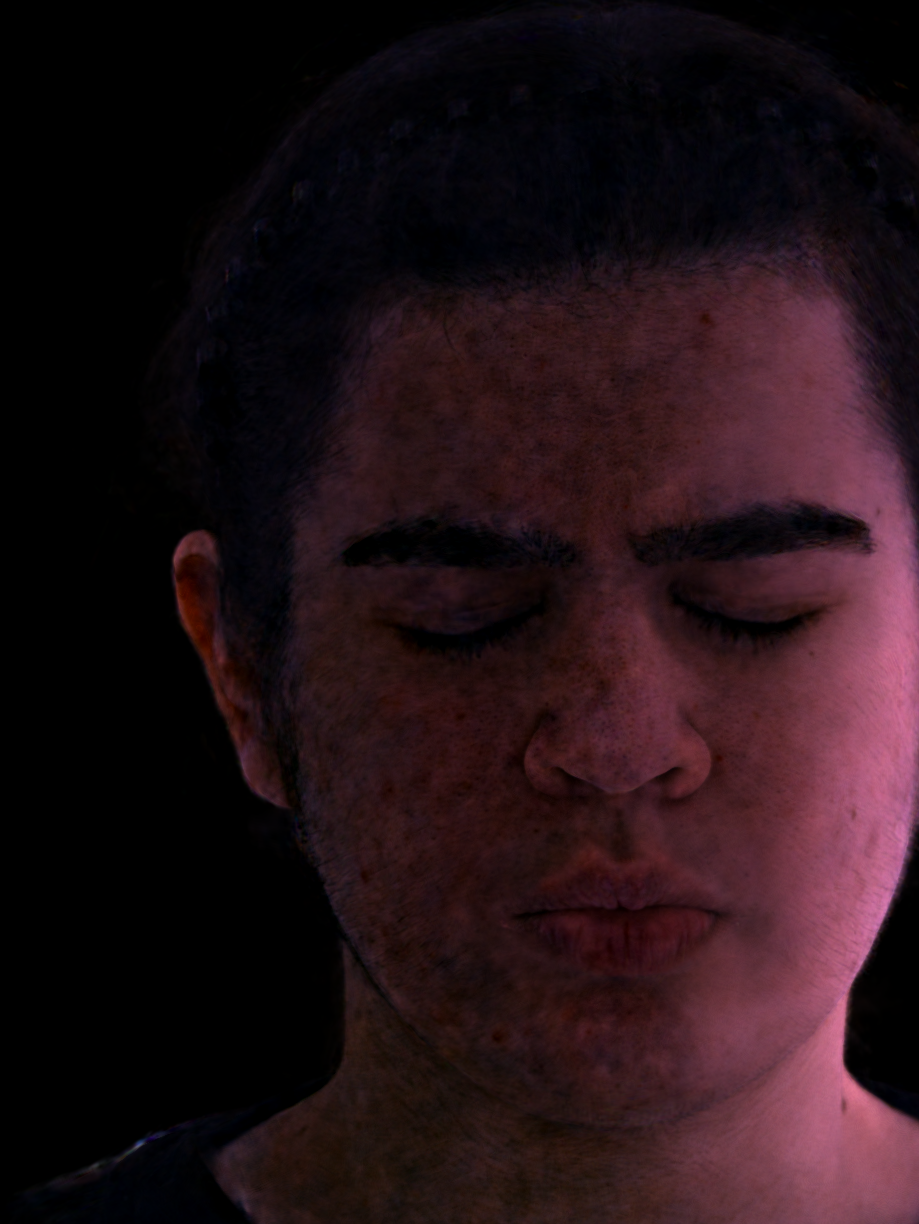}
        \includegraphics[width=\onesixthfigurewidth\linewidth]{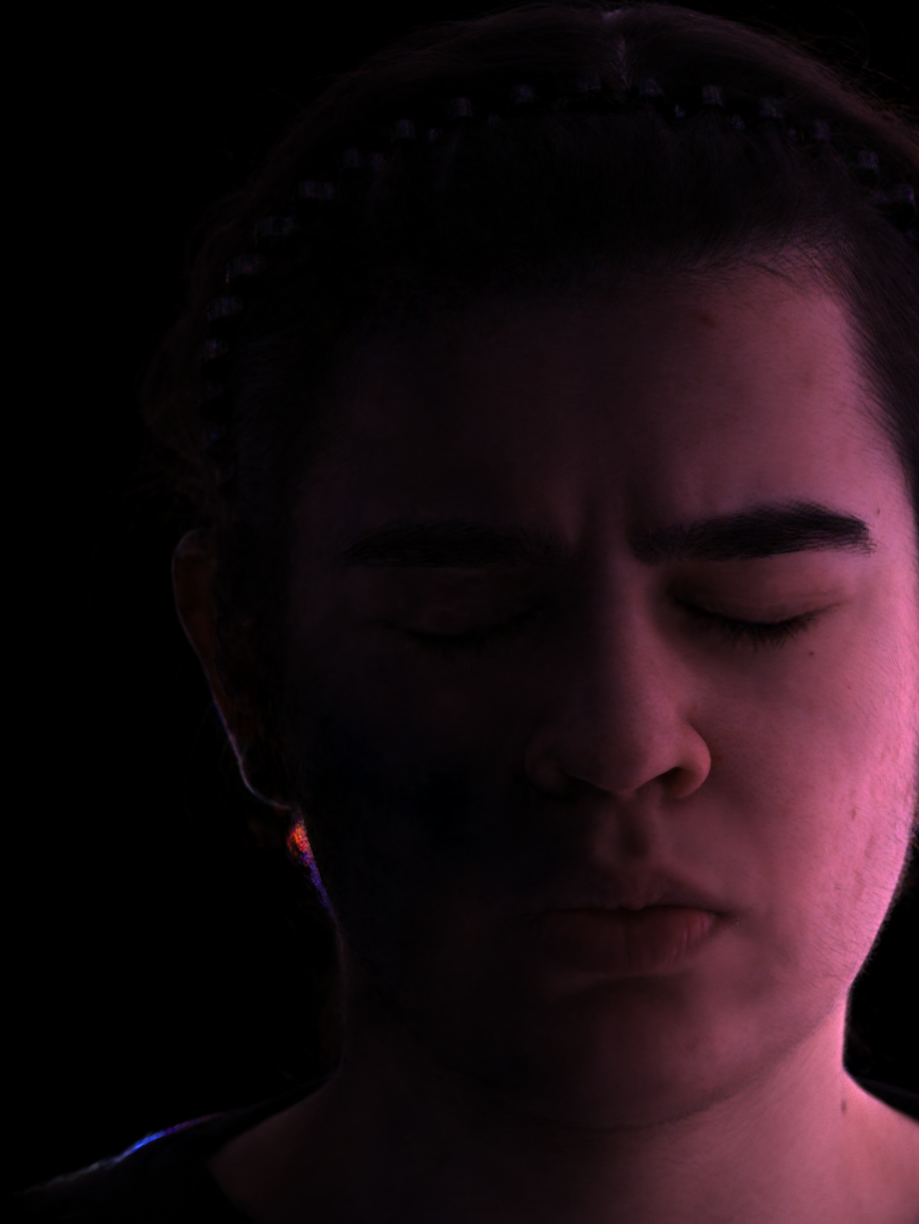}
        \includegraphics[width=\onesixthfigurewidth\linewidth]{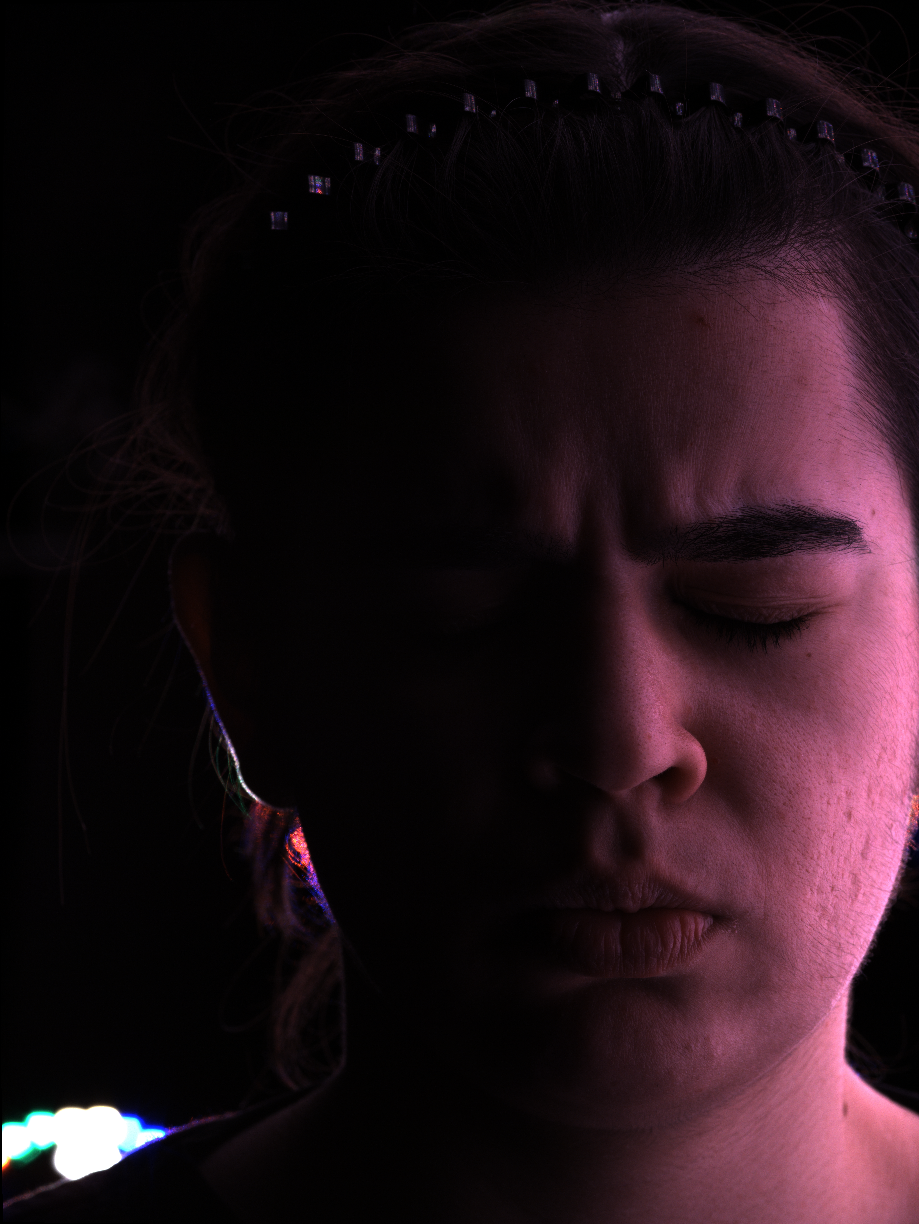}
    \end{minipage}

    \vspace{5pt}
    \begin{minipage}{\linewidth}
        \centering
        \begin{minipage}[t]{\onesixthfigurewidth\linewidth}
            \centering
            \subfloat{NL}
        \end{minipage}
        \begin{minipage}[t]{\onesixthfigurewidth\linewidth}
            \centering
            \subfloat{NL + ENV}
        \end{minipage}
        \begin{minipage}[t]{\onesixthfigurewidth\linewidth}
            \centering
            \subfloat{NL + LCL}
        \end{minipage}
        \begin{minipage}[t]{\onesixthfigurewidth\linewidth}
            \centering
            \subfloat{NL + TS}
        \end{minipage}
        \begin{minipage}[t]{\onesixthfigurewidth\linewidth}
            \centering
            \subfloat{Ours}
        \end{minipage}
        \begin{minipage}[t]{\onesixthfigurewidth\linewidth}
            \centering
            \subfloat{Ground truth}
        \end{minipage}
    \end{minipage}
    \caption{Ablation study results on Subjects A (top) and B (bottom) about our physically inspired linear light branch for the appearance decoder $\colordecoder$.
    From left to right: relighting results of four alternative baselines (see detailed explanations in Section~\ref{sec:ablation_study}), our results, and the ground truth. \revision{Note that here we use simulated environment map light which is similar to the lighting conditions that NL + ENV is trained on. Therefore, the results of NL + ENV are comparable to ours in this figure but downgrades significantly when using real HDR environments for testing (see more results in our supplementary materials).}
    }
\label{fig:ablation_gt}
\end{figure*}

We capture \revision{600} frames for each subject under various preset lighting conditions in \yl{a} Light Stage as ground truth for quantitative evaluation. Quantitative results are summarized in Table~\ref{tab:ablation} and qualitative comparisons are shown in \revision{Figure~\ref{fig:ablation_gt}}.
Note that not all the lighting conditions can be simulated in a Light Stage due to hardware limitations such as the maximum brightness of a lighting unit.
The results demonstrate that our linear lighting branch of $\colordecoder$ significantly enhances the generalization performance for relighting.


\subsection{Video-Driven Animation}

Our volumetric avatar can be animated by replacing the motion encoder $\motionencoder$ with an application-specific module predicting the low-dimensional expression code $\expressioncode$.
Existing methods perform domain adaptation on synthetic datasets~\cite{lombardi2018deep} or use triplet supervision~\cite{zhang2022video} to train the expression code predictor.
In our implementation, we simply use an off-the-shelf expression regressor similar to \cite{weise2011realtime} to predict the identity-independent blendshape weights of each frame from the frontal view in our captured data. Then we train a three-layer MLP to predict the expression code $\expressioncode$ from the blendshape weights. Our volumetric avatar can be animated by the extracted blendshape weights from monocular videos. 

We find that the rigid head rotation and translation are successfully disentangled from the expression code even without explicit constraint.
Thanks to the consistent structures of the base mesh and volumetric primitives, we can explicitly constrain the motion beyond face, achieving plausible animation results.
Figure~\ref{fig:animation} shows some performance-driven animation results.
\ifthenelse{\equal{\final}{1}}
{
Please refer to our accompanying video for the corresponding animations results.
}
{}

\ifthenelse{\equal{\final}{1}}
{

\begin{figure}
    \centering
    \begin{minipage}{\linewidth}
        \centering
        \begin{minipage}[t]{\onethirdfigurewidth\linewidth}
            \centering
            \includegraphics[width=\linewidth]{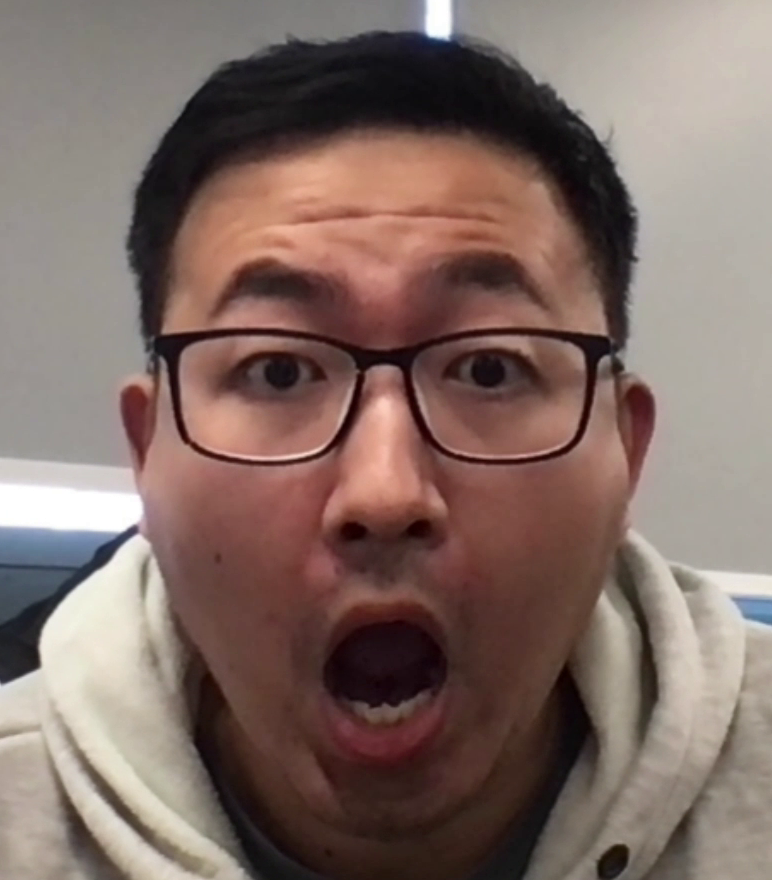}
            \includegraphics[width=\linewidth]{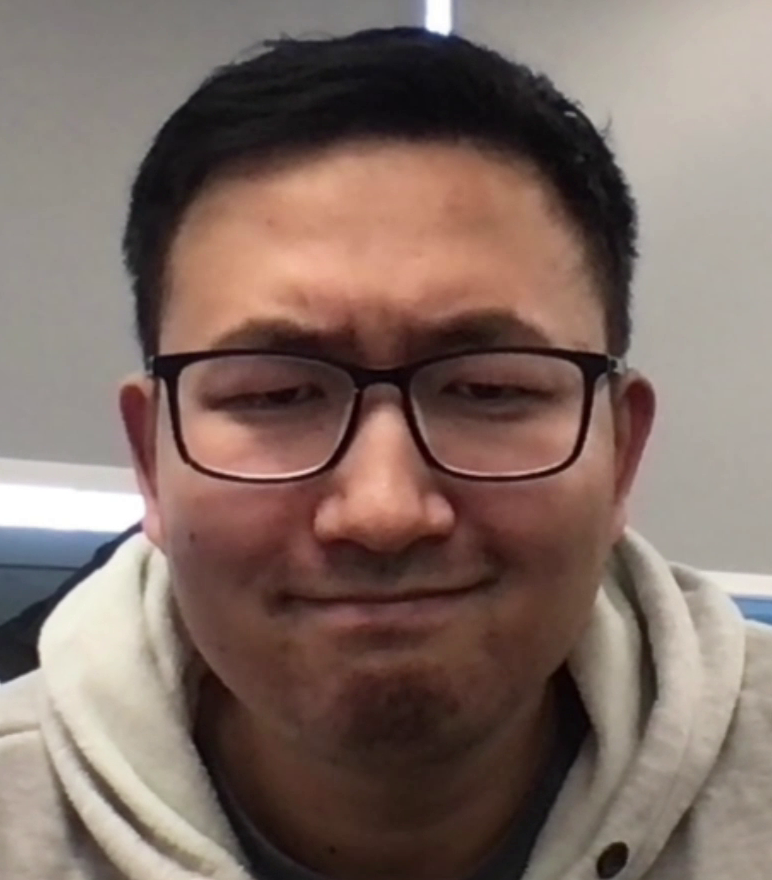}
            \includegraphics[width=\linewidth]{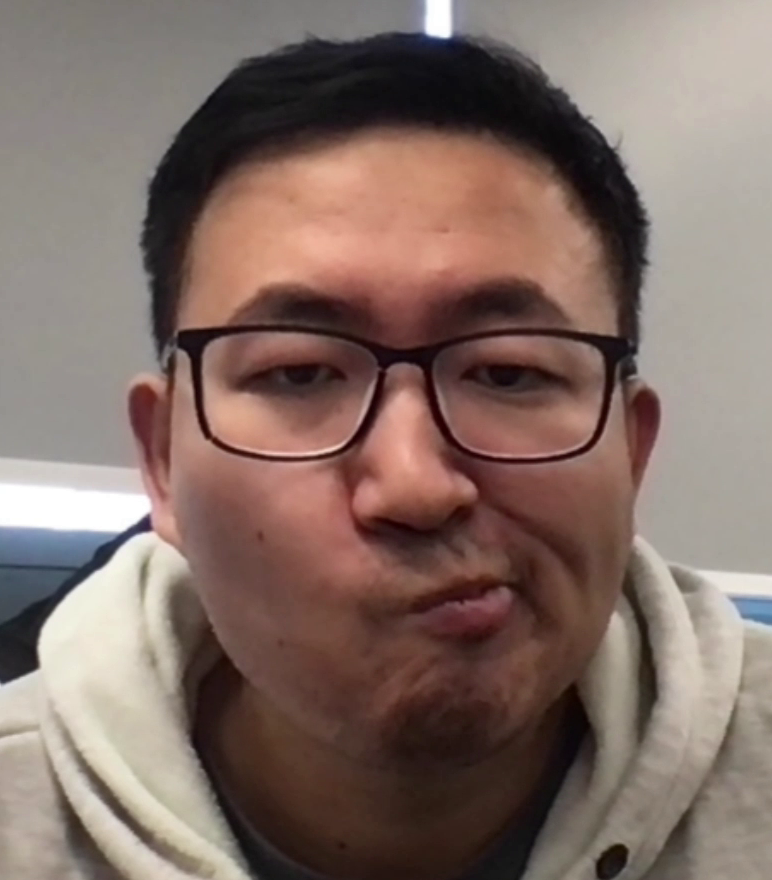}
            \subfloat{Input}
        \end{minipage}
        \begin{minipage}[t]{\onethirdfigurewidth\linewidth}
            \centering
            \includegraphics[width=\linewidth]{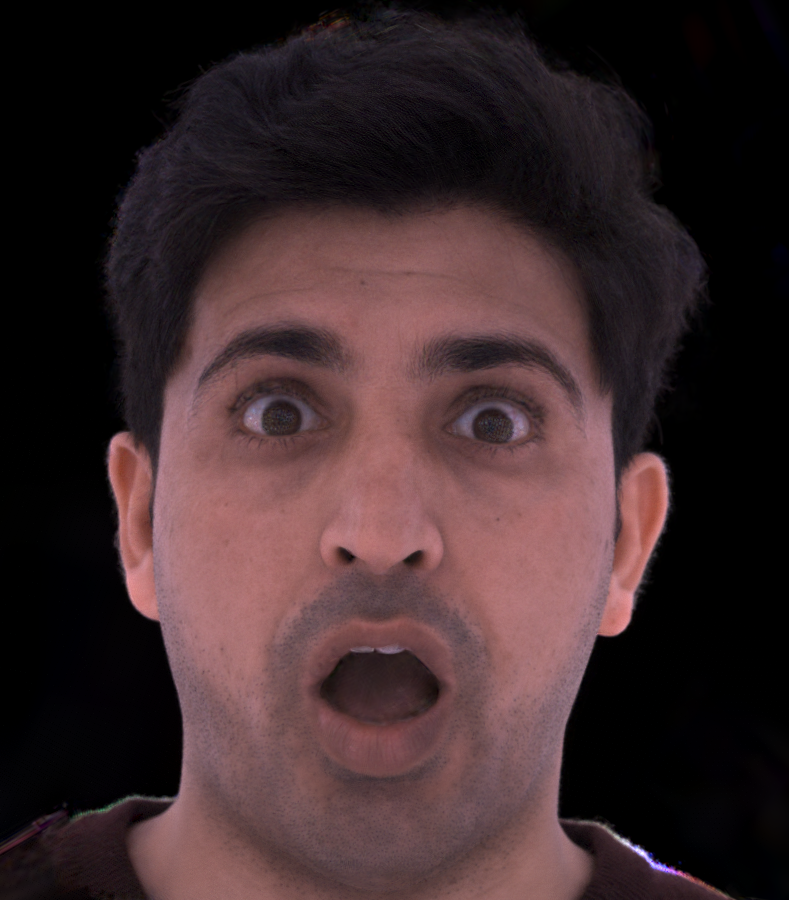}
            \includegraphics[width=\linewidth]{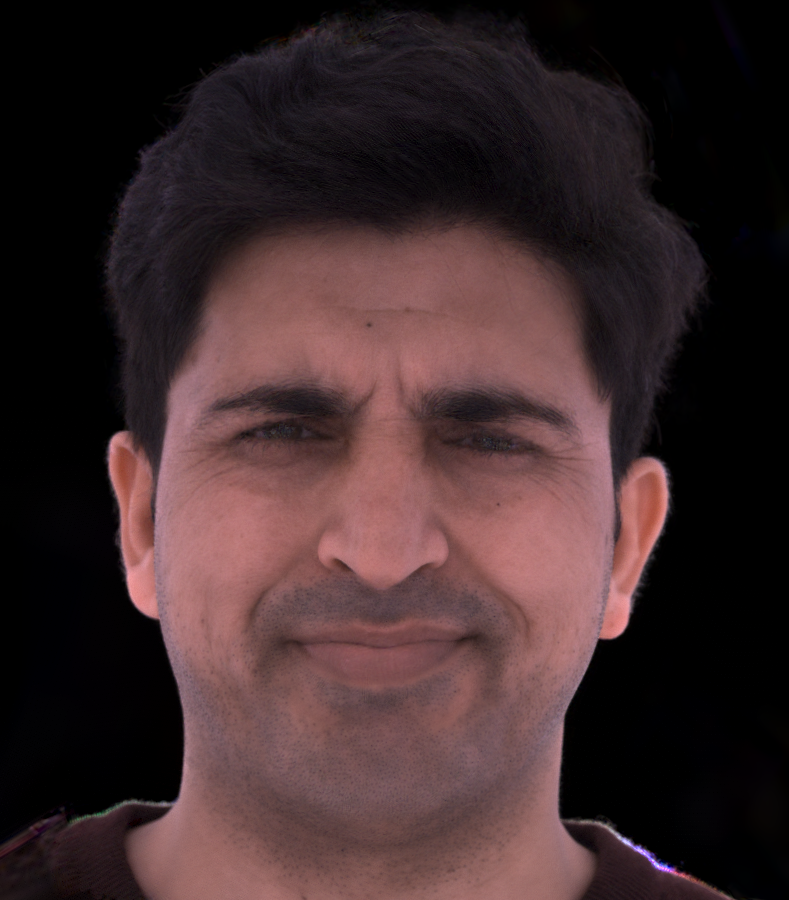}
            \includegraphics[width=\linewidth]{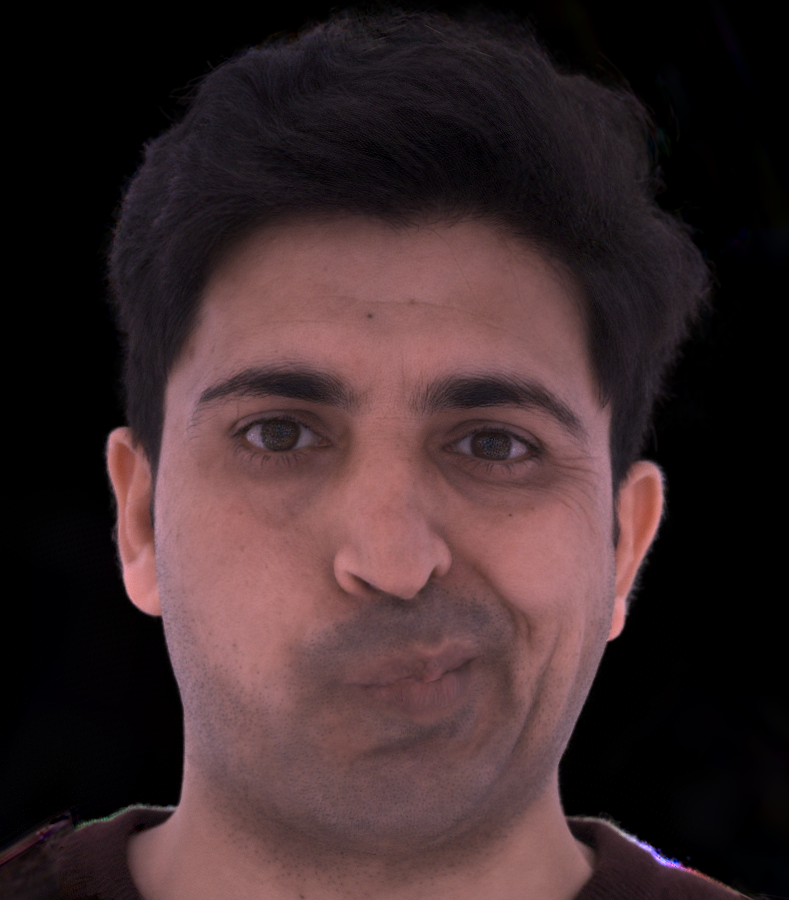}
            \subfloat{Subject B}
        \end{minipage}
        \begin{minipage}[t]{\onethirdfigurewidth\linewidth}
            \centering
            \includegraphics[width=\linewidth]{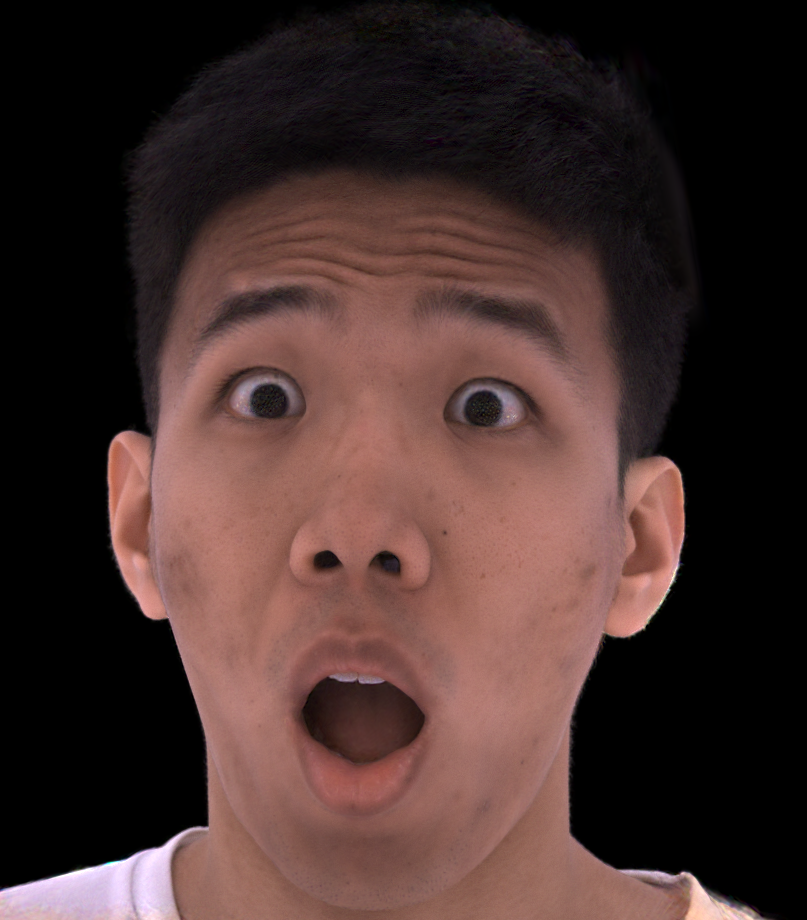}
            \includegraphics[width=\linewidth]{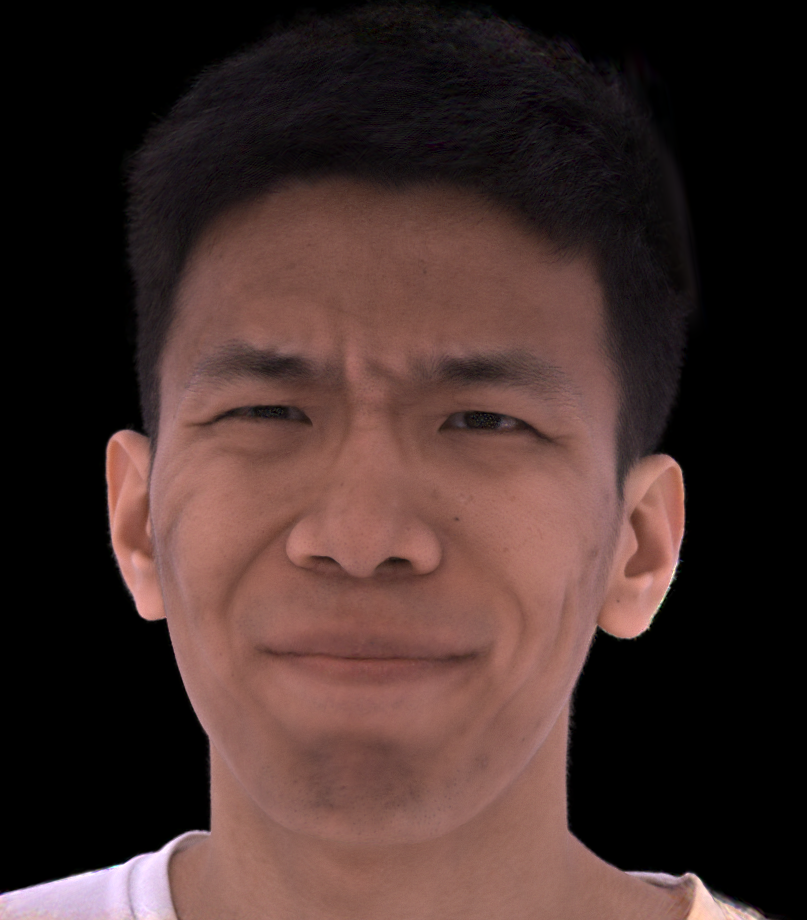}
            \includegraphics[width=\linewidth]{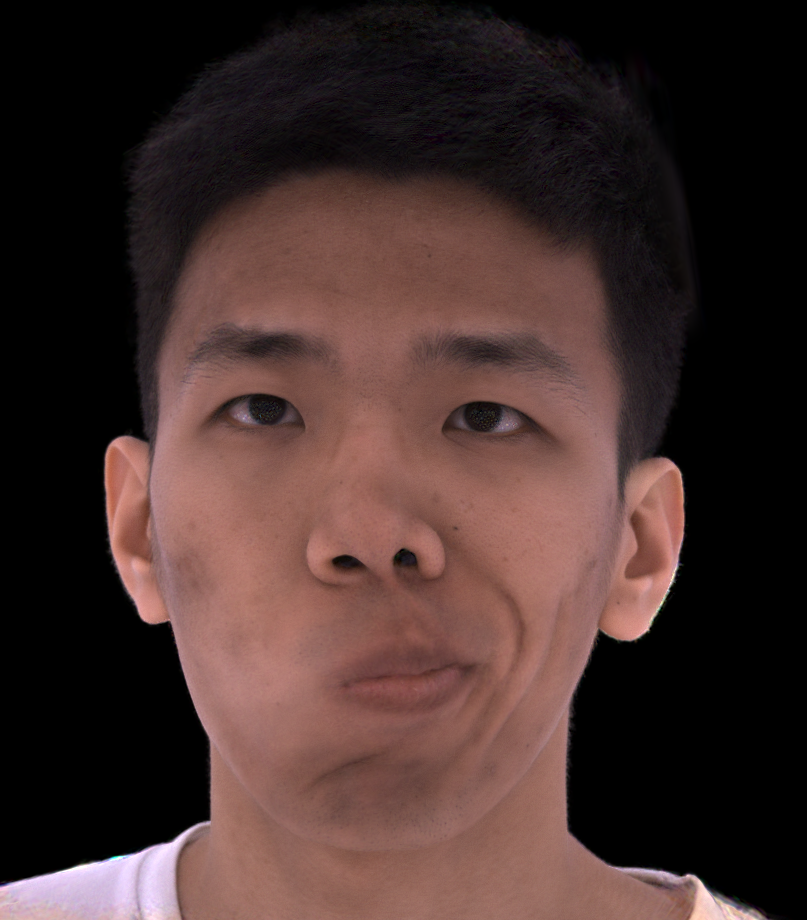}
            \subfloat{Subject C}
        \end{minipage}
    \end{minipage}
    \caption{Video-driven animation results. Our method can faithfully generate identity-specific dynamic wrinkle details for different expressions.}
\label{fig:animation}
\end{figure}

}
{}


\section{Conclusion and Future Work}

In this work, we propose a novel framework, \yl{named} \sysname, for \yl{capturing and reconstructing} high-fidelity and relightable 3D avatars in a practical and efficient setting.
We train the framework with dynamic image sequences captured in a Light Stage under varying lighting conditions, enabling natural relighting and video-driven animation.

Our contributions are two-fold. First, we present a novel network architecture that satisfies the linear nature of lighting, allowing for real-time appearance prediction and high-quality relighting effects. Second, we propose to jointly optimize facial geometry and relightable appearance based on image sequences, with 
\yl{the deformation of the base mesh implicitly learned.}
Our tracking-free scheme provides robustness for establishing temporal correspondences between frames under different lighting conditions.
Both qualitative and quantitative experiments demonstrate that our framework achieves superior performance in photorealistic avatar animation and relighting, facilitating further advancements in content creation of 3D avatars.

Despite our promising results, there are some limitations to be addressed in future work.
First, the data capturing apparatus employed in our framework is expensive, which may limit its applicability and adoption. 
Second, due to the lack of sufficient surface constraints, it becomes challenging to perform precise manual control on the learned implicit representation. Future work could explore methods to create relightable avatars with more affordable equipment and investigate representations that offer more flexible control.
\revision{Finally, we are interested in extending our method to handle near-field and high-frequency relighting~\cite{bi2021deep, sun2020light} as well as accessories such as glasses~\cite{li2023megane}.}

\begin{acks}
We would like to thank Qianfang Zou and Xuesong Niu for being our capture subjects, the authors of \cite{lombardi2021mixture} and \cite{zhou2019deep} for releasing their source code, as well as anonymous reviewers for their valuable feedback.
\end{acks}


\appendix

\setcounter{table}{0}  
\setcounter{figure}{0}
\renewcommand\thesection{\Alph{section}}
\renewcommand{\thetable}{A\arabic{table}}
\renewcommand{\thefigure}{A\arabic{figure}}

\begin{figure*}
\centering
    \includegraphics[width=\linewidth]{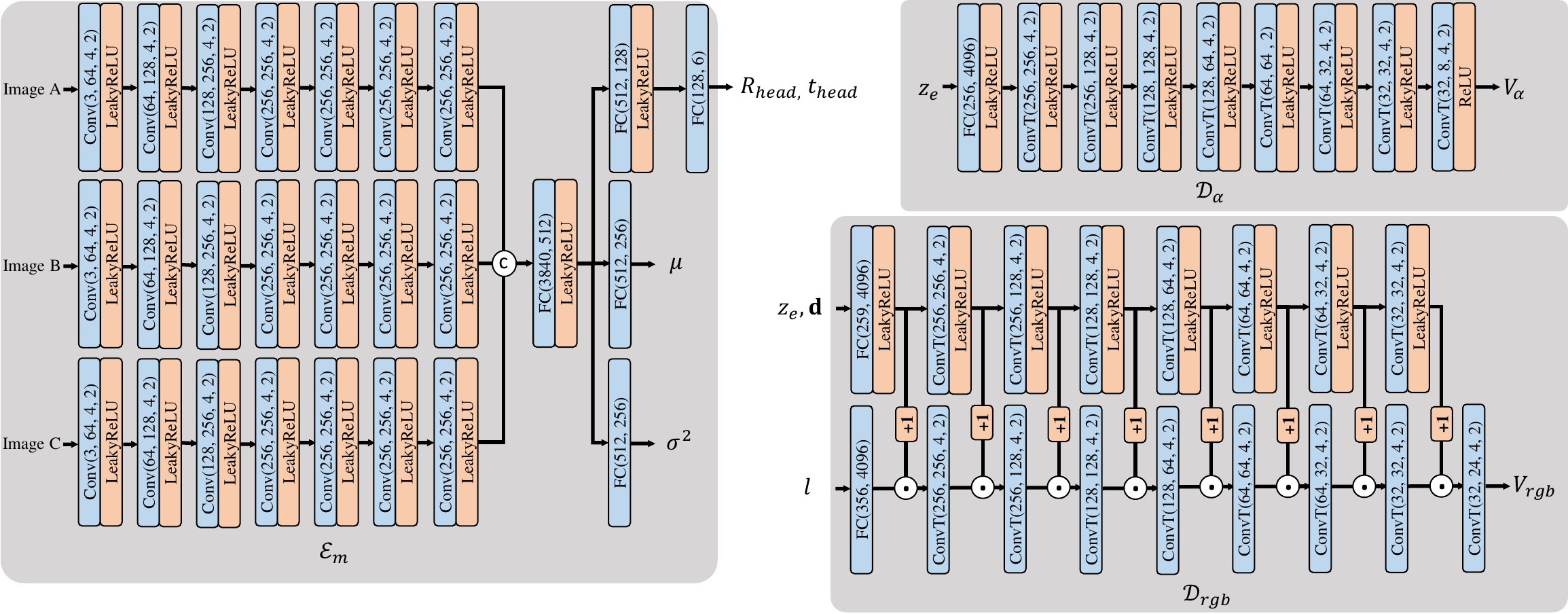}
    \captionof{figure}{Detailed architecture of our motion encoder $\motionencoder$, opacity decoder $\opacitydecoder$, and appearance decoder $\colordecoder$. The convolutional layer is represented as $Conv(inchs, outchs, kernel size, stride)$, where $inchs$ is the number of input channels and $outchs$ is the number of output channels. The representation of transposed convolutional layers are similar. The fully connected layer is represented as $FC(inchs, outchs)$. 
    The architecture of the transformation decoder $\transformdecoder$ is similar to the opacity decoder, except that there is no activation layer at the end.
    The mesh decoder $\meshdecoder$ is a three-layer MLP with LeakyReLU activation layers, which is omitted in this figure.}
\label{fig:networks}
\end{figure*}


\appendix

\section{Implementation Details}

\paragraph{Network architectures and hyperparameters.}
We provide detailed architectures of our neural networks in Figure~\ref{fig:networks}.
The values of hyperparameters in our implementation are provided in Table~\ref{tab:hyperparameter}, which are identical in all our experiments.

\begin{table}[]
\centering
\caption{Values of our hyperparameters.}
\begin{tabular}{cr|cr|cr}
\hline
Parameter & Value & Parameter & Value & Parameter & Value \\ \hline
$N_{mesh}$     & 7306  & $N_{prim}$     & 16384 & $M$         & 8     \\
${\lambda}_\mathrm{VGG}$    & 0.1   & ${\lambda}_\mathrm{GAN}$    & 0.005 & ${\lambda}_\mathrm{Lap}$    & 0.01  \\
${\lambda}_{pR}$     & 10    & ${\lambda}_{vol}$    & 0.01  & ${\lambda}_\mathrm{KLD}$    & 0.001 \\ \hline
\end{tabular}
\label{tab:hyperparameter}%
\end{table}

\begin{figure}
    \centering
    \begin{minipage}[t]{0.99\linewidth}
        \centering
        \includegraphics[width=\linewidth]{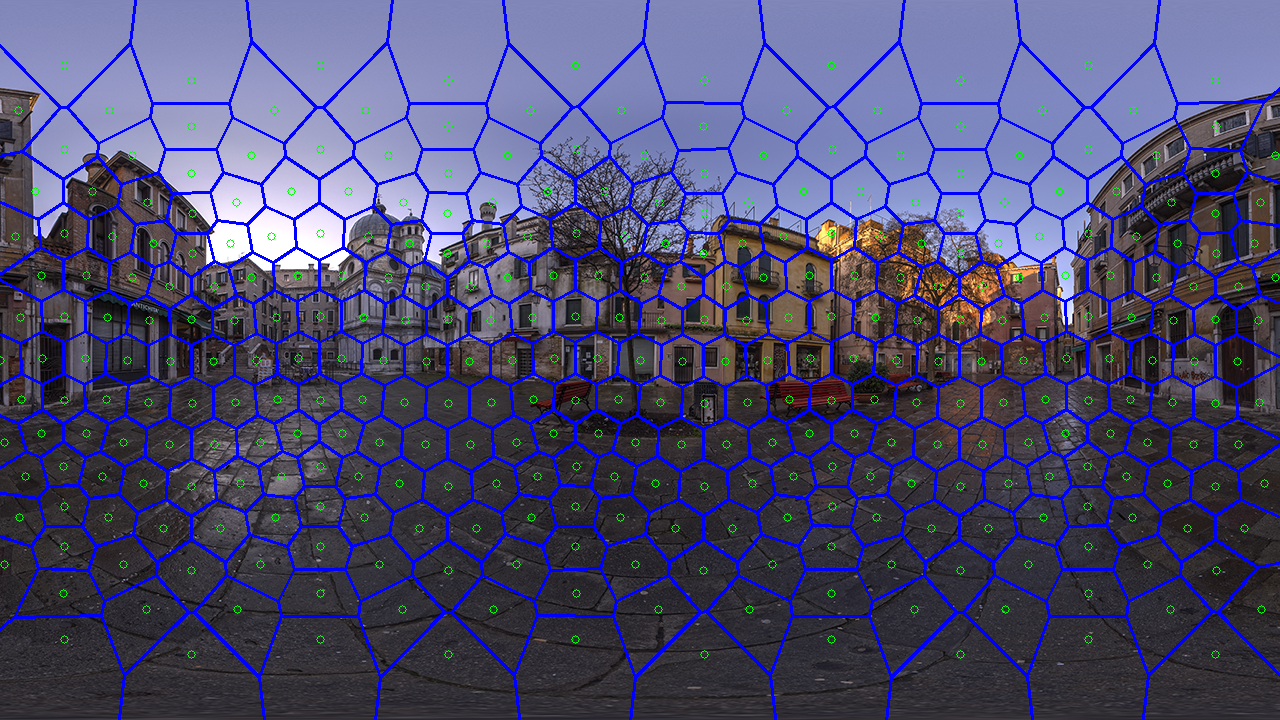}
    \end{minipage}
    \caption{The projected light positions on an environment map. Green circles: positions of 356 lighting units projected on an environment map in the longitude-latitude format. Blue lines: edges of the Voronoi diagram.}
\label{fig:lightposition}
\end{figure}

\paragraph{Environment map relighting.}
Given a high dynamic range environment map in the longitude-latitude format, we extract the lighting condition $\light$ for our appearance decoder. Specifically, we project the position of each light unit of the Light Stage onto the environment map image and split the space using a Voronoi diagram~\cite{aurenhammer1991voronoi}. The corresponding value of $\light$ is set according to the weighted-average pixel values in the cell. We show the projected light positions in Figure~\ref{fig:lightposition}.

\revision{
\section{Additional Results}
\begin{figure*}
    \centering
    \begin{minipage}[t]{\linewidth}
        \centering
        \includegraphics[width=\onesixthfigurewidth\linewidth]{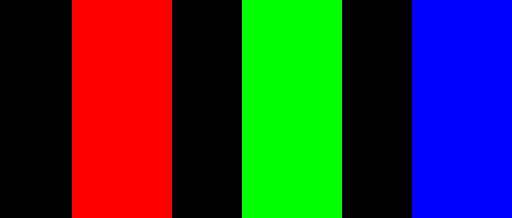}
        \includegraphics[width=\onesixthfigurewidth\linewidth]{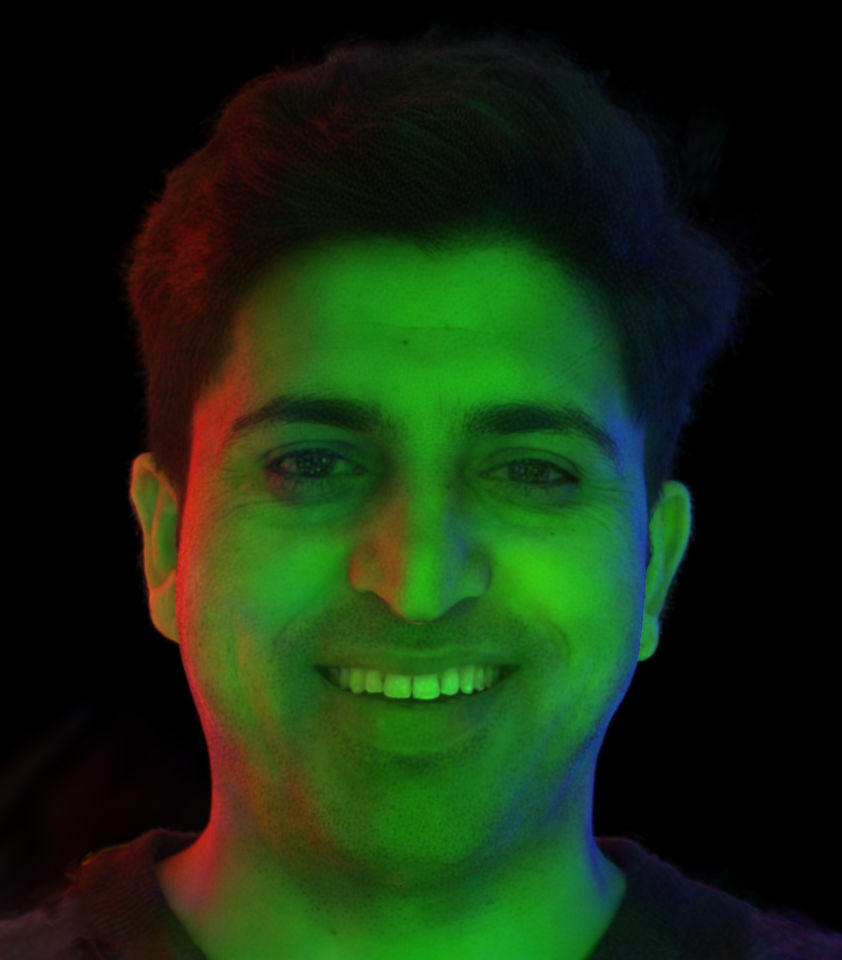}
        \includegraphics[width=\onesixthfigurewidth\linewidth]{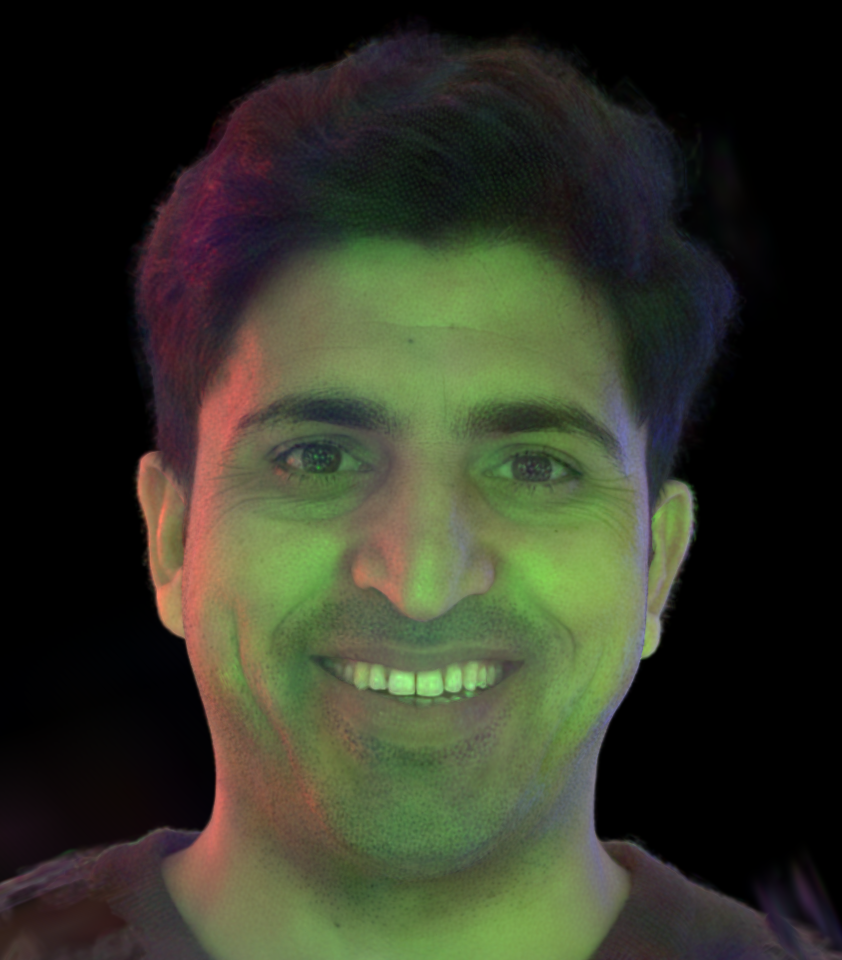}
        \includegraphics[width=\onesixthfigurewidth\linewidth]{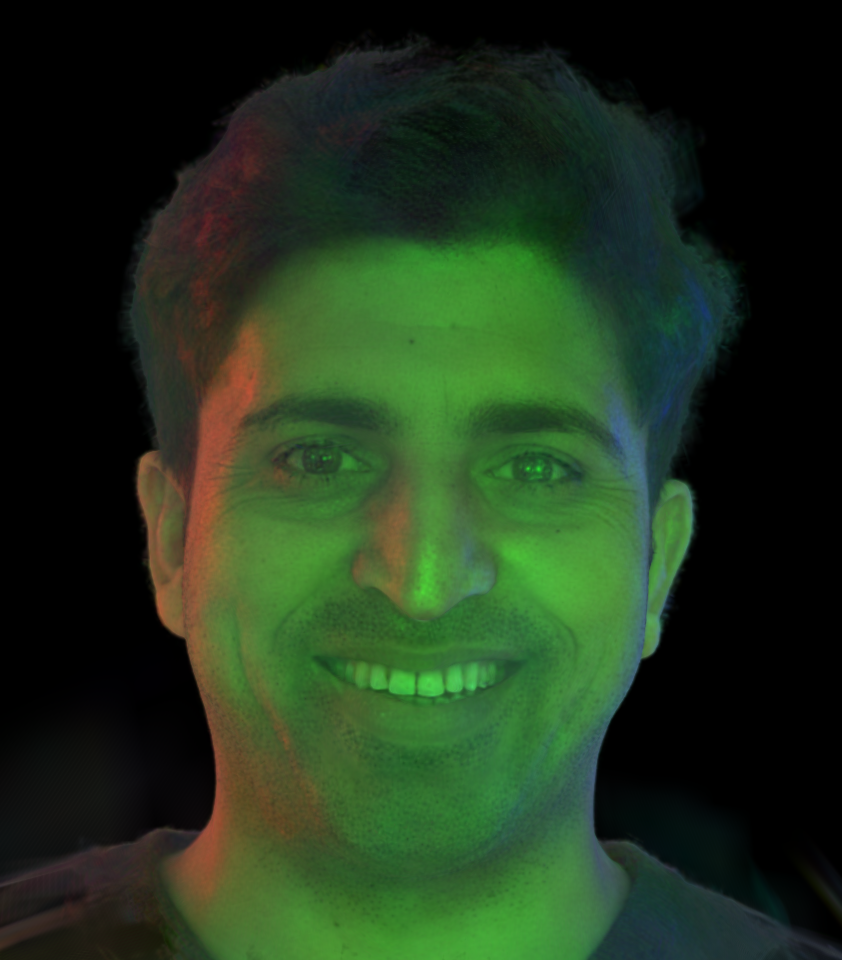}
        \includegraphics[width=\onesixthfigurewidth\linewidth]{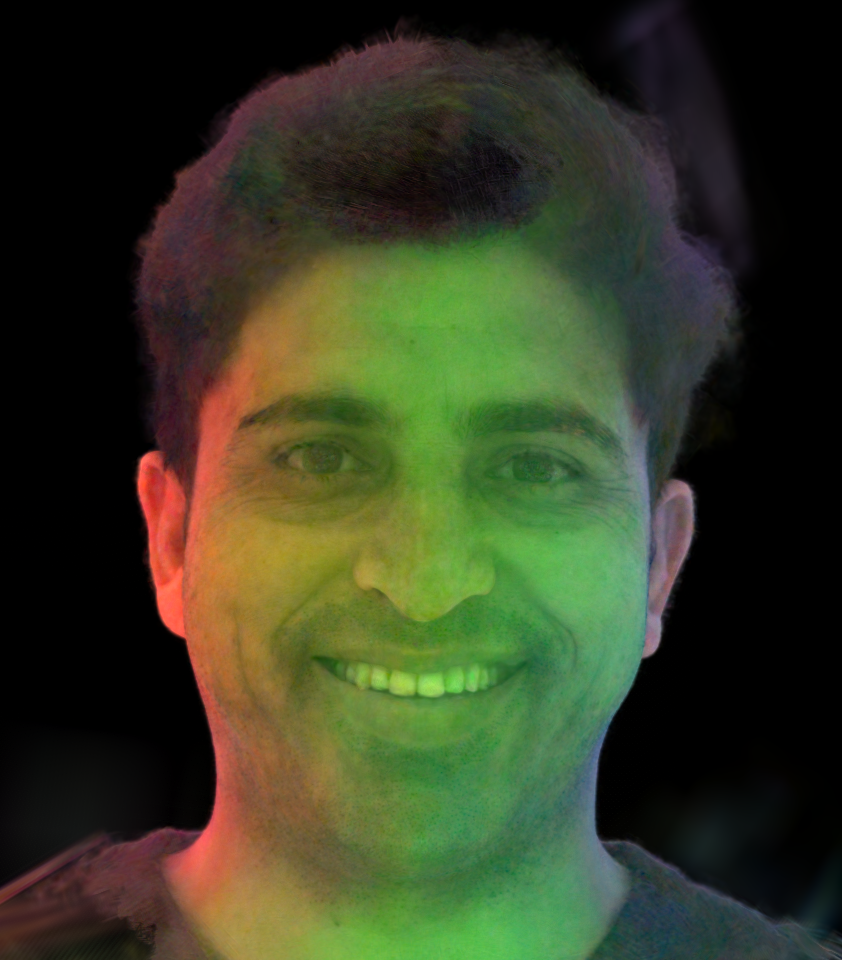}
        \includegraphics[width=\onesixthfigurewidth\linewidth]{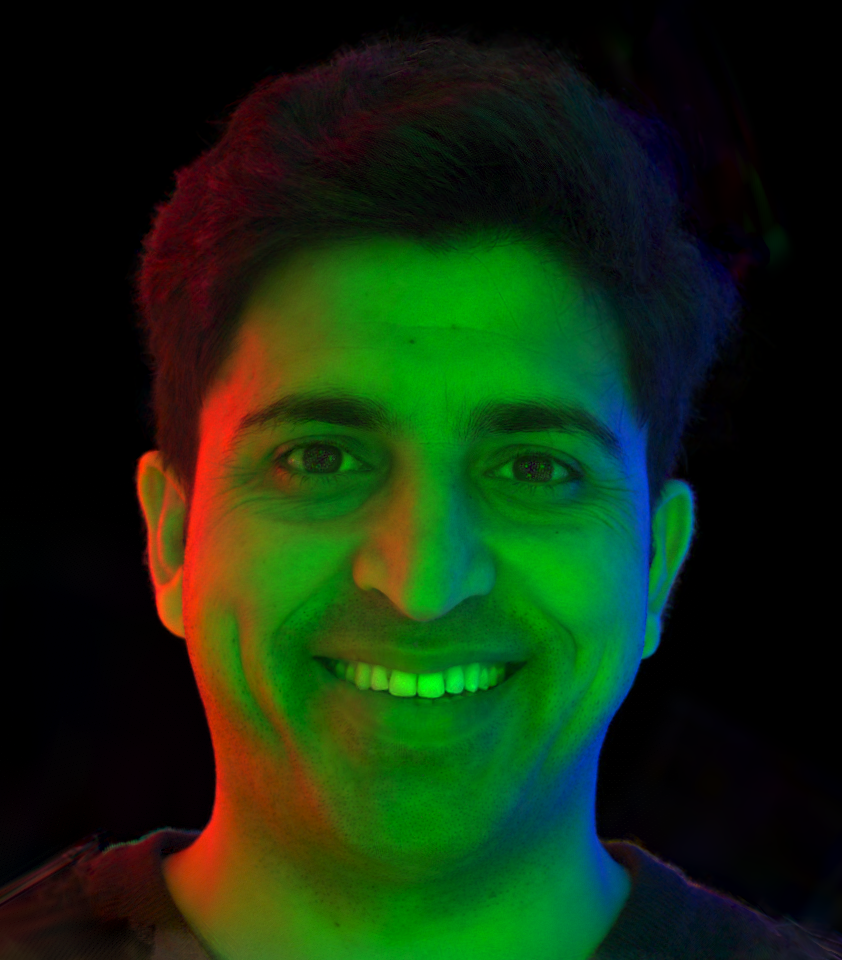}
    \end{minipage}

    \begin{minipage}[t]{\linewidth}
        \centering
        \includegraphics[width=\onesixthfigurewidth\linewidth]{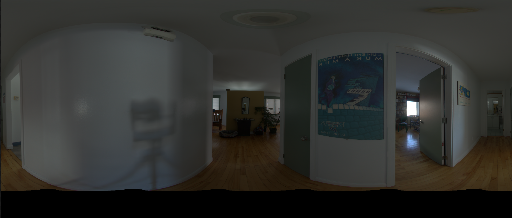}
        \includegraphics[width=\onesixthfigurewidth\linewidth]{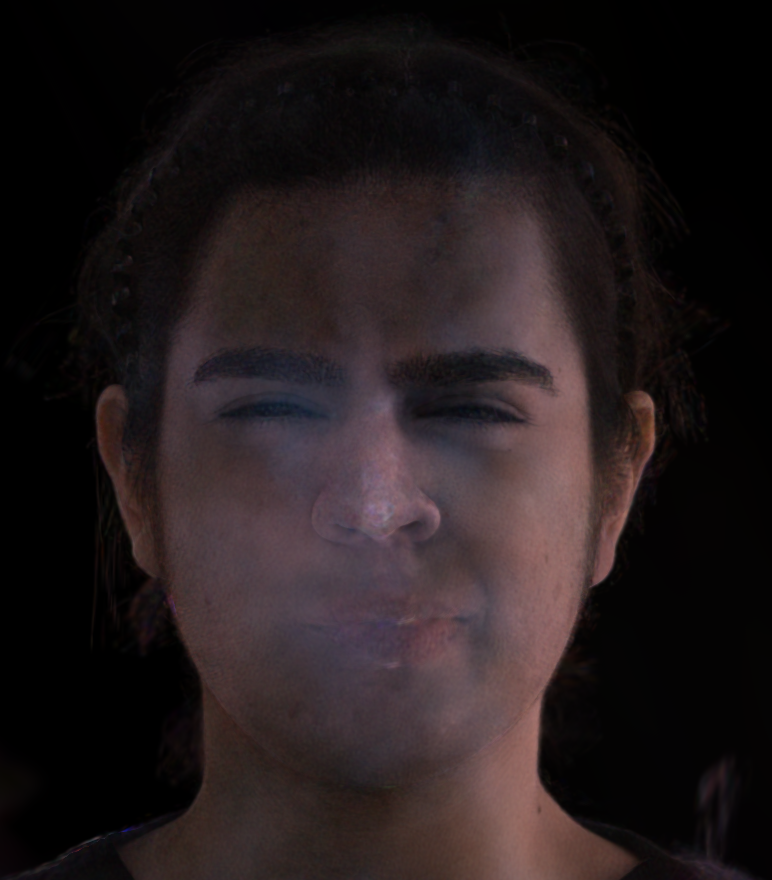}
        \includegraphics[width=\onesixthfigurewidth\linewidth]{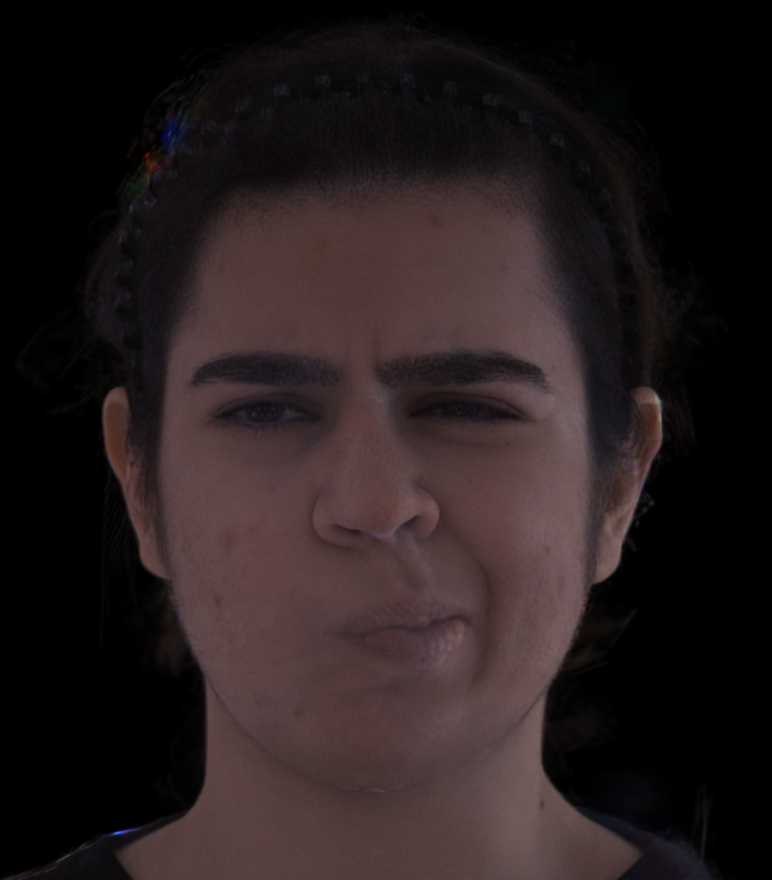}
        \includegraphics[width=\onesixthfigurewidth\linewidth]{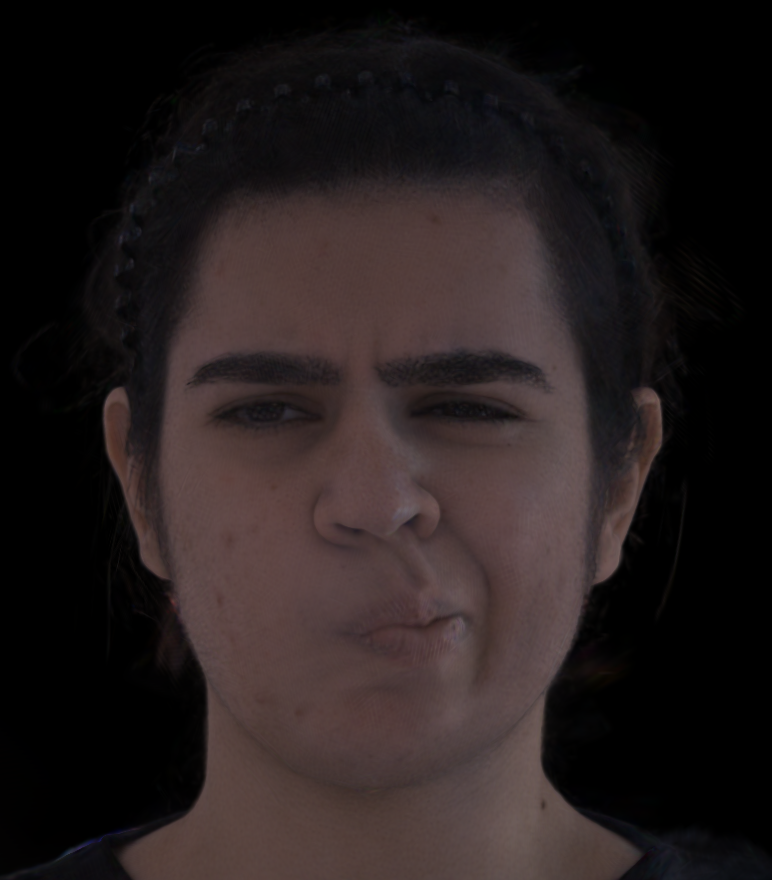}
        \includegraphics[width=\onesixthfigurewidth\linewidth]{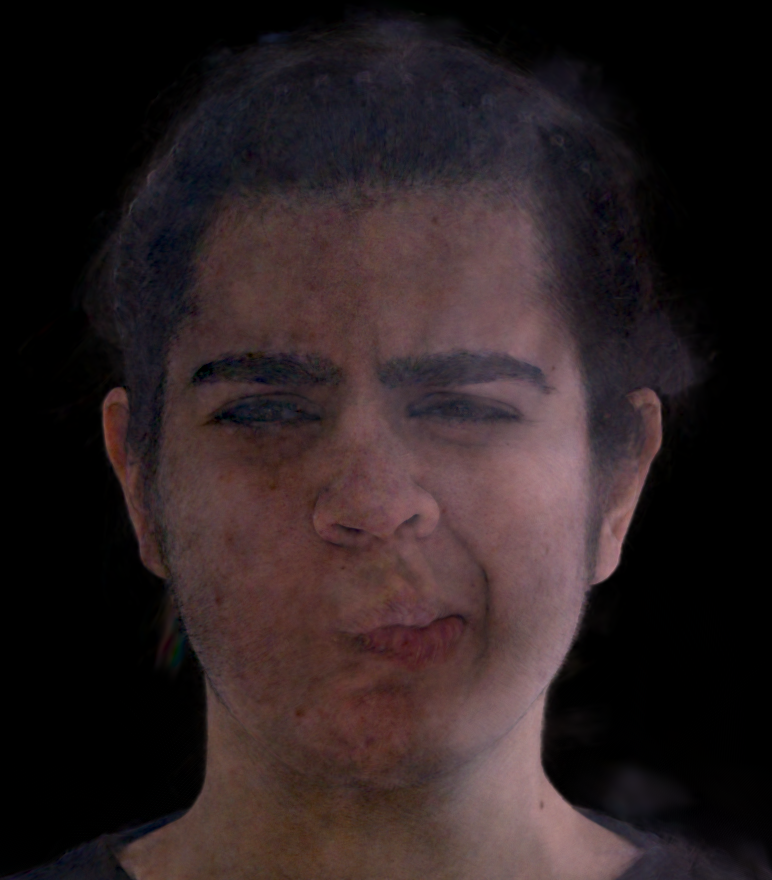}
        \includegraphics[width=\onesixthfigurewidth\linewidth]{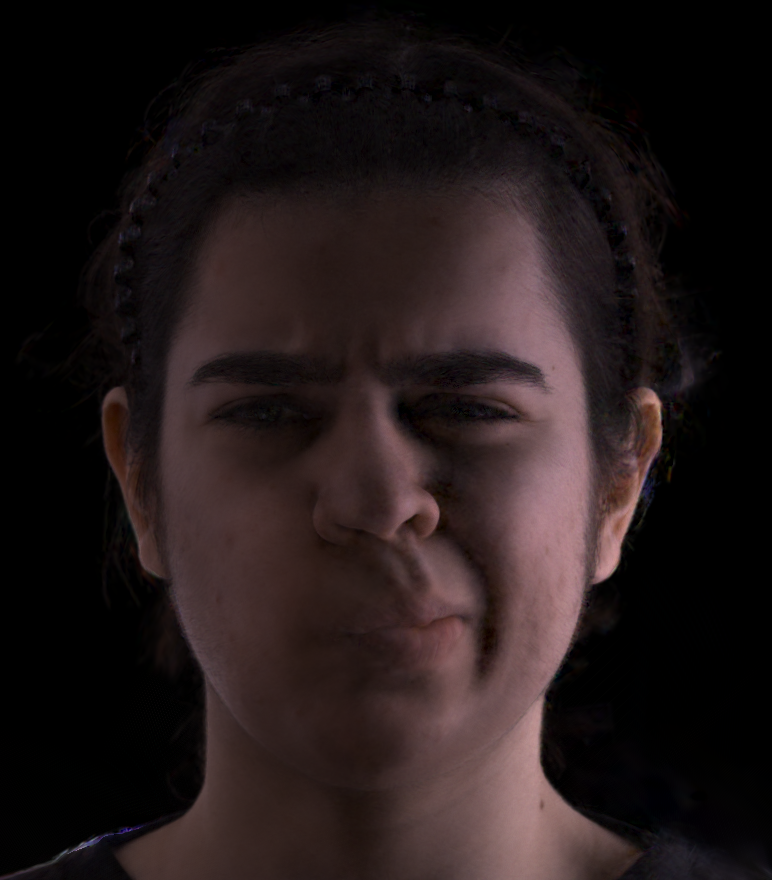}
    \end{minipage}

    \vspace{5pt}
    \begin{minipage}{\linewidth}
        \centering
        \begin{minipage}[t]{\onesixthfigurewidth\linewidth}
            \centering
            \subfloat{Light}
        \end{minipage}
        \begin{minipage}[t]{\onesixthfigurewidth\linewidth}
            \centering
            \subfloat{NL}
        \end{minipage}
        \begin{minipage}[t]{\onesixthfigurewidth\linewidth}
            \centering
            \subfloat{NL + ENV}
        \end{minipage}
        \begin{minipage}[t]{\onesixthfigurewidth\linewidth}
            \centering
            \subfloat{NL + LCL}
        \end{minipage}
        \begin{minipage}[t]{\onesixthfigurewidth\linewidth}
            \centering
            \subfloat{NL + TS}
        \end{minipage}
        \begin{minipage}[t]{\onesixthfigurewidth\linewidth}
            \centering
            \subfloat{Ours}
        \end{minipage}
    \end{minipage}
    \caption{Ablation study results about our physically inspired linear lighting branch for the appearance decoder $\colordecoder$.
     The input environment maps are shown on the left.
     The relighting results of four alternative baseline approaches (see detailed explanations in Section~\ref{sec:ablation_study}) and ours are shown on the right.
}
\label{fig:ablation}
\end{figure*}

Figure~\ref{fig:ablation} shows additional ablation study results using two environment maps, which demonstrate that the linear lighting branch of our appearance decoder $\colordecoder$ can significantly enhance the generalization performance for relighting.
}

\bibliographystyle{ACM-Reference-Format}
\bibliography{arxiv}

\end{document}